\documentclass[10pt,journal,compsoc]{IEEEtran}
\newcommand{\eg}{\textit{e}.\textit{g}.}
\newcommand{\etal}{\textit{et al}.}
\newcommand{\ie}{\textit{i}.\textit{e}.}
\newcommand{\etc}{\textit{etc}}
\newcommand{\aka}{\textit{a.k.a.}}

\ifCLASSOPTIONcompsoc
  \usepackage[nocompress]{cite}
\else
  \usepackage{cite}
\fi
\usepackage{times} 
\usepackage[table]{xcolor}
\usepackage{graphicx}  
\usepackage{float} 
\usepackage{subfigure}
\usepackage{amsmath}
\usepackage{url}
\usepackage{multirow}
\usepackage{makecell, multirow, tabularx}
\usepackage{adjustbox}
\usepackage{tabularx}
\usepackage{xspace}
\usepackage{booktabs}
\usepackage{color}
\usepackage{amssymb,mathrsfs,amsmath}
\usepackage{arydshln} 
\usepackage{amsfonts} 

\usepackage{tikz}
\usepackage{collcell}
\usepackage{etoolbox}

\newtoggle{inTableHeader}
\toggletrue{inTableHeader}
\newcommand*{\StartTableHeader}{\global\toggletrue{inTableHeader}}%
\let\OldTabular\tabular%
\let\OldEndTabular\endtabular%
\renewenvironment{tabular}{\StartTableHeader\OldTabular}{\OldEndTabular\StartTableHeader}%

\newcommand*{\MinNumber}{0.0}%
\newcommand*{\MidNumber}{1} %
\newcommand*{\MaxNumber}{12.75}%

\newcommand{\ApplyGradient}[1]{%
  \iftoggle{inTableHeader}{#1}{
    \ifdim #1 pt > \MidNumber pt
        \pgfmathsetmacro{\PercentColor}{max(min(100.0*(#1 - \MidNumber)/(\MaxNumber-\MidNumber),100.0),0.00)} %
        \hspace{-0.33em}\colorbox{green!\PercentColor!yellow}{#1}
    \else
        \pgfmathsetmacro{\PercentColor}{max(min(100.0*(\MidNumber - #1)/(\MidNumber-\MinNumber),100.0),0.00)} %
        \hspace{-0.33em}\colorbox{red!\PercentColor!yellow}{#1}
    \fi
  }}

\newcolumntype{R}{>{\collectcell\ApplyGradient}c<{\endcollectcell}}

\definecolor{hollywoodcerise}{rgb}{0.96, 0.0, 0.63}
\definecolor{lasallegreen}{rgb}{0.03, 0.47, 0.19}
\definecolor{hanpurple}{rgb}{0.32, 0.09, 0.98}
\definecolor{green(pigment)}{rgb}{0.0, 0.65, 0.31}
\usepackage{bbding} 

\usepackage{hyperref}
\hypersetup{colorlinks,linkcolor={red},citecolor={hollywoodcerise},urlcolor={red}}  

\begin{document}

\title{Deep Learning for Event-based Vision: \\A Comprehensive Survey and Benchmarks}

\author{ Xu~Zheng$^*$, Yexin~Liu$^*$, Yunfan~Lu, Tongyan~Hua, Tianbo~Pan, Weiming~Zhang, Dacheng~Tao, Lin~Wang$^\dagger$
}
\markboth{Journal of \LaTeX\ Class Files,~Vol.~14, No.~8, August~2015}%
{Shell \MakeLowercase{\textit{et al.}}: Bare Advanced Demo of IEEEtran.cls for IEEE Computer Society Journals}

\IEEEdisplaynontitleabstractindextext

\IEEEpeerreviewmaketitle

\IEEEtitleabstractindextext{%
\begin{abstract}
Event cameras are bio-inspired sensors that capture the per-pixel intensity changes asynchronously and produce event streams encoding the time, pixel position, and polarity (sign) of the intensity changes. Event cameras possess a myriad of advantages over canonical frame-based cameras, such as high temporal resolution, high dynamic range, low latency, etc. Being capable of capturing information in challenging visual conditions, event cameras have the potential to overcome the limitations of frame-based cameras in the computer vision and robotics community. In very recent years, deep learning (DL) has been brought to this emerging field and inspired active research endeavors in mining its potential. However, there is still a lack of taxonomies in DL techniques for event-based vision. We first scrutinize the typical event representations with quality enhancement methods as they play a pivotal role as inputs to the DL models. We then provide a comprehensive survey of existing DL-based methods by structurally grouping them into two major categories: 1) image/video reconstruction and restoration; 2) event-based scene understanding and 3D vision. We conduct benchmark experiments for the existing methods in some representative research directions \ie, image reconstruction, deblurring, and object recognition, to identify some critical insights and problems. Finally, we have discussions regarding the challenges and provide new perspectives for inspiring more research studies. 
\end{abstract}
}
\maketitle

\ifCLASSOPTIONcompsoc
\IEEEraisesectionheading{
\section{Introduction}\label{sec:introduction}}
\else
\section{Introduction}
\label{sec:introduction}
\fi
\IEEEPARstart{T}{he} 
breakthrough in neuromorphic engineering has recently extended the realms of sensory perception and initiated a novel paradigm that mimics biological vision. The bio-inspired sensors are also called \textbf{event} cameras~\cite{eventsurvey}. 
The difference is that each pixel in the camera operates independently, triggering a response (\ie, event) only when there is a brightness change, probably caused by motion or other visual changes~\cite{davies2021advancing}. 
In practice, the event streams captured by an event camera are sparse and are more concentrated along object boundaries. 
By contrast, canonical frame-based cameras record 
either a still image or video with a fixed frame rate.
Event cameras offer some other benefits, such as \textit{high temporal resolution} and \textit{high dynamic range}~\cite{baldwin2022time}. This means that an event camera can capture high-quality data either in extreme lighting or high-speed motion conditions~\cite{issafe}. 
Therefore, event cameras have the potential to overcome the limitations of frame-based cameras in the computer vision and robotic fields and have helped the community in solving various downstream tasks, \eg, corner tracking\cite{gehrig2020eklt}, Simultaneous Localization And Mapping (SLAM)\cite{hidalgo2022event,2021Comparing}, image and video restoration\cite{chen2021indoor,MostafaviIsfahani2018EventBasedHD,messikommer2022multi,scheerlinck2020fast}, object detection and segmentation\cite{mondal2021moving,2019Event,perot2020learning,evsegnet,videotoevent,issafe}. 

\begin{figure*}[t!]
\centering
\includegraphics[width=.97\textwidth]{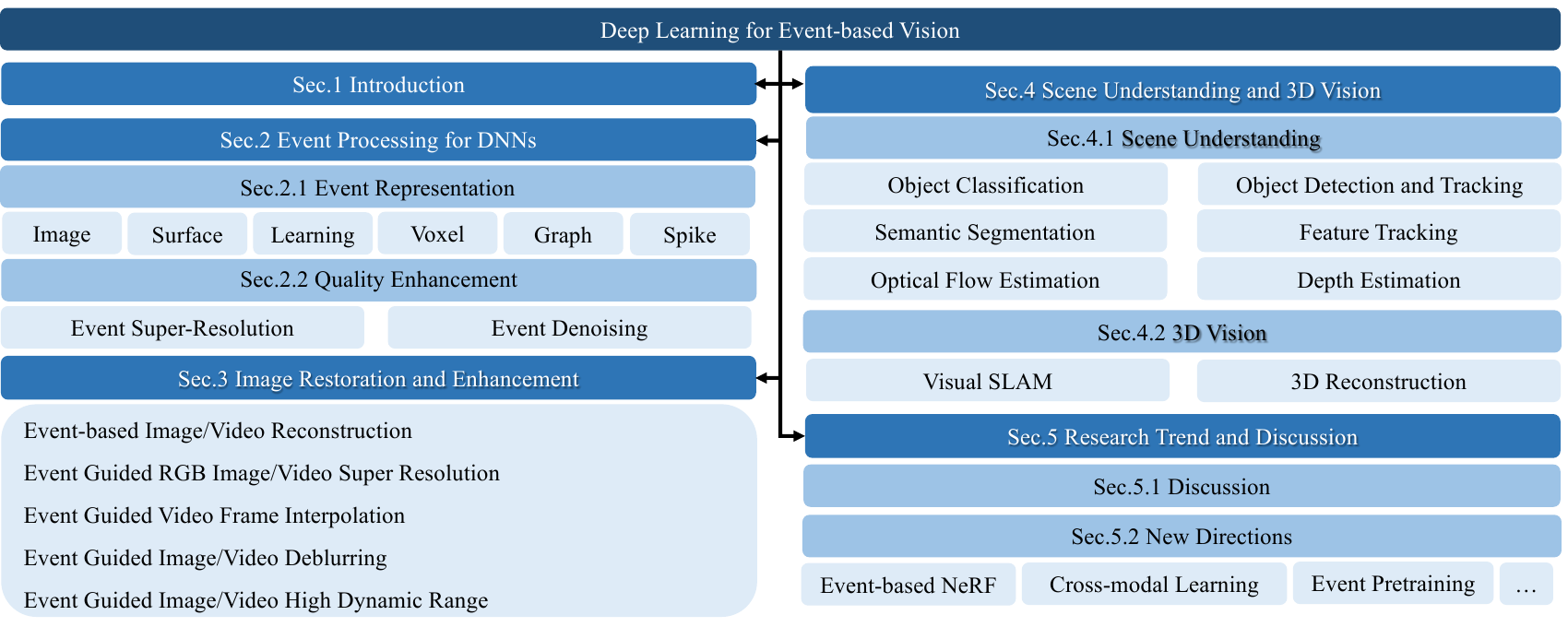}
\caption{The structural and hierarchical taxonomy of event-based vision with deep learning.} \label{fig:1-Overall}
\end{figure*}

\noindent \textbf{Motivation:} 
The first application of an event camera is in a hardware-based Deep Neural Network (DNN) system called CAVIAR~\cite{serrano2009caviar}, predating the first software-based DNN system from the computer vision community by approximately a decade. This early development also includes significant contributions from solid-state circuit papers that detailed the design and initial applications of camera chips~\cite{5537149,4444573}. Recently, deep learning (DL) has received great attention in this emerging area with amounts of techniques developed based on various purposes. For example, DL-based feature trackers~\cite{messikommer2023data} achieve better accuracy than the non-DL-based methods (See Tab.~\ref{Tab:featuretracking}). 
We explore some fundamental questions that have been driving this research area when examining the representative methods.

Previously, Gallego \etal~\cite{eventsurvey} provided the first overview of the event-based vision with a particular focus on the principles and conventional algorithms. However, DL has invigorated almost every field of event-based vision recently, and remarkable advancements in methodologies and techniques have been achieved. Therefore, a more up-to-date yet insightful survey is urgently needed for capturing the research trends while clarifying the challenges and potential directions.

We survey the DL-based methods by focusing on three important aspects: 
1) DNN input representations for event data and quality enhancement
(Sec.~\ref{sec:Event Processing for DNNs}); 2) Current research highlights by typically analyzing two hot fields, image restoration and enhancement (Sec.\ref{sec:Image Reconstruction and Restoration}), scene understanding and 3D vision (Sec.~\ref{sec:Scene Understanding and 3D Vision}); 3) Potential future directions, such as event-based neural radiance for 3D reconstruction, cross-modal learning, and event-based model pertaining {(Sec.~\ref{sec:new-direction})}. 

\noindent \textbf{Contributions:} In summary, the main contributions of this paper are five-fold: 
(I) We provide a comprehensive overview of the existing event representations as well as the quality enhancement methods for events.
(II) We provide a summary of how existing DL-based approaches for event-based vision address challenges and offer insights for the computer vision community, including image restoration and enhancement and high-level scene understanding tasks.
(III) We discuss some open problems and challenges on DL with events and identify future research directions, providing guidance for future developments in this field.
(IV) We ask and discuss some broadly focused questions as well as some potential problems to answer the concern and dive deeply into the event-based vision.
(V) Furthermore, we create an open-source repository that provides a taxonomy of all mentioned papers and code links. Our open-source repository will be updated regularly with the latest research progress, and we hope this work can bring sparks to the research of this field. The repository link is \url{https://github.com/vlislab22/Deep-Learning-for-Event-based-Vision}.
Meanwhile, we benchmark and highlight some representative event-based and event-guided vision tasks, \eg, in Tab.~\ref{Tab:Classification_exp} and \ref{tab:optical_flow}, to identify the critical insights and problems for future studies. \textit{Due to the lack of space, some experimental results can be found in the supplementary material}.

In the following sections, we discuss and analyze the recent advances in DL methods for event-based vision. The structural and hierarchical taxonomy of this paper is depicted in Fig.~\ref{fig:1-Overall}.

\section{Event Processing for DNNs}
\label{sec:Event Processing for DNNs}
Event cameras generate events when individual pixels detect the relative logarithm intensity change. Consequently, sparse and asynchronous event streams are generated. This inherent sparsity offers immediate advantages, such as low latency and low computational requirements for postprocessing systems~\cite{liu2019event}. Event cameras pose a distinctive shift in the imaging paradigm regarding how visual information is captured, making it impossible to directly apply the DNN models taking the image- or tensor-like inputs. 
Therefore, we first analyze the event representations, which are used as inputs to DNNs (Sec.~\ref{sec:Event Representation}). Moreover, as event data are often hampered by noises (especially in low-light conditions) and limited spatial resolution (in particular for DAVIS cameras), we analyze the DL methods that super-resolve and denoise the event streams for improving the learning performance (Sec.~\ref{sec:Quality Enhancement for Events}).

\subsection{Event Representation}
\label{sec:Event Representation}
We first review how an event camera responds asynchronously to each independent pixel and generates a stream of events. 
An event is interpreted as a tuple  $(\textbf{u}, t, p)$, which is triggered whenever a change in the logarithmic intensity $L$ surpasses a constant value (threshold) $C$, formulated as follows: 
\begin{equation}
    p = \left\{
    \begin{aligned}
         +1 ,& L(\textbf{u},t) -L(\textbf{u}, t-\Delta t) \geq C \\
         -1 ,& L(\textbf{u},t) -L(\textbf{u}, t-\Delta t) \leq -C \\
         0  ,& other\\
    \end{aligned}
    \right.
\end{equation}

where $\textbf{u}=(x,y)$ is the pixel location, $t$ is the timestamp and $p\in \{-1, 1\}$ is the polarity, indicating the sign of brightness changes 
(1 and -1 represent positive and negative events, respectively), and $p=0$ means that there are no events. 
$\Delta t$ is a time interval since the last event at pixel $\textbf{u}=(x,y)$. A number (or stream) of events are triggered 
which can be denoted as: 

\begin{equation}
    \mathcal{E}=\{e_i\}_{i=1}^N = \{\textbf{u}_i,t_i,p_i\}, i \in N,
\end{equation}
For more details of the event generation model, we refer readers to \cite{eventsurvey}.
In a nutshell, this particular type of data makes it difficult to apply the DNN models predominantly designed for frame-based cameras. 
Therefore, it is pivotal to exploit the effective alternative representations of event data for mining their visual information and power~\cite{ChainSAE}.
In the following, we review the representative event representation methods, which can be divided into six categories: image-based, surface-based, learning-based, voxel-based, graph-based, and spike-based representations, as shown in Tab.~\ref{tab:eventrepresentations}.

\begin{table}[t!]
\centering
\caption{Representative event representations, 
(SP: Steering Prediction; OF: Optical Flow Estimation; Cls: Classification; CD: Corner Detection; GR: Gesture recognition; Recon: Reconstruction; DE: Depth Estimation; N/A: Not Available.)}
\label{tab:eventrepresentations}
\resizebox{\linewidth}{!}{
\begin{tabular}{llll}
\toprule
Categories  & Event Representation & Tasks &Dimensions \\  \midrule
\multirow{5}{*}{Image} & Maqueda \etal \cite{maqueda2018event} & SP & (2,H,W) \\ 
& EV-FlowNet \cite{zhu2018ev} &OF & (4,H,W) \\ 
& Event Image~\cite{wang2019ev} & Cls & (4,H,W)  \\
&AMAE\cite{deng2020amae} & Cls & (2,H,W) \\ 
&Bai \etal \cite{bai2022accurate} &Cls & (3,H,W)  \\ 
& MVF-Net \cite{deng2021mvf} &Cls & N/A \\ 

\midrule
\multirow{2}{*}{Learning}& EST~\cite{EST} & Cls\&OF & (2,B,H,W) \\
\multicolumn{1}{l}{} & \multicolumn{1}{l}{Matrix-LSTM~\cite{Matrixlstm}} & \multicolumn{1}{l}{Cls\&OF}  & (B,H,W) \\ \midrule
\multirow{13}{*}{Surface} & Timestamps Image~\cite{park2016performance} & Cls & N/A   \\
& SAE~\cite{SAE} & Cls & (2,H,W) \\
& Time Surface~\cite{Hots} & Cls & (2,H,W)  \\
& Sorted Time Surface~\cite{alzugaray2018ace} & Cls & (2,H,W) \\
& Event Histogram~\cite{maqueda2018event} & Cls & (2,H,W)  \\
& HATS~\cite{sironi2018hats} & Cls & (2,H,W)  \\
& IETS~\cite{IETS} & Cls & (3,H,W) \\
& SITS~\cite{2019F25} & CD & (2,H,W)\\ 
& Chain SAE~\cite{ChainSAE} & Cls & (2,H,W) \\
& DiST~\cite{N-ImageNet} & Cls & (2,H,W)  \\
& TORE~\cite{baldwin2022time} & Cls & (2,K,H,W)\\ \midrule

\multirow{4}{*}{Voxel} 
&  Zhu \etal \cite{zihao2018unsupervised} &OF & (B,H,W) \\
&  Rebecq \etal \cite{gu2020tactilesgnet}& Recon & (B,H,W) \\
& Ye \etal \cite{2020Unsupervised30}& DE\&OF  & (B,H,W) \\
& TORE \cite{baldwin2022time}& Recon, \etc & (2,K,H,W) \\ 
\midrule

\multirow{2}{*}{Graph} 
&  RG-CNN \cite{RGCNNs} & Cls & N/A \\
&  EV-VGCNN \cite{Ev-VGCNN} & Cls & (H,W,A)  \\
\midrule

\multirow{3}{*}{Spike} 
& Tactilesgnet \cite{gu2020tactilesgnet}& Cls & N/A \\ & Botzheim \etal \cite{botzheim2012human}&GR & N/A  \\ 
\bottomrule
\end{tabular}}
\end{table}

\subsubsection{Image-based Representation}
The straightforward solution to adopt events to the existing DL methods is to stack (or convert) events to synchronous 2D image representations (similar to frame-based cameras) as the inputs to DNNs. 
For example, Moeys \etal ~\cite{moeys2016steering} proposed the first CNN driven by DVS frames to address the blurring issue in a predator-prey robot scenario. This study also marks the initial utilization of event count DVS images to guide a DNN using DVS data. The channels of image-based representation are often set to preserve polarities, timestamps, and event counts~\cite{maqueda2018event,zhu2018ev,deng2020amae,bai2022accurate,deng2021mvf}. Based on how images are formed, we divide the prevailing methods into four types.

\noindent \textbf{Stack based on polarity:} Maqueda \etal \cite{maqueda2018event} set up two separate channels to evaluate the histograms for positive and negative events to obtain two-channel event images, which are finally merged together into synchronous event frames. 

\noindent \textbf{Stack based on timestamps: }To consider the importance of event counts and timestamps for holistic information, \cite{deng2020amae, bai2022accurate,MostafaviIsfahani2018EventBasedHD} take the timestamps of events into consideration.%

\noindent\textbf{Stack based on the number of events:} 
Due to the uneven triggers of events within fixed time intervals, 
another stacking strategy is proposed to sample and stack events in a fixed constant number~\cite{ multiplemodalities, messikommer2020event}.

\noindent \textbf{Stack based on timestamps and polarity:}
In Ev-gait ~\cite{wang2019ev}, event streams are converted to frame-like
representation with four channels, containing the positive or negative polarities in two channels and the temporal characteristics in another two channels. Also, some research has focused on exposure time control in event-based systems. Liu \etal~\cite{liu2018adaptive,liu2022edflow} have explored dynamic control of exposure time and inter-slice time interval to optimize the quality of slice features. This adaptive control helps to ensure the robustness of the model in dynamic scenes with varying motion speeds and scene structures.
\textit{Detailed mathematical formulations can be found in Sec.1.1 of the supplementary material.}




\subsubsection{Surface-based Representation}
The first surface-based representation, \ie, Surface of Active Events (SAE)~\cite{SAE}, it maps the event streams to a time-dependent surface and tracks the activity around the spatial location of the latest event $e_i$. 
Different from the basic image-based representation which utilizes intensity images to provide context content, the SAE achieves this through a totally different perspective, \ie, the temporal-spatial perspective. Specifically, the time surface of the $i$ th event $e_i$ can be formulated as a spatial operator acting on the neighboring region of $e_i$:
\begin{equation}
\label{SAE}
    \tau_i([x_n,y_n]^T,p) = \underset{j\le i}{max}\{t_j|[x_i+x_n,y_i+y_n],p_j=p\} 
\end{equation}
where $x_n \in \{-r,r\}$ is the horizontal coordinate of $e_i$, $y_n \in \{-r,r\}$ is the vertical coordinate of $e_i$, $p_j \in \{-1,1\}$ is the polarity of the $j$ th event $e_j$, $t_j$ is the timestamp of $e_j$ and $r$ is the radius of the neighboring region used to obtain the time surface. As shown in Eq.~\ref{SAE}, the time surface $\tau_i([x_n,y_n]^T,p)$ encodes the time context in the $(2r+1)\times(2r+1)$ neighborhood region of $e_i$, hence maintaining both temporal and spatial information for downstream tasks. 

However, the timestamps of events monotonically increase, which causes the temporal values in the surface from 0 to infinity~\cite{ChainSAE}. 
Therefore, appropriate normalization approaches are required to preserve the temporal-invariant data representation from raw SAE by mapping the timestamps to $[0,1]$. Basic normalization methods are directly applied to time surfaces~\cite{alzugaray2018ace, Hots, time-window}, such as the min-max~\cite{alzugaray2018ace}, time window~\cite{time-window}, \etc. 

All these above normalization methods rely on empirical parameter tuning, leading to additional computational costs. To avoid this problem, sort normalization is employed by Alzugaray \etal ~\cite{alzugaray2018ace} to sort all the timestamps within an SAE at each pixel. 
However, though this method alleviates the dependence on parameter tuning, the by-product of time complexity impedes the whole procedure's efficiency.
To build an efficient SAE and achieve robust speed-invariant characteristics, Manderscheid \etal ~\cite{2019F25} introduced a normalization scheme to obtain the Speed Invariant Time Surface (SITS). The SITS updates the time surface of each incoming event according to its neighborhood with the radius $r$.
Overall, when large $r$ is adopted, the SITS updates the time surface when a new event is triggered, thus leading to inefficiency in the on-demand tasks. Lin \etal ~\cite{ChainSAE} suggested solving and alleviating this imbalance between normalization and the number of events by using a chain update strategy. 


\subsubsection{Voxel-based Representation}
The voxel-based representations map the raw events into the nearest temporal grid within temporal bins.
The first spatial-temporal voxel grid is proposed in ~\cite{zihao2018unsupervised}, inserting events into volumes using a linearly weighted accumulation to improve the resolution along the temporal domain.
This spatial-temporal voxel grid is also used in some following works~\cite{rebecq2019events, 2020Unsupervised30}.
More recently, a time-ordered recent event volume is proposed in ~\cite{baldwin2022time}, aiming at compactly maintaining raw spike temporal information with minimal information loss.


\subsubsection{Graph-based Representation}
Aiming at preserving the sparsity of events, graph-based approaches transform the raw event streams within a time window into a set of connected nodes. Bi \etal~first proposed a residual graph CNN architecture to obtain a compact graph representation for object classification~\cite{ASL-DVS, RGCNNs}. The graph CNN preserves the spatial-temporal coherence of input events while avoiding large computational costs. More recently, Deng \etal~proposed a voxel graph CNN which aims at exploiting the sparsity of event data~\cite{Ev-VGCNN}. The proposed EV-VGCNN~\cite{Ev-VGCNN} is a lightweight voxel graph CNN while achieving the SOTA classification accuracy with very low model complexity.


\subsubsection{Spike-based Representation}
Due to the sparsity and asynchronous nature of event streams, most of the above representations consider the timestamps. Different from the standard DNN models, spiking neural networks (SNNs)\cite{Hfirst,gu2020tactilesgnet} are advantageous in that they incorporate the concept of time into operational models. Therefore, SNNs better fit the biological neuronal mechanism by using signals in the form of pulses (discontinuous values) to convey visual information. SNNs are applied to extracting features from events asynchronously to solve diverse tasks, such as object classification\cite{gu2020tactilesgnet}. However, due to the complex dynamics and non-differentiable nature of the spikes, two challenges exist: 1) Well-established back-propagation methods cannot be applied to the training process, leading to a long training time and high costs; 2) Specialized and effective hardware and algorithms are lacking. Consequently,
their accuracy cannot exceed SOTA methods. Future research could explore more in this direction. 


\subsubsection{Learning-based Representation}
The aforementioned event representations are mainly designed for a specific task and cannot be generally and flexibly applied to other tasks. 
To this end, Gehrig \etal ~\cite{EST} proposed the first learning-based approach to convert asynchronous raw events into tensor-like inputs, which can be flexibly applied to diverse downstream tasks. 
In particular, a multi-layer perceptron (MLP) is adopted to learn the coordinates and timestamps of events to obtain grid-like representations. 
Moreover, some methods~\cite{phasedlstm, Matrixlstm} extract features from events using Long Short-Term Memory (LSTM). A representative approach, Matrix-LSTM~\cite{Matrixlstm}, utilizes a grid of LSTM cells to integrate information in the temporal axis. This approach follows a fully differentiable procedure that extracts the most relevant event representations for downstream tasks. 


\noindent \textbf{\textit{Remarks}}
Table~\ref{tab:eventrepresentations} outlines six types of mainstream representation methods for DNNs. Different event representations offer unique advantages and considerations when applied to various tasks. 
(1) Image-based representation enables seamless integration with traditional deep learning algorithms, allowing for applications in scene understanding and 3D vision tasks. (2) Surface-based representation provides spatial-temporal context and preserves temporal information to a certain extent. (3) Voxel-based representation enhances resolution and preserves raw spike temporal information by mapping raw events into temporal grids. (4) Graph-based representation maintains sparsity and coherence while minimizing computational costs, achieving high classification accuracy. (5) Spike-based representation, buttressed by SNNs, offers advantages in asynchronous processing, efficiency, noise robustness, and compatibility with neuromorphic hardware. (6) Learning-based representation aims to discover optimal event representations, adapting to task-specific requirements. However, practical factors such as computational complexity and data availability should be considered. 
\textbf{\textit{Overall, the selection of event representation should be considered based on the requirements of the task, and the trade-offs between complexity, computational efficiency, and interpretability.}} Further studies are urgently needed to explore more generic event representations for a wider range of tasks.

\subsection{Quality Enhancement for Events}
\label{sec:Quality Enhancement for Events}

Event cameras, \eg, DAVIS346~\cite{brandli2014real}, are with relatively low resolution\textemdash 346$\times$240, while some event cameras, \eg, EVK4 from Prophesse \footnote{https://www.prophesee.ai/event-camera-evk4/}, show higher spatial resolution up to, \eg,  1280$\times$720 px. These cameras often suffer from unexpected noise in the captured event data, especially in challenging visual conditions and event representation processes. 
Research has been recently conducted to improve the spatial resolution of events and denoise the events to achieve higher quality. 

\subsubsection{Event Super-Resolution}
Different from image super-solution (SR), event SR requires distribution estimation in both spatial and temporal dimensions.
Li \etal ~\cite{li2019super} first proposed to solve the spatial-temporal SR problem of the LR event image.
Later on,  Wang \etal ~\cite{wang2020joint} proposed to bridge intensity images and events from the Dynamic Vision Sensor (DVS) via joint image filtering, so as to obtain motion-compensated event frames with high-resolution (HR) and less noise.
The first DL-based approach is proposed by Duan \etal~\cite{duan2021eventzoom}, which addresses the joint denoising and SR by using a multi-resolution event recording system and a 3D U-Net-based framework, called EventZoom. In particular, it incorporates event-to-image reconstruction to achieve resolution enhancement.
Furthermore, Li \etal \cite{li2021event} proposed an SNN framework with a constraint learning mechanism to simultaneously learn the spatial and temporal distributions of event streams.
Recently, Weng \etal \cite{weng2022boosting} introduced a Recurrent Neural Network (RNN) that employs temporal propagation and spatial-temporal fusion net to ensure the restoration abilities of fine-grained event details without any auxiliary high-quality and HR frames.

\noindent \textbf{\textit{Remarks}:} 
Though these methods achieve plausible SR results, the spatial-temporal distribution estimation leads to high latency for large factor SR, \eg, $\times 16$. \textbf{\textit{Future research could focus on reducing inference latency and lightweight network design.}}
\subsubsection{Event Denoising}
\label{subsec:event+denoising}

The presence of random noise, such as thermal noise and junction leakage currents, leads to background activity (BA) where an event is generated without any log-intensity change. To address this issue, both handcrafted and learning-based denoisers are proposed. Handcrafted filter-based methods utilize various types of filters to eliminate background activity. These methods include bio-inspired filters\cite{barrios2018less}, hardware-based filters\cite{khodamoradi2018n}, spatial filters\cite{SAE}, and temporal filters\cite{IETS,wang2020joint}. Guo \etal \cite{guo2022low} introduced a novel framework that quantifies denoising algorithms more effectively by measuring receiver operating characteristics using known mixtures of signal and noise DVS events. \textit{For experiment results of different filter-based methods, refer to Fig. {\color{red}6} in the supplementary material.}

DL-based methods have also been introduced in ~\cite{baldwin2020event, fang2022aednet}. One representative framework is EDnCNN proposed by Baldwin \etal \cite{baldwin2020event}. It transforms neighboring events into voxels and differentiates noise using an Event Probability Mask (EPM). 
Event data is conventionally converted into regular tensors for DNNs, compromising its inherent asynchronous properties. To address this issue, aAEDNet\cite{fang2022aednet} adeptly utilizes the correlation of the irregular signal within the spatial-temporal range, preserving its original structural integrity without the need for transformation.
\textit{For detailed qualitative and quantitative experiment results, refer to Fig. {\color{red} 5} and Tab. {\color{red} 4} in the supplementary material.}
~\cite{lin2022dvs,hu2021v2e,videotoevent,rebecq2018esim}.

For both kinds of methods, the availability of labeled data is paramount. In this context, DVS simulation models~\cite{lin2022dvs,hu2021v2e,videotoevent,rebecq2018esim} prove to be invaluable. Primarily, these models facilitate the prediction of events, whether ideal or non-ideal, from known visual inputs. Secondly, they are capable of generating labeled noise events. This dual functionality not only enhances the training and validation of algorithms but also supports the development of more robust event-based vision systems by providing a comprehensive framework for data generation and analysis.

\noindent \textbf{\textit{Remarks}:}  
Due to its differential imaging mechanism, the event camera is sensitive to various types of noise.
Thus, denoising methods are one of the bases of event-based vision. \textbf{\textit{In future research, a more general DL-based denoising pipeline, which can be applied to various event-based vision tasks while has similar computation efficiency compared with prior methods, is worth exploring.}
}
\section{Image Restoration and Enhancement}
\label{sec:Image Reconstruction and Restoration}
Event cameras hold immense potential for leveraging event cameras in the reconstruction and restoration of HDR images and high frame-rate videos. 
However, their unique imaging paradigm presents a challenge when applying vision algorithms designed for frame-based cameras.  To address this challenge and bridge the gap between event-based and standard computer vision, many methods have been proposed to reconstruct intensity video frames or images from events.

In this paper, we group the prevailing methods into two major types: event-based image (or video) reconstruction (with only events as inputs) and event-guided image restoration (hybrid inputs of events and frames). For the former, the main problem is how to fully explore the visual information, \eg, edge, from events with DNNs to reconstruct high-quality intensity images or video frames; while the latter explores how to fuse frames and events while leveraging the advantages of events, \eg, HDR, to benefit the image restoration process. We now review the state-of-the-art (SOTA) techniques in the following sections.  

\subsection{Event-based Intensity Reconstruction}
\label{event-to-image}

\begin{figure}[t!]
    \centering    \includegraphics[width=0.49\textwidth]{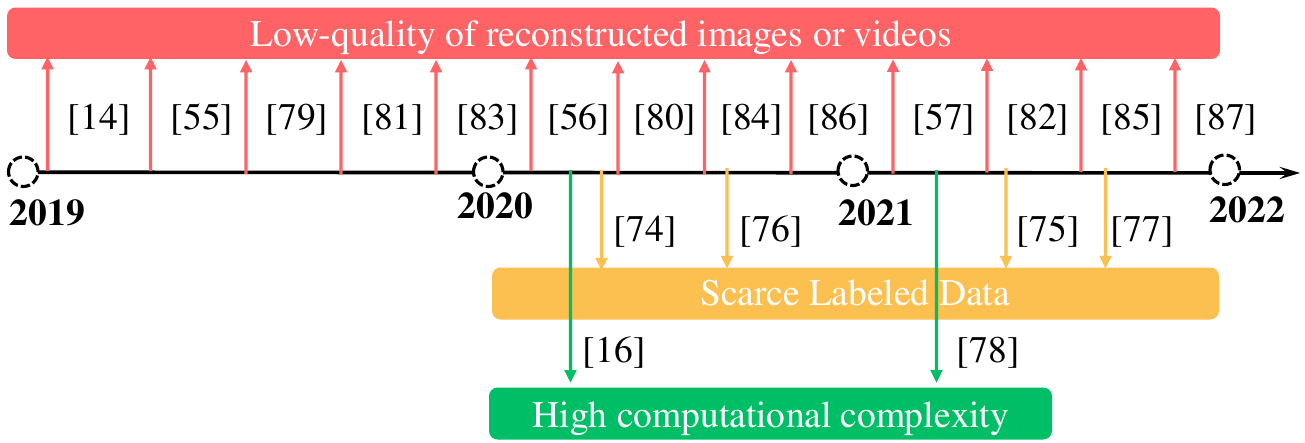}
    \caption{Methods for event-based intensity reconstruction.}
    \label{fig: Image Reconstruction}
\end{figure}

\begin{figure}[t!]
    \centering
    \includegraphics[width=0.50\textwidth]{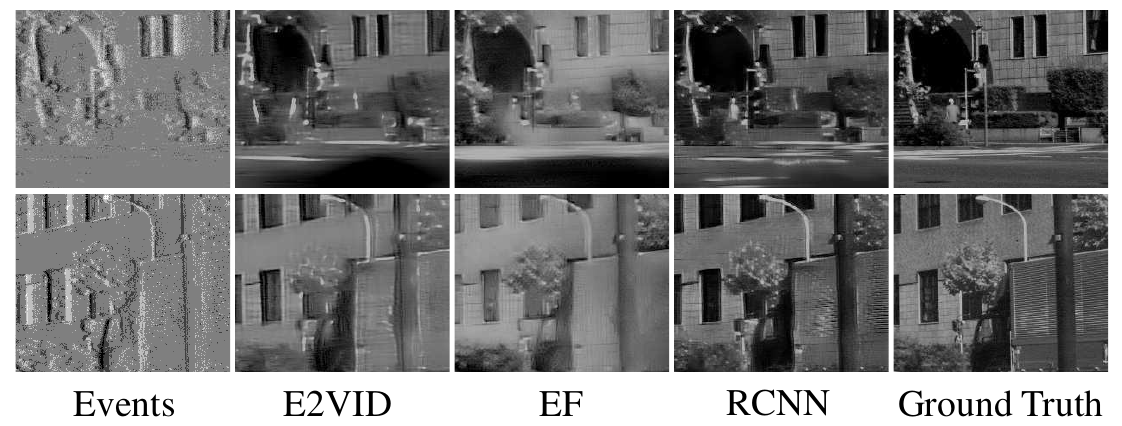}
    \caption{Visual examples of some SOTA methods for video reconstruction (E2VID ~\cite{rebecq2019high}, EF~\cite{stoffregen2020reducing}, RCNN~\cite{zou2021learning}).}
    \label{fig: results video recons}
\end{figure}

\textit{\textbf{Insight}}: This task learns a mapping from a stream of events to a single intensity image or sequence of images (\ie, video). The mapped results allow for applying the off-the-shelf DL algorithms\textemdash developed for frame-based cameras\textemdash to learning downstream tasks. From our review, intensive research has been devoted to achieving this task, as summarized in Fig.~\ref{fig: results video recons}, Fig.~\ref{fig: Image Reconstruction}, and Tab.~\ref{Reconstruction}.

Early methods rely on the assumption about the scene structure (or motion dynamics)~\cite{cook2011interacting} or event integration
with regularization terms~\cite{munda2018real} to reconstruct intensity images.  
However, these methods suffer from artifacts due to the direct event integration, and the reconstructed intensity images are not photo-realistic enough. DL-based methods, by contrast, bring significant accuracy gains. 
In this paper, we analyze the SOTA deep learning methods based on the following challenges:
1) A lack of large-scale datasets for training deep networks; 
2) High computational complexity and high latency; 
3) The low-quality of reconstructed images or videos, \eg, relatively low resolution and blurred images. 

\begin{table}[t!]
\centering
\caption{Qualitative comparison results of some image reconstruction methods~\cite{zhang2022formulating} on event dataset~\cite{mueggler2017event}. The experiment utilized the Pylops library on an Intel i7-8650U CPU laptop, assuming known optical flow and NIWE parameters.}
\resizebox{0.95\linewidth}{!}{
\begin{tabular}{cccccc}
\toprule
Method & Type & MSE~$\downarrow$ & SSIM~$\uparrow$ & LPIPS~$\downarrow$ & Time\\ \midrule
E2VID ~\cite{rebecq2019high} & DL-based & 0.069 & 0.395 &   0.438 &   \textbf{0.2448 s}\\
ECNN~\cite{stoffregen2020reducing} & DL-based &   \textbf{0.056} & 0.416 & 0.442 & 0.2839 s \\
BTEB~\cite{paredes2021back} & DL-based & 0.090 & 0.357 & \textbf{0.520} & 0.4059 s\\
Tikhonov ~\cite{zhang2022formulating} & Model-based &  0.121 &   0.356 & 0.485 & 0.4401 s\\
TV~\cite{zhang2022formulating} & Model-based & 0.113 & 0.386 & 0.502 & 4.0443 s\\
CNN~\cite{zhang2022formulating} & DL-based & 0.080 & \textbf{0.437} & 0.485 & 28.3904 s\\
\bottomrule
\end{tabular}}
\label{Reconstruction}
\end{table}

\begin{table*}[t!]
\centering
\caption{
Comparison of the representative event-guided video frame interpolation (VFI) methods.
}
\resizebox{\linewidth}{!}{
\begin{tabular}{llllccccc}
\toprule
Publication                                 & Methods & Highlight                                               & Event Representation                          & Optical Flow & Deblurring & Supervised & Backbone    & Dataset\\ \midrule
CVPR 2021   &TimeLens\cite{tulyakov2021time}         &  synthesis-based and ﬂow-based                           & voxel grid\cite{zihao2018unsupervised}        & \Checkmark   & \XSolidBrush & \Checkmark      & CNN   & HQF,Vimeo90k,GoPro,Middlebury,HS-ERGB    \\
CVPR 2021   &EFI-Net\cite{paikin2021efi}             &  different spatial resolutions                           & voxel grids\cite{zihao2018unsupervised}       & \XSolidBrush & \XSolidBrush & \Checkmark       & CNN   & Samsung GE3 DVS      \\
ICCV 2021   &Yu \etal \cite{yu2021training}          &  \textbf{weakly supervised}                              & Image-based                                   & \Checkmark   & \XSolidBrush & \XSolidBrush    & ViT+CNN & GoPro, SloMo-DVS\\
CVPR 2022   & Time Replayer\cite{he2022timereplayer}  &  \textbf{unsupervised} cycle-consistent style            & 4-channel frames\cite{park2016performance}    & \Checkmark   & \XSolidBrush & \XSolidBrush      & CNN   & GoPro, Adobe240, Vimeo90k      \\
CVPR 2022   & TimeLens++\cite{tulyakov2022time}       & multi-scale feature-level fusion                        & voxel grid\cite{zihao2018unsupervised}        & \Checkmark   & \XSolidBrush & \Checkmark    & CNN  & BS-ERGB, HS-ERGB \\
ECCV 2022   & $A^2OF$\cite{wu2022video}               & optical flows adjustment                & four-channel frame\cite{park2016performance}                  & \Checkmark    & \XSolidBrush & \Checkmark     & CNN    & Adobe240, GoPro, Middlebury, HS-ERGB, HQF     \\
ECCV 2022   & $A^2OF$\cite{wu2022video}               & optical flows adjustment                & four-channel frame\cite{park2016performance}                  & \Checkmark    & \XSolidBrush & \Checkmark     & CNN    & Adobe240, GoPro, Middlebury, HS-ERGB, HQF     \\
\midrule
ECCV2020    & Lin \etal \cite{lin2020learning}        & physical model inspired                                & stream and frame-based                         & \XSolidBrush & \Checkmark & \Checkmark     & CNN   & GoPro,Blur-DVS      \\ 
CVPR 2022   & E-CIR\cite{song2022cir}                 & parametric intensity function                           & polynomial\cite{song2022cir}                  & \XSolidBrush & \Checkmark & \Checkmark     & CNN   & REDS      \\
CVPR 2022   & Zhang \etal \cite{zhang2022unifying}    & deblurring and frame interpolation                      & event streams                                 & \XSolidBrush & \Checkmark & \XSolidBrush    & CNN  & GoPro, HQF, RBE       \\
CVPR 2023  & CBMNet~\cite{Kim_2023_CVPR}  & bidirectional motion fields & voxel grids & \Checkmark & \XSolidBrush & \Checkmark    & CNN  & HQF, BS-ERGB       \\
\midrule
ECCV2020    & Lin \etal \cite{lin2020learning}        & physical model inspired                                & stream and frame-based                         & \XSolidBrush & \Checkmark & \Checkmark     & CNN   & GoPro,Blur-DVS      \\ 
CVPR 2022   & E-CIR\cite{song2022cir}                 & parametric intensity function                           & polynomial\cite{song2022cir}                  & \XSolidBrush & \Checkmark & \Checkmark     & CNN   & REDS      \\
CVPR 2022   & Zhang \etal \cite{zhang2022unifying}    & deblurring and frame interpolation                      & event streams                                 & \XSolidBrush & \Checkmark & \XSolidBrush    & CNN  & GoPro, HQF, RBE       \\
\bottomrule
\end{tabular}
}
\label{tab:vfi_comparison}
\end{table*}

\begin{figure*}[t!]
\centering
\includegraphics[width=.98\textwidth]{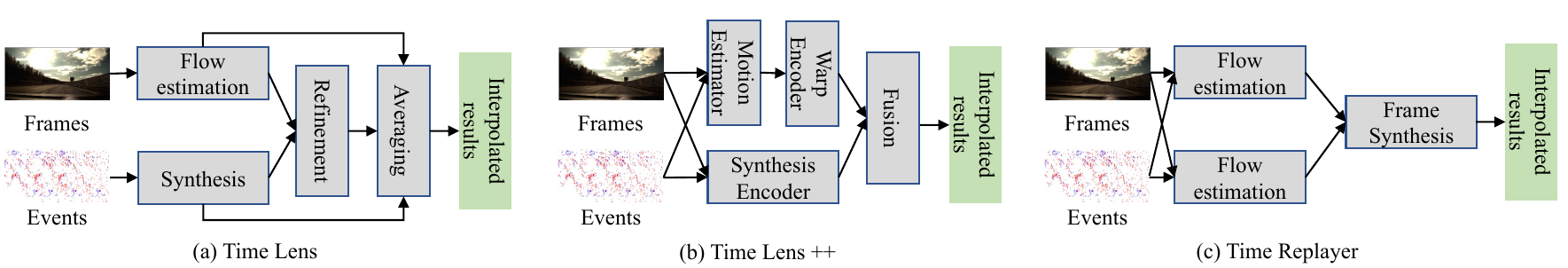}
\caption{Representative VFI methods, including, \eg, (a) TimeLens~\cite{tulyakov2021time}, the first event-guided VFI method (b) TimeLens++\cite{tulyakov2022time}, the SOTA event-based VFI method (c) TimeReplayer\cite{he2022timereplayer}, the first unsupervised event-guided VFI method.} 
\end{figure*}

For the first challenge, ~\cite{yu2020event} and ~\cite{zhu2021eventgan} are representative works leveraging generative adversarial networks (GANs) to bridge knowledge transfer between events and RGB images to alleviate the scarce labeled data problem.
Moreover, Stoffregen \etal ~\cite{stoffregen2020reducing} found that the contrast threshold is a key factor in synthesizing data to match the real event data well.
Further, Vall{\'e}s \etal ~\cite{paredes2021back} explored the theoretical basis of event cameras and proposes self-supervised learning to reduce the dependence on the ground truth video (including synthetic data) based on the photometric constancy of events.

For addressing the second challenge, 
Scheerlinck \etal ~\cite{scheerlinck2020fast} employed recurrent connections to build a state over time, allowing a much smaller recurrent neural network that reuses previously reconstructed results. Interestingly, Duwek \etal ~\cite{duwek2021image} combined CNNs with SNNs based on Laplacian prediction and Poisson integration to achieve video reconstruction with fewer parameters. 
To solve the third challenge, GANs, the double integral model~\cite{pan2019bringing}, and RNNs are applied to avoid generating blurred results and obtain high-speed and HDR videos from events in \cite{MostafaviIsfahani2018EventBasedHD,zou2021learning,rebecq2019high,zhang2022formulating}.
As for generating super-resolution (SR) images/videos from events, we divide the prevailing works into three categories, including optimization-based\cite{li2019super}, supervised \cite{duan2021eventzoom, jing2021turning, wang2020joint,liu2023sensing}, and adversarial learning methods \cite{wang2020eventsr}. 
Optimization-based methods, \eg, \cite{li2019super}, adopts a two-stage framework to solve the SR image reconstruction problem based on the non-homogeneous Poisson point process. Supervised methods either utilize residual connections to prevent the network models from the problem of gradients vanishing when generating SR images or estimate optical flow and temporal constraints to learn the motion cues, so as to reconstruct SR videos. For adversarial learning methods, Wang \etal \cite{wang2020eventsr} propose a representative end-to-end SR image reconstruction framework without access to the ground truth (GT), \ie, HR images.

\noindent\textit{\textbf{Remarks}}: 
In this section, we discuss various techniques for event-based intensity reconstruction. However, we acknowledge the need for a brief comparison to determine which technique is more suitable for different scenarios. While DL-based methods have shown significant accuracy gains in terms of accuracy and photorealism compared to early approaches relying on assumptions and regularization, they also come with increased computational complexity. On the other hand, traditional methods based on direct event integration may suffer from artifacts and produce less photo-realistic results. Furthermore, the choice of event representation remains an open question, and existing learned models often exhibit limited generalization capability. Noise in event data also poses a significant challenge, and the reconstruction of color images/videos from events is a particularly difficult problem. Future research efforts could focus on addressing these aspects to improve the quality and fidelity of reconstructed results.


\subsection{Event-guided Image/Video Super-Resolution (SR)}
\label{subsec:event+sr}
\textit{\textbf{Insight:}} The goal is to explore the visual information, \eg, edge, and high temporal resolution of events, which are fused with the low-resolution (LR) image/video to recover the high-resolution (HR) image/video, as shown in Tab. \ref{tab:sr_comparison}.

\noindent \textbf{Image SR:} eSL-Net \cite{wang2020event} is the first work that introduces events for guiding image SR. It proposes a unified event-guided sparse learning framework that simultaneously denoises, deblurs, and super-resolves the low-quality active pixel sensor (APS)  \footnote{A type of frame-based sensor, embedded in DAVIS event cameras.} images to recover high-quality images in an end-to-end learning manner. However, due to the limitations of sparse coding, this method performs poorly on more complex datasets~\cite{wang2020event,han2021evintsr}. 
EvIntSR \cite{han2021evintsr} achieves the goal of image SR in two steps: 1) Synthesizing a sequence of latent frames by combing events and blurry LR frames; 2) Merging latent frames to obtain a sharp HR frame. In general, EvIntSR explores the distinctive properties of events more directly than eSL and achieves better SR results on the simulation dataset. However, this method has two drawbacks: 1) Errors are accumulated in the two-stage training procedure; 2) The visual information of events is less explored in the second stage.

\noindent \textbf{Video SR:} Compare with image SR, video SR pays more attention to the relationship between multiple frames. 
E-VSR~\cite{jing2021turning} is the first VSR framework with events.
Similar to EvIntSR, it also consists of two sub-tasks: video frame interpolation and video SR and is limited by the accumulated errors.
Recently, EG-VSR \cite{lu2023learning} employs implicit functions for learning the continuous representation of videos. This method enables end-to-end upsampling at arbitrary scales, offering advantages in the video SR task.

\begin{table}[t!]
\centering
\caption{
Comparison of the representative event-guided image/video SR methods.
}
\begin{tabular}{llllccccc}
\toprule
Publication     & Methods                                   & Highlight             & Backbone   \\ 
\midrule
ECCV 2020       & eSL-Net \cite{wang2020event}              & sparse learning       & CNN \\
ICCV 2021       & EvIntSR \cite{han2021evintsr}             & two-step methods      & CNN \\
CVPR 2021       & E-VSR~\cite{jing2021turning}              & two-step methods      & CNN \\
CVPR 2023       & EG-VSR \cite{lu2023learning}              & SR with arbitrary scales  & ViT+CNN \\
\bottomrule
\end{tabular}
\label{tab:sr_comparison}
\end{table}

\noindent\textit{\textbf{Remarks}}: 
While significant progress has been made, including the ability to perform upsampling at arbitrary scales, there are still areas that need further investigation.
For example, the research ignores the distinct modality differences between events and RGB frames. Therefore, the directly fusing features of two modalities might degrade the performance of SR as events are often disturbed by unexpected noises, \eg, in low-light scenes  Future research could explore more to tackle these problems.

\begin{figure}[t!]
    \centering
    \includegraphics[width=1\columnwidth]{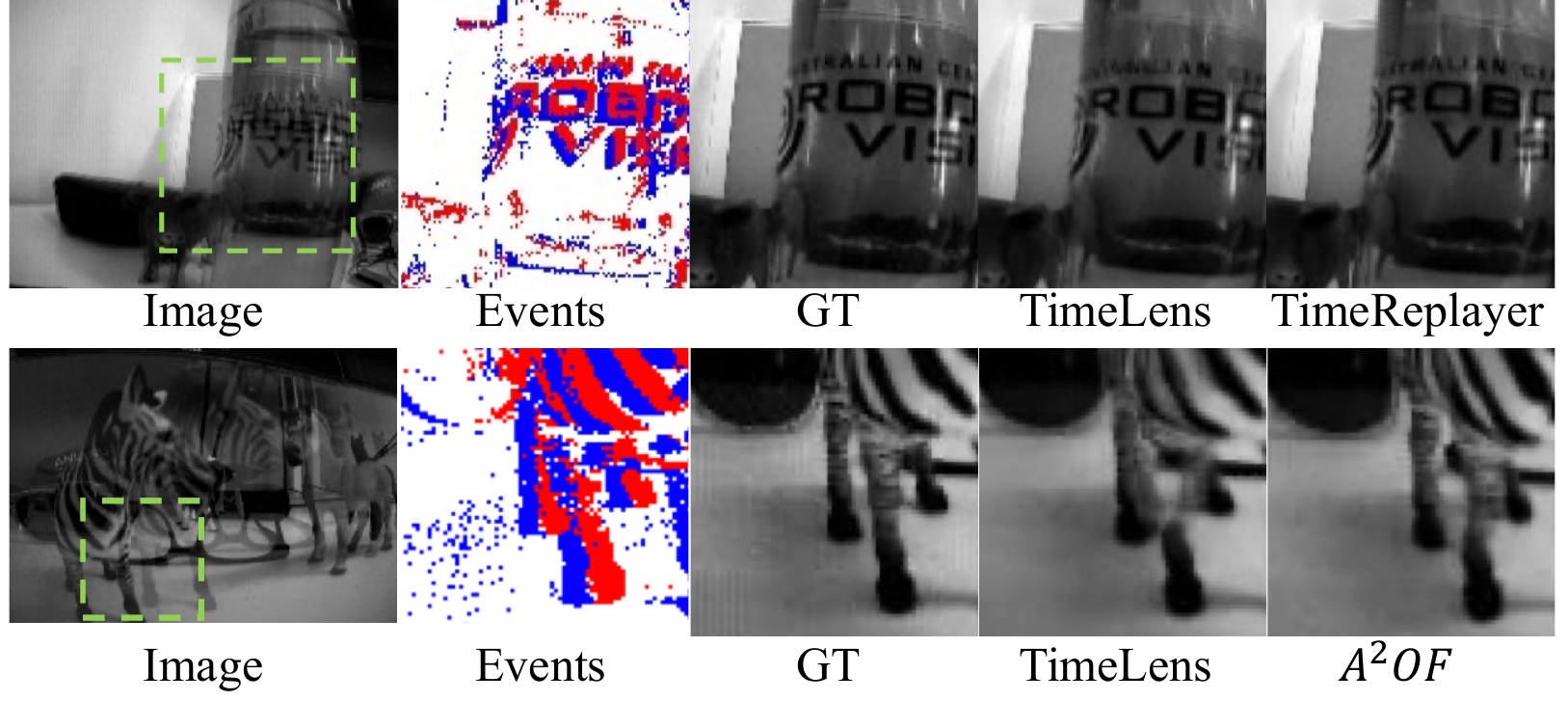}
    \caption{Visual results of VFI by three different methods. (TimeLens\cite{tulyakov2021time}, TimeReplayer\cite{he2022timereplayer}, $A^2OF$\cite{wu2022video}).}
    \label{fig:vfi-v}
\end{figure}

\begin{figure*}[t!]
\centering
\includegraphics[width=.95\textwidth]{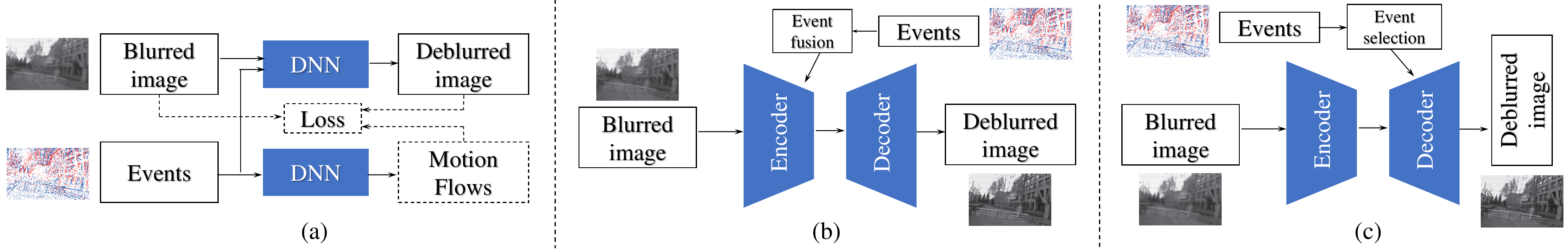}
\caption{Representative deblurring methods, including, \eg, (a) Interaction-based methods, (b) Event fusion-based methods, \eg, and (c) Event selection-based methods.} 
\label{fig:Deblurr}
\end{figure*}

\subsection{Event-guided Video Frame Interpolation (VFI)}
\label{subsec:event+vfi}

\textit{ \textbf{Insight} }: 
This task leverages the high temporal resolution of events and aims to estimate the non-linear motion information between frames, so as to insert latent frames between two consecutive frames.
Based on how the VFI frameworks are learned, we categorize them into three types: supervised, weakly supervised, and unsupervised methods, as shown in Tab.~\ref{tab:vfi_comparison}.
The VFI results of some representative methods are visualized in Fig.~\ref{fig:vfi-v}.

\noindent \textbf{Supervised methods:} 
TimeLens~\cite{tulyakov2021time} is the first and representative work, which employs four modules to fuse features to achieve warping-based and synthesis-based interpolation (See Fig.~\ref{fig:vfi}(a)). A dataset with spatially aligned events and high-speed videos was also released.
However, it has limitations in that 
1) the optical flow estimated from events limits the warped frames;
2) the noisy events restrict the quality of optical flow; and 
3) it is learned sequentially (\ie, not in an end-to-end manner). 
Therefore, training TimeLens is difficult, and errors are accumulated, degrading the performance. These problems are better addressed later on by Timelens++\cite{tulyakov2022time}, $A^2OF$~\cite{wu2022video}, and EFI-Net\cite{paikin2021efi}. 

In particular, Timelens++\cite{tulyakov2022time} proposes a framework, comprised of four modules including motion estimation, warping encoder, synthesis encoder, and fusion module, as depicted in Fig.~\ref{fig:vfi}(b). This method introduces multi-scale feature-level fusion and computes one-shot non-linear inter-frame motion, which could effectively be sampled for image warping based on events and frames. $A^2OF$~\cite{wu2022video} focuses on generating the anisotropic optical flow from events. However, such an approach cannot model the complicated motion in real-world scenes; therefore, $A^2OF$ employs the distribution masks for optical flow from events to achieve the intricate intermediate motion interpolation.

It is worth noting that the events used by the aforementioned methods have the same spatial resolution as RGB frames. Unfortunately, it is quite expensive to match a RGB sensor's resolution with an event sensor in real scenarios. Therefore, EFI-Net\cite{paikin2021efi} proposes a multi-phase CNN-based framework, which can fuse the frames and events with various spatial resolutions. In summary, supervised methods rely on paired data with high-frame-rate videos and events. HS-ERGB \cite{tulyakov2021time} and BS-ERGB\cite{tulyakov2022time} are representative datasets. However, these datasets suffer from strict pixel alignments between events and frames and are expensive to collect.
Therefore, some weakly-supervised and unsupervised methods have been proposed recently.

\noindent \textbf{Weakly-supervised methods:} 
Yu \etal \cite{yu2021training} proposed the first weakly-supervised event-based VFI method. In practice, it extracts complementary information from events to correct image appearance and employs an attention mechanism to support correspondence searching on the low-resolution feature maps.
Meanwhile, a real-world dataset, namely SloMo-DVS, is also released.

\noindent \textbf{Unsupervised methods:}
TimeReplayer~\cite{he2022timereplayer} is the first unsupervised method, trained in a cycle-consistent manner, as shown in Fig.~\ref{fig:vfi} (c).
It directly estimates the optical flow between the key-frame and the input frame, instead of computing intermediate frames as a proportion of the computed optical flow between input frames.
In this way, the complex motion can be estimated.
Then, input frames can be reconstructed by key-frame and inverse optical flow. 
Totally, this cycle consistency method not only models complex nonlinear motion but also avoids the need for a large amount of paired high-speed frames and events.

All the above-mentioned methods have the strong assumption that the exposure time of RGB frames is very short, and there are no blur artifacts in frames.
However, this assumption is overly harsh because the practical exposure time can be long and results in blur artifacts in frames, particularly in complex lighting scenes.
When the exposure time is longer, the issue of interpolation needs to be re-examined.
For this reason, \cite{lin2020learning,zhang2022unifying,song2022cir} jointly address the interpolation problem and deblurring. 
For example, E-CIR \cite{song2022cir} transforms a blurry image into a sharp video that is represented as a time-to-intensity parametric function with events.
Similarly, Zhang \etal~\cite{zhang2022unifying} employed a learnable double integral network to map blurry frames to sharp latent images with event guidance.
Lin \etal~\cite{lin2020learning} emphasized that the residuals between a blurry image and a sharp image are event integrals.
Based on this perspective, they proposed a network that uses events to estimate residuals for sharp frame restoration.

\noindent\textbf{\textit{Remarks}:} From our review, the majority of the methods are based on supervised learning, and weakly supervised or unsupervised methods still have a lot of room for further research. For example, mutual supervision could be performed through the imaging relationship between events and interpolated frames to relieve the need for ground truth, \ie, high frame rate video for training.
Also, dense optical flow estimated by events\cite{gehrig2021raft}, could be used as a constraint between interpolation results to improve VFI accuracy in an unsupervised manner.

\subsection{Event-guided Image/Video Deblurring}
\textit{\textbf{Insight}}: This task gets inspired by the no-motion-blur property of events and aims to restore a sharp image/video from a blurry image/video sequence under the guidance of events.
Because supervised methods tend to achieve higher PSNR and SSIM~\cite{he2022timereplayer,tulyakov2021time}.
Traditional deblurring methods rely on the physical event generation model~\cite{wang2020joint}. In particular,  Pan \etal~\cite{pan2019bringing} proposed an event-based double integral model for recovering latent intensity images.  Based on this model, sharp images and videos could be generated by solving the non-convex optimization problems under adverse visual conditions. However, it suffers from the problem of accumulated error caused by noise in the sampling process. By contrast, learning-based methods directly explore the relationship between blurry and sharp images with the help of events and show more plausible deblurring results. In this paper, we divide the learning-based methods into three categories: 1) interaction-based methods; 2) fusion-based methods; and 3) selection-based methods (See Fig.~\ref{fig:Deblurr}). The deblur results of some representative methods are shown in Tab.~\ref{Deblurring}.

\begin{table}[t!]
\centering
\caption{Qualitative comparison of deblurring methods on GoPro and HQF dataset from ~\cite{zhang2022unifying}. `N/A' means no results are available. Param. refers to the number of parameters in the framework.}
\resizebox{0.94\linewidth}{!}{
\begin{tabular}{ccccccccc}
\toprule
\multirow{2}{*}{Method} & \multicolumn{3}{c}{GoPro} & \multicolumn{3}{c}{HQF} & \multirow{2}{*}{Param.}\\ \cmidrule{2-7}
& PSNR$\uparrow$ & SSIM $\uparrow$ & LPIPS $\downarrow$ & PSNR $\uparrow$ & SSIM $\uparrow$& LPIPS$\downarrow$ \\ \midrule
LEVS ~\cite{jin2018learning} & 20.84 & 0.5473 & 0.1111 & 20.08 & 0.5629 & 0.0998 & 18.21M \\
EDI ~\cite{pan2019bringing} & 21.29 & 0.6402 & 0.1104 & 19.65 & 0.5909 & 0.1173 &  N/A \\
eSL-Net ~\cite{wang2020event} & 17.80 & 0.5655 & 0.1141 & 21.36 & 0.6659 & 0.0644 &  0.188M \\
LEDVDI ~\cite{lin2020learning} &  25.38 & 0.8567 & 0.0280 & 22.58 & 0.7472 & 0.0578 & 4.996M \\
RED~\cite{xu2021motion} & 25.14 & 0.8587 & 0.0425 & 24.48 & 0.7572 & 0.0475 & 9.762M \\
EVDI~\cite{zhang2022unifying} & 30.40 &  0.9058 &  0.0144 &  24.77 &  0.7664 &  0.0423 & 0.393M \\
\bottomrule
\end{tabular}}
\label{Deblurring}
\end{table}


\noindent \textbf{Interaction-based methods} usually input the blurry image and events into two different networks and then carry out information interaction after encoding the features in each branch to improve the deblurring effect (See Fig.~\ref{fig:Deblurr} (a)). For example, in \cite{xu2021motion}, a self-supervised framework was proposed to reduce the domain gap between simulated and real-world data. Specifically, they first estimated the optical flow and exploited the blurry and photometric consistency to enable self-supervision on the deblurring network. Lin \etal  
\cite{lin2020learning} introduced a CNN framework to predict the residual between sharp and blurry images for deblurring, and the residual between sharp frames for interpolation. Jiang \etal~\cite{jiang2020learning} explored long-term, local appearance/motion cues and novel event boundary priors to solve motion deblurring. Zhang \etal~\cite{zhang2022unifying} utilized low latency of events to alleviate motion blur and facilitate the prediction of intermediate frames.

\noindent \textbf{Fusion-based methods} aims to design a principled framework for video and event-guided deblurring~\cite{wang2020event,zhang2023event}, as shown in Fig.~\ref{fig:Deblurr} (b). For instance, Shang \etal~\cite{shang2021bringing} proposed a two-stream framework to explore the non-consecutively blurry frames and bridge the gap between event-guided and video deblurring. 

\noindent \textbf{Selection-based methods}, \eg, \cite{kim2022event}, formulate the event-guided motion deblurring by considering the unknown exposure and readout time in the video frame acquisition process. The main challenge is how to selectively use event features by estimating the cross-modal correlation between the blurry frame features and the events. Therefore, the proposed event selection module subtly selects useful events, and the fusion module fuses the selected event features and blur frames effectively, as shown in Fig.~\ref{fig:Deblurr} (c).

\noindent\textit{\textbf{Remarks}}: 
Most of the aforementioned deblurring methods are limited to some specific scenes. 
In some scenes with large or fast motions, the model's accuracy may deteriorate dramatically.

\label{subsec:event+deblurring}

\begin{table*}[]
\centering
\caption{Experiments of representative methods on event object classification. N/A means no results available. Param. refers to the number of parameters in the framework.}
\label{Tab:Classification_exp}
\resizebox{0.99\textwidth}{!}{
\begin{tabular}{llcccccccc}
\toprule
\multirow{2}{*}{Publication} & \multirow{2}{*}{Methods} & \multicolumn{6}{c}{Dataset}                                           &   &\multirow{2}{*}{Param.}         \\ \cmidrule{3-9}&                            
& N-MINIST~\cite{orchard2015converting} & MINIST-DVS~\cite{mnistdvs} & N-Caltech101~\cite{orchard2015converting} & CIFAR10-DVS~\cite{cifar10-dvs} & N-Cars~\cite{sironi2018hats} & ASL-DVS~\cite{ASL-DVS} & N-ImageNet~\cite{N-ImageNet} \\ \midrule
TPAMI 2015 & HFirst~\cite{Hfirst} &0.712 & N/A &0.054 & N/A &0.561 & N/A & N/A & 21.79M\\ 
TPAMI 2016 & HOTS~\cite{Hots} &0.808 &0.803 & 0.210 & 0.271 & 0.624 & N/A & N/A & 21.79M\\ 
CVPR 2018 & HATS~\cite{sironi2018hats} &0.991 &0.984 &0.642 &0.524 & 0.902 & N/A & 0.471 & 21.79M \\ 
ICCV 2019 & EST~\cite{EST} & N/A & N/A & 0.817 & N/A & 0.925 & N/A &  \textbf{0.489} & 21.79M \\ 
ICCV 2019 & RG-CNNs~\cite{RGCNNs} &0.990 &0.986 &0.657 &0.540 &0.914  &0.901 & N/A & 19.46M\\ 
TPAMI 2019 & DART~\cite{ramesh2019dart} & 0.979 & 0.985 & 0.664 & 0.658 & N/A & N/A  & N/A  & N/A \\ 
ECCV 2020 & Matrix-LSTM~\cite{Matrixlstm} &0.989 & N/A &0.843 &N/A &0.943 &  \textbf{0.997} &0.322 &25.56M \\ 
ECCV 2020 & ASCN~\cite{messikommer2020event} & N/A & N/A & 0.745 & N/A & 0.944  & N/A & N/A & 9.47M  \\ 
ICCV 2021 & EvS~~\cite{EvS} & N/A &  \textbf{0.991} & 0.761 &  0.680 & 0.931 & N/A & N/A & N/A\\ 
ICCV 2021 & DiST~\cite{N-ImageNet} & N/A & N/A & N/A & N/A & N/A & N/A & 0.484   & 21.79M   \\ 
TCSVT 2021 & MVF-Net~\cite{deng2021mvf} & 0.993 & N/A &  \textbf{0.871} & 0.663 & 0.968  & 0.996 & N/A  & 21.79M \\ 
CVPR 2022 & AEGNNs~\cite{schaefer2022aegnn} & N/A & N/A & 0.668 & N/A & 0.945  & N/A & N/A & N/A\\ 
CVPR 2022 & EV-VGCNN~\cite{Ev-VGCNN} &  \textbf{0.994} & N/A & 0.748 & N/A & 0.953  & 0.983 & N/A & 21.79M  \\ 
TPAMI 2022 & TORE~\cite{baldwin2022time} & 0.994 &N/A & 0.798 &N/A & \textbf{0.977} & 0.996 & N/A & 5.94M \\ 
ICCV 2023 & GET~\cite{peng2023get} &  \textbf{0.997} &N/A & N/A & \textbf{0.848} & 0.967 & N/A & N/A & 4.50M\\ 
\bottomrule
\end{tabular}}
\end{table*}

\subsection{Event-based Deep Image/Video HDR}
\label{subsec:event+hdr}

{\textit{\textbf{Insight}}: The HDR of events makes it naturally more advantageous to reconstruct an HDR image/video. The predominant methods can be divided into two main categories: event-based HDR image/video HDR methods~\cite{rebecq2019high,zou2021learning,MostafaviIsfahani2018EventBasedHD} and event-guided image/video HDR methods (a hybrid of event and frame data)~\cite{Han2020NeuromorphicCG,messikommer2022multi}.}

\begin{figure}[t!]
\centering
\includegraphics[width=0.48\textwidth]{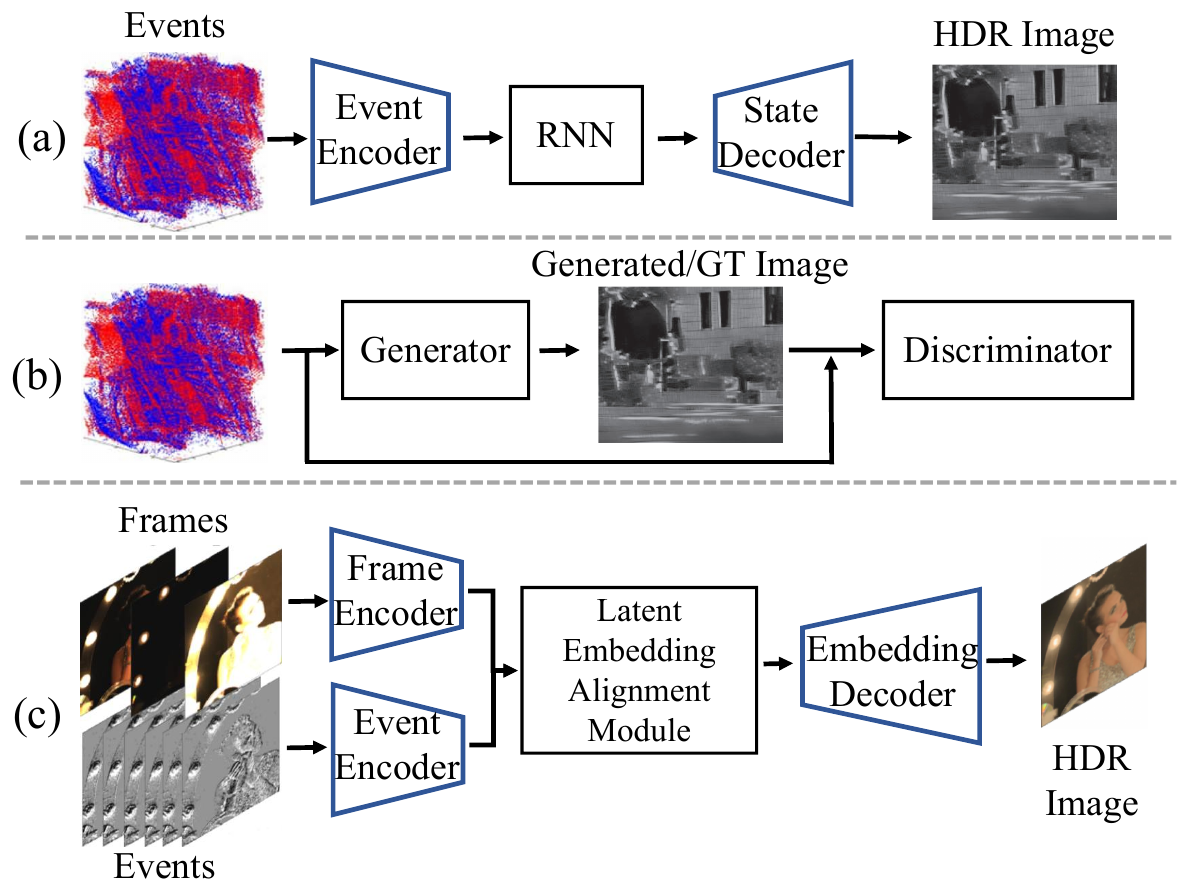} 
\caption{Representative DL-based HDR imaging methods. (a) RNN-based methods~\cite{rebecq2019high,zou2021learning} and (b) GAN-based method~\cite{MostafaviIsfahani2018EventBasedHD} that use events only, and (c) event-frame fusion~\cite{Han2020NeuromorphicCG, messikommer2022multi}. (\textit{latent embedding} denotes information learned by feature extractor, filter,\etc.) }
\label{fig:HDR}
\end{figure}

\noindent \textbf{Event-based image/video HDR} typically employs the idea of event-to-image translation\textemdash reconstructing HDR images from events, as mentioned in Sec.~\ref{event-to-image}. Representative works are based on the recurrent neural networks (RNNs)~\cite{rebecq2019high, zou2021learning} (See Fig.~\ref{fig:HDR} (a)) or generative adversarial networks (GANs)~\cite{MostafaviIsfahani2018EventBasedHD} (See Fig.~\ref{fig:HDR} (b)). However, the reconstructed HDR results intrinsically lack textural details, especially in the static scene, as events are sparse and motion-dependent.

\noindent \textbf{Event-guided image/video HDR:} HDR imaging methods are categorized into two types: single-exposure HDR and multi-exposure HDR (See \cite{wang2021deep} for details). Event-guided image/video also follows these two paradigms. ~\cite{Han2020NeuromorphicCG, messikommer2022multi} explore the potential of merging both events and frames for this task, as shown in Fig.~\ref{fig:HDR} (c). In particular, Han \etal ~\cite{Han2020NeuromorphicCG} proposed the first single-exposure HDR imaging framework to recover an HR HDR image by adding the LR intensity map generated from events. This framework addresses the gaps in spatial resolution, dynamic range, and color representation of the hybrid sensor system to pursue a better fusion.
By contrast, EHDR ~\cite{messikommer2022multi} is the first multi-exposure HDR imaging framework that combines bracketed LDR images and synchronized events to recover an HDR image. To alleviate the impact of scene motion between exposures, 
EHDR employs events to learn a deformable convolution kernel, which can align feature maps from images with different exposure times. By contrast, HDRev-Net~\cite{yang2023learning} implicitly mitigates the misalignment of multi-modal representations by aligning them in the shared latent space and fusing them with a confidence-guided fusion module.

\noindent\textit{\textbf{Remarks}}: Based on the review, only two research works have been proposed for deep HDR imaging. 
The most possible reason is that it is practically difficult to collect paired datasets for training, especially for multi-exposure HDR imaging. Future directions could consider directly fusing LDR images and events and learning a unified HDR imaging framework without relying on intensity reconstruction. Also, it is promising to explore how to leverage events to guide color image HDR imaging.  

\section{Scene Understanding and 3D Vision}
\label{sec:Scene Understanding and 3D Vision}
\subsection{Scene Understanding}
\subsubsection{Object Classification}
\textit{\textbf{Insight}}: Event-based object classification aims to identify and classify objects from an event stream based on their visual characteristics. 
This allows for real-time object classification with high temporal resolution and low latency, making it suitable for applications in robotics, autonomous vehicles, and other mobile systems.
Intuitively, we divide the event-based classification methods into three categories according to the input event representations and DNN types: 1) learning-based; 2) graph-based; and 3) asynchronous model-based methods.

\noindent \textbf{Learning-based methods}
Gehrig \etal~\cite{EST} proposed the first end-to-end framework to learn event representation for object classification. In particular, it converts event streams into grid-like tensors, \ie, Event Spike Tensor (EST), through a sequence of differentiable operations.
Though EST achieves high accuracy, it also brings redundant computation costs and high latency.
To tackle this problem, Cannici \etal~\cite{Matrixlstm} proposed Matrix-LSTM to adaptively integrate and utilize information of events by the memory mechanism of LSTM. This makes it possible to efficiently aggregate the temporal information of event data. 

\noindent \textbf{Graph-based methods}
Some works also utilize graphs for representing events for the computational efficiency of the Graph CNNs. Yin \etal~\cite{ASL-DVS} proposed a representative approach that represents event data as a graph and introduced residual graph CNNs (RG-CNNs).

\noindent \textbf{Asynchronous model-based methods}
Though the learning-based methods obtain plausible classification results, they fail to fully explore the inherent asynchronicity and sparsity of event data.
Consequently, Nico \etal ~\cite{messikommer2020event} converted the classification models trained on the synchronous frame-like event representations into models taking asynchronous events as inputs.

\begin{table}[t!]
\centering
\caption{Comparison of existing representative event classification benchmarks. MR denotes Monitor Recording. MR is the process of capturing the visual output displayed on a computer monitor or screen. }
\label{Tab:clsdataset}
\resizebox{0.49\textwidth}{!}{
\begin{tabular}{ccccc}
\toprule
Dataset & \# of Samples & \# of Classes &Sources & Paired RGB Data \\ \midrule
N-Cars~\cite{sironi2018hats} & 24029 & 2 & Real & N/A \\
N-Caltech101~\cite{orchard2015converting} & 8709 & 101 & MR & Caltech101 \\
CIFAR10-DVS~\cite{cifar10-dvs} & 10000& 10 & MR & CIFAR10 \\
ASL-DVS~\cite{ASL-DVS}& 100800 & 24 & Real & N/A \\
N-MNIST~\cite{orchard2015converting} & 70000 & 10 & MR & MNIST \\
MNIST-DVS~\cite{mnistdvs} & 30000 & 10 & MR & MNIST \\
N-ImageNet~\cite{N-ImageNet} & 1781167 & 1000 & MR & ImageNet \\ \bottomrule
\end{tabular}}
\end{table}

Besides, to fit with the sparse event data, VMV-GCN\cite{xie2022vmv} first considers the relationships between vertices of the graph and then groups the vertices according to the proximity both in the original input and feature space. 
Furthermore, for computational efficiency, AEGNN~\cite{schaefer2022aegnn} proposes to process events sparsely and asynchronously as temporally evolving graphs. Meanwhile, EV-VGCNN\cite{Ev-VGCNN} utilizes voxel-wise vertices rather than point-wise inputs to explicitly exploit the regional 2D semantics of event streams while maintaining the trade-off between accuracy and model complexity.

\begin{table*}[t!]
\centering
\caption{Comparison of existing representative event object detection methods.}
\label{Tab:object_detection}
\resizebox{0.99\textwidth}{!}{
\begin{tabular}{lllllcc}
\toprule
Publications & Method            & Representations & Highlight                                     & Backbone           & Frame Images & Multi-modal \\ \midrule
WACV 2022    & PointConv~\cite{spacedetection} & Image-based           & pint-cloud feature extractor                                 & CNN                & \XSolidBrush                 & \XSolidBrush         \\ 
MFI 2022     & GFA-Net~\cite{eventkitti}           & Image-based           & Edge information \& Temporal information across event frames & CNN \& ViT & \XSolidBrush                 & \XSolidBrush           \\ \midrule
ECCV 2020    & NGA~\cite{perot2020learning}               & Image-based           & Grafting pre-trained deep network for novel sensors          & CNN                & \XSolidBrush                 & \Checkmark           \\ 
NeurIPS 2020 & RED~\cite{perot2020learning}               & Image-based           & Recurrent architecture and temporal consistency              & RNN                & \XSolidBrush                 & \XSolidBrush           \\ 
TIP 2022     & ASTMNet~\cite{spatiotemporalmemory}           & Image-based           & Continuous event stream with lightweight RNN                 & RNN                & \XSolidBrush                 & \XSolidBrush           \\ \midrule
ICRA 2019    & Mixed-Yolo~\cite{mixedframeevent}        & Image-based           & Mixed APS frame and DVS frame                                & CNN                & \Checkmark                  & \XSolidBrush           \\ 
ICME 2019    & JDF~\cite{jointdetection}               & Spike-based           & Joint detection with event streams and frames                & CNN \& SNN         & \Checkmark                  & \XSolidBrush           \\ 
ICRA 2022    & FPN-fusion events~\cite{fusingdetection} & Image-based           & Robust detection with RGB and event-based sensors            & CNN                & \XSolidBrush                 & \Checkmark            \\ \bottomrule
\end{tabular}}
\end{table*}

\noindent\textbf{Benchmark datasets} are vital foundations for the development of event-based vision, given that sufficient event data is barely available due to the novelty of event sensors. 
The existing event datasets can be briefly divided into two categories according to the captured scenes, \ie, the real and the simulated ones.
Gehrig \etal ~\cite{videotoevent} proposed to convert video datasets into event datasets by adaptive upsampling and using an event camera simulator (ESIM)~\cite{rebecq2018esim}. Models trained on the simulated dataset generalize well on the real data. 
More recently, N-ImageNet \cite{N-ImageNet} serves as the first real large-scale fine-grained benchmark, which provides various validation sets to test the robustness of event-based object recognition approaches amidst changes in motion or illumination. 
We summarize the existing datasets for event recognition in Tab.~\ref{Tab:clsdataset} and conduct a benchmark evaluation for the representative event-based classification methods in Tab.~\ref{Tab:Classification_exp}.

\noindent\textit{\textbf{Remarks}}: The efficacy of event-based object classification is often constrained by the paucity of annotated datasets. As a result, significant efforts have been directed toward synthesizing event data through tools like ESIM or generating event data from images displayed on monitors. There is also a noteworthy potential in adapting models, initially trained on synthetic data, for application to real-world event data~\cite{kim2022evtta, planamente2021da4event}. Furthermore, an emerging area of interest involves exploiting large volumes of unlabeled data through techniques such as active learning, wherein the classifier is designed to request further labeled examples as necessary to enhance its performance. Additionally, there have been recent forays into knowledge transfer from extensively trained models in the RGB domain~\cite{zhou2023clip}.

\subsubsection{Feature Tracking}
\textit{\textbf{Insight}}:
In recent years, researchers have focused on event-based feature tracking for its robustness in fast motion capture and extreme lighting conditions ~\cite{messikommer2023data, gehrig2020eklt}.
Early event-based feature trackers treat events as point sets and use Iterative Closest Point (ICP) to track features
, 
and there are also works use B-splines~\cite{chui2021event} and some other techniques to obtain the feature trajectories~\cite{hu2022ecdt}.

Recently, deep learning-based method, \aka, data-driven methods for event-based feature tracking method has been proposed the performance comparison is shown in Table.~\ref{Tab:featuretracking}. The most representative one is ~\cite{messikommer2023data} which serves as the first work of introducing a data-driven feature tracker for event cameras, leveraging low-latency events to track features detected in a grayscale frame. The data-driven tracker outperforms the existing non-DL-based methods in relative feature age by up to 120\% while keeping the lowest latency.



\noindent\textit{\textbf{Remarks}}: 
From our review, we find that deep learning is just introduced to event-based feature tracking recently and it is worth exploring this direction.

\begin{table}[]
\caption{The quantitative results of the evaluated trackers on the EDS~\cite{hidalgo2022event} and EC~\cite{mueggler2017event} dataset are reported in terms of "Feature Aeg (FA)" of the stable tracks and the "Expected FA", which is the multiplication of the feature age by the ratio of the number of table tracks over the number of initial features. This table is from ~\cite{messikommer2023data}. }
\label{Tab:featuretracking}
\setlength{\tabcolsep}{1.5mm}
\resizebox{\columnwidth}{!}{
\begin{tabular}{ccccc}
\toprule
\multirow{2}{*}{Method} & \multicolumn{2}{c}{EDS~\cite{hidalgo2022event}} & \multicolumn{2}{c}{EC~\cite{mueggler2017event}} \\ \cmidrule{2-5} 
 & Feature Age (FA) & Expected FA & Feature Age (FA) & Expected FA \\ \midrule
HASTE~\cite{alzugaray2020haste} & 0.096 & 0.063 & 0.442 & 0.427 \\ \midrule
EKLT~\cite{gehrig2020eklt} & 0.325 & 0.205 & 0.811 & 0.775 \\ \midrule
DDFT~\cite{messikommer2023data} (zero-shot) & 0.549 & 0.451 & 0.811 & 0.787 \\ \midrule
DDFT~\cite{messikommer2023data} (fine-tuned) &  \textbf{0.576} &  \textbf{0.472} & \textbf{ 0.825} &  \textbf{0.818} \\ \bottomrule
\end{tabular}}
\end{table}

\subsubsection{Object Detection and Tracking}
\noindent \textbf{Object Detection}: Event cameras bring a new perspective in dealing with the challenges in object detection~\cite{cao2023chasing} (\eg, motion blur, occlusions, and extreme lighting conditions).
In reality, the RGB-based detection fails to enable robust perception under image corruptions or extreme weather conditions. Meanwhile, auxiliary sensors, such as LiDARs, are extremely bulky and costly~\cite{fusingdetection}. Therefore, event-based detectors are introduced to overcome the dilemma, especially in challenging visual conditions~\cite{eventkitti}. 
In this work, we divide the event-based object detection methods into three categories according to the input data formats and data representations, as summarized in Tab.~\ref{Tab:object_detection}. 

The first category simply converts the raw event data into frame-based images~\cite{eventkitti, gehrig2023recurrent}, \eg, the recurrent vision transformer (RVT)~\cite{gehrig2023recurrent} which takes 2-channel frames within a time duration. However, this kind of method loses the raw spatial-temporal information in the event stream.
For this reason, event volume and some other formats tailored for object detectors are used in the methods of the second category. Some works~\cite{multiplemodalities, perot2020learning} obtain event volumes by taking the linear or convolve kernels to integrate the asynchronous events into multiple slices within the equal temporal volume. 

However, the event volume still follows a frame-like 2D representation, and critical temporal information is lost. Recently, ASTMNet ~\cite{spatiotemporalmemory} exploits the spatial-temporal information by directly processing the asynchronous events instead of the 2D frame-like representations. Furthermore, it also serves as the first end-to-end pipeline for continuous object detection. Notably, a branch of research has introduced temporal hints by integrating recurrent neural network layers, resulting in significant enhancements in detection accuracy ~\cite{gehrig2023recurrent,spatiotemporalmemory}. 
The third category of attempts combines the advantages of event images and RGB images~\cite{mixedframeevent, jointdetection}. Tomy \etal ~\cite{fusingdetection} proposed a representative framework that fuses the information from the event- and frame-based cameras for better detection accuracy in normal conditions and robust performance in the presence of extreme scenarios. 

\noindent \textbf{Object Tracking}: 
Tracking dynamic objects is an essential task in mobile robots, which requires the basic functionality of obstacle avoidance. 
RGB camera-based trackers perform poorly for high-speed and dynamic objects because of motion blur and time-delayed transmission. In such cases, introducing event cameras to address this problem is of great value. 

For mobile robots, earlier methods proposed to track moving objects on the conditions of geometric priors , known shape , and motion-compensation model \cite{2019Event}, \cite{2020Event}. More recently, many DL-based methods designed for canonical image data have undergone a paradigm shift and have been applied successfully to event data. 
Many more endeavors have been made to enable the onboard inference ability of the deep learning models. 
EV-Catcher~\cite{2022EV-Catcher} trains a small CNN to process single-channel event images, achieving an inference speed of 2ms, which is faster than its predecessor by a large margin \cite{2020Fast}. Because EV-Catcher only regresses the real-time target position and its uncertainty at x-coordinate from the DNNs, further estimation of hitting position and timing is based on the linear motion assumptions. Others tend to proceed with this problem in an end-to-end manner, showing marginal benefits.

\begin{table*}[t!]
\centering
\caption{Experiments of representative methods on event-based optical flow estimation from ~\cite{DECIFlow}. \\ 
UL: unsupervised learning. SL: supervised learning. MB: model-based methods. ($\cdot$): evaluation on both \textit{outdoor\_day1} and \textit{outdoor\_day2} sequences. 
[$\cdot$]: evaluation on \textit{outdoor\_day2} sequences. N/A means no results are available.
The results without any brackets mean that they are not trained on any sequence of MVSEC. Param. refers to the number of parameters in the framework.)
}
\label{tab:optical_flow}
\setlength{\tabcolsep}{4mm}
\resizebox{0.99\textwidth}{!}{
\begin{tabular}{cccccccccccc}
\toprule
\multirow{2}{*}{\begin{tabular}[c]{@{}c@{}}Type\end{tabular}} & \multirow{2}{*}{Method} & \multirow{2}{*}{Metric} & \multicolumn{2}{c}{indoor\_flying1} & \multicolumn{2}{c}{indoor\_flying2} & \multicolumn{2}{c}{indoor\_flying3} & \multicolumn{2}{c}{outdoor\_day1} & \multirow{2}{*}{Param.}\\ \cmidrule{4-11} 
 &  &  & EPE & \%Out & EPE & \%Out & EPE & \%Out & EPE & \%Out \\ \midrule
\multirow{9}{*}{UL} & Ev-FlowNet~\cite{zhu2018ev} & sparse & (\ApplyGradient{1.03}) & (2.2) & (1.72) & (15.1) & (1.53) & (11.9) & {[}0.49{]} & {[}0.2{]} & N/A \\  \cmidrule{2-12}
 & Zhu \etal ~\cite{zhu2019unsupervised} & sparse & (0.58) & (0.0) & (1.02) & (4.0) & (0.87) & (3.0) & {[}0.32{]} & {[}0.0{]} & N/A \\ \cmidrule{2-12}
 & Matrix-LSTM~\cite{Matrixlstm} & sparse & (0.82) & (0.53) & (1.19) & (5.59) & (1.08) & (4.81) & N/A & N/A & N/A\\ \cmidrule{2-12}
 & Spike-FLowNet~\cite{lee2020spike} & sparse & {[}0.84{]} & N/A & {[}1.28{]} & N/A & {[}1.11{]} & N/A & {[}0.49{]} & N/A & 13.039M\\ \cmidrule{2-12}
 & Paredes \etal ~\cite{paredes2021back} & sparse & (0.79) & (1.2) & (1.40) & (10.9) & (1.18) & (7.4) & {[}0.92{]} & {[}5.4{]} & N/A\\ \cmidrule{2-12}
 & LIF-EV-FlowNet~\cite{LIF-EV-FlowNet} & sparse & 0.71 & 1.41 & 1.44 & 12.75 & 1.16 & 9.11 & 0.53 & 0.33 & N/A\\ \cmidrule{2-12}
 & Deng \etal~\cite{deng2021} & sparse & (0.89) & (0.66) & (1.31) & (6.44) & (1.13) & (3.53) & N/A & N/A & N/A\\ \cmidrule{2-12}
 & Li \etal~\cite{Li2020}  & sparse & (0.59) & (0.83) & (0.64) & (2.26) & N/A & N/A & {[}0.31{]} & {[}0.03{]} & N/A\\\cmidrule{2-12}
 & STE-FlowNet~\cite{STE-FlowNet} & sparse & {[}0.57{]} & {[}0.1{]} & {[}0.79{]} & {[}1.6{]} & {[}0.72{]} & {[}1.3{]} & {[}0.42{]} & {[}0.0{]} & N/A\\ \midrule
\multirow{4}{*}{SL} & Stoffregen \etal ~\cite{stoffregen2020reducing}&  dense &  \textbf{0.56} & 1.00 & 0.66 & 1.00 & 0.59 & 1.00 & 0.68 & 0.99 & N/A\\ \cmidrule{2-12}
 & EST~\cite{EST} & sparse & (0.97) & (0.91) & (1.38) & (8.20) & (1.43) & (6.47) & N/A & N/A & N/A\\ \cmidrule{2-12}
 & {DCEIFlow~\cite{DECIFlow}} & \multicolumn{1}{c}{dense} & \multicolumn{1}{c}{ \textbf{0.56}} & \multicolumn{1}{c}{0.28} & \multicolumn{1}{c}{0.64} & \multicolumn{1}{c}{ \textbf{0.16}} & \multicolumn{1}{c}{ \textbf{0.57}} & \multicolumn{1}{c}{ \textbf{0.12}} & \multicolumn{1}{c}{0.91} & \multicolumn{1}{c}{0.71} & \multicolumn{1}{c}{N/A}\\ \cmidrule{2-12}
 & {DCEIFlow~\cite{DECIFlow}} & \multicolumn{1}{c}{sparse} & \multicolumn{1}{c}{0.57} & \multicolumn{1}{l}{0.30} & \multicolumn{1}{c}{0.70} & \multicolumn{1}{c}{0.30} & \multicolumn{1}{c}{0.58} & \multicolumn{1}{c}{0.15} & \multicolumn{1}{c}{0.74} & \multicolumn{1}{c}{0.29} & \multicolumn{1}{c}{N/A}\\ \midrule 
\multirow{3}{*}{MB} 
 & Shiba~\cite{shiba2022secrets} & sparse & 0.42 &  \textbf{0.10} &  \textbf{0.60} & 0.59 & 0.50 & 0.28 &  \textbf{0.30} &  \textbf{0.10} & N/A\\\cmidrule{2-12}
 & $\text{Fusion-FlowNet}_{Early}$~\cite{Fusion-FlowNet} & dense & ( \textbf{0.56}) & N/A & (0.95) & N/A & (0.76) & N/A & {[}0.59{]} & N/A & 12.269M\\ \cmidrule{2-12}
 & $\text{Fusion-FlowNet}_{Late}$~\cite{Fusion-FlowNet} & sparse & (0.57) & N/A & (0.99) & N/A & (0.79) & N/A & {[}0.55{]} & N/A &   7.549M\\ \bottomrule
\end{tabular}
}
\end{table*}
\begin{figure}[t!]
    \centering
    \includegraphics[width=0.49\textwidth]{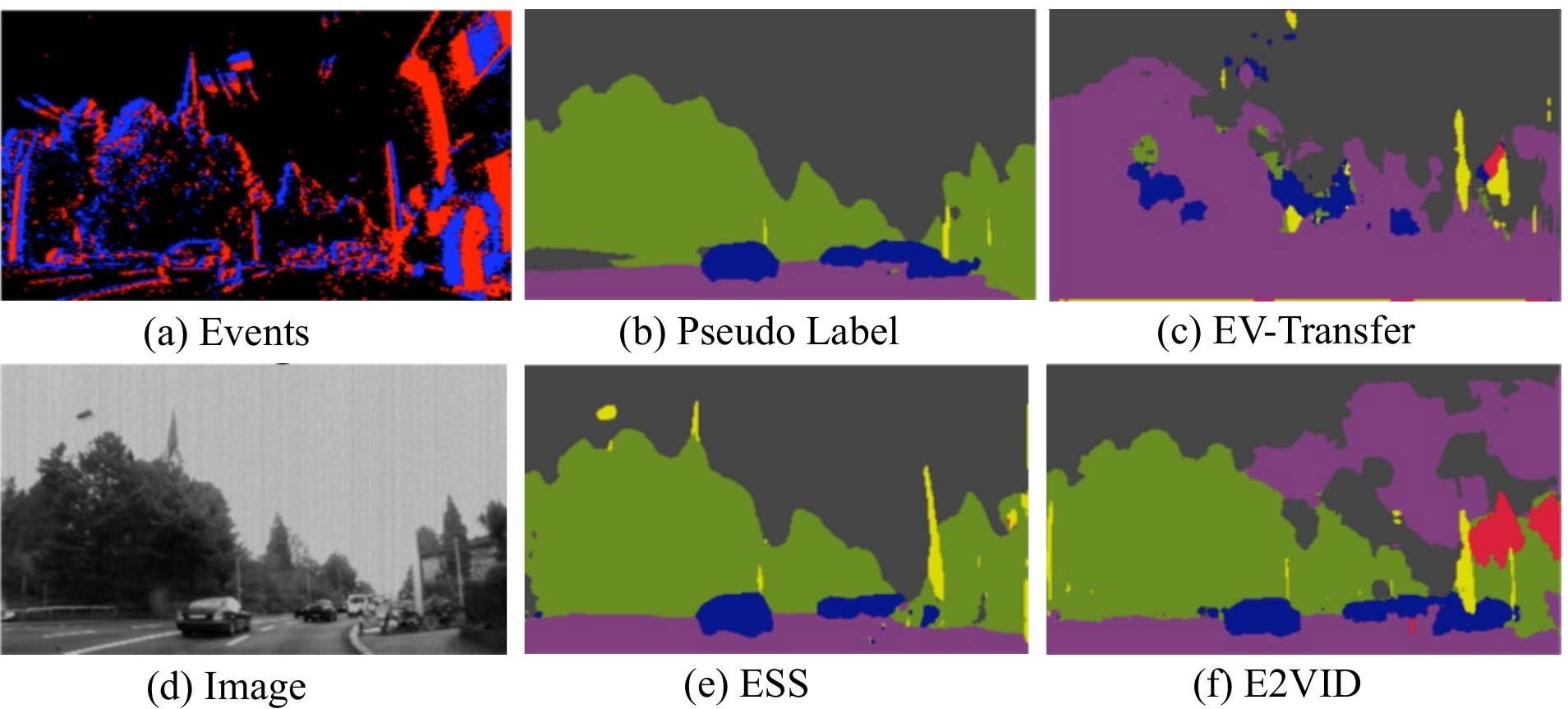}
    \caption{Visualization results of semantic segmentation with events, (a) Events, (b) Pseudo Label, (c) Ev-Transfer~\cite{Ev-Transfer}, (d) Image, (e) ESS~\cite{ESS}, (f) E2VID~\cite{rebecq2019high}.}
    \label{fig: sematinc seg}
\end{figure}

\noindent\textit{\textbf{Remarks:}} 
Exploring object detectors that utilize event cameras can address frame-based detectors' shortcomings in challenging conditions, such as high-speed motion or low light. Integrating deep learning-based detectors, like recurrent vision transformers, with event data is promising for enhancing detection capabilities.
While RGB cameras dominate object detection, fusing event and frame data could boost accuracy, especially where events are scarce, like in static scenarios.
The focus is shifting towards multi-modal sensor fusion in detection and tracking, suggesting that event-based approaches could replace costlier sensors like LiDARs, potentially delivering equal or better performance.

\subsubsection{Semantic Segmentation}
Image segmentation~\cite{segsurvey} is a fundamental vision task with many pivotal applications
including robotic perception, scene understanding, augmented reality, \etc. In these practical scenarios, the segmentation models always fail in the non-ideal weather and lighting conditions~\cite{issafe}, leading to poor scene perception of intelligent systems. Event-based semantic segmentation, which is first proposed in Ev-SegNet~\cite{evsegnet}, achieves a significant improvement by utilizing the asynchronous event data. Ev-SegNet also introduces a dataset extended from the DDD17 dataset~\cite{ddd17}. However, the resolution and image quality is less satisfactory for the semantic segmentation task.

To address this problem, Gehrig \etal ~\cite{videotoevent} proposed to convert video data to synthetic events. This work unlocks the usage of a large number of existing video datasets for event-based semantic segmentation. Inspired by this synthetic data source, Wang \etal ~\cite{evdistill,wang2020dual} suggested combining the labeled RGB data and unlabeled event data in a cross-modal knowledge distillation setting, so as to alleviate the shortage of labeled real event data. For higher segmentation results, Zhang \etal ~\cite{issafe} constructed a multi-modal segmentation benchmark model by using the complementary information in both event and RGB branches. More recently, ESS~\cite{ESS} proposed an unsupervised domain adaptation (UDA) framework that leverages the still images without paired events and frames. Jia~\etal~\cite{jia2023event} proposed a posterior attention module to adjust the standard attention based on event data. 

\noindent\textit{\textbf{Remarks:}} 
Due to the lack of precisely annotated large-scale real-world event datasets, existing works mostly focus on generating pseudo labels. However, the labels are not precise enough, rendering the learned segmentation models less robust, as demonstrated by the visual results in Fig.~\ref{fig: sematinc seg}. Future work could further explore the multi-modal domain adaption from RGB data to event data for semantic segmentation.

\subsubsection{Optical Flow Estimation}
\textit{\textbf{Insight}}: 
Optical flow estimation is the process of estimating the motion field within an image sequence.
Conventional RGB-based methods remain unsatisfying in extreme lighting conditions, \eg, at night and in high-speed motion. To overcome these limitations, event cameras have been introduced. The SOTA event-based methods for optical flow estimation can be classified into two categories: traditional methods and DL-based methods. DL-based methods further encompass supervised and unsupervised approaches. Table.~\ref{tab:optical_flow} presents the results achieved by several representative methods in the field.


\begin{figure}[t!]
\centering
\includegraphics[width=0.88\columnwidth]{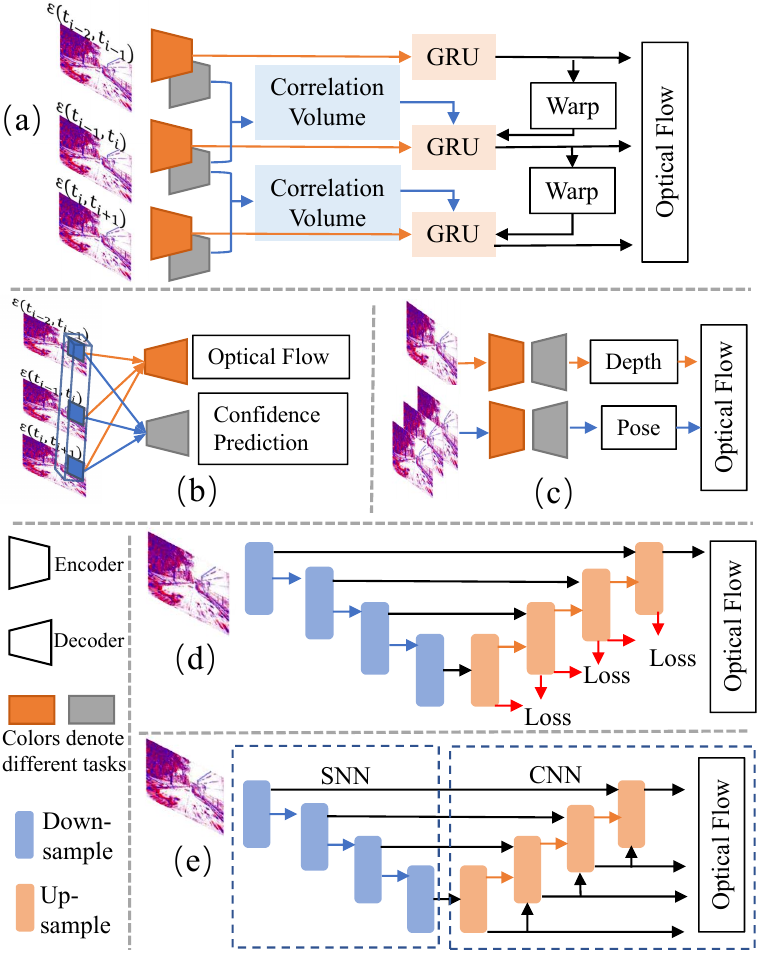} 
\caption{Representative optical flow estimation methods, including supervised methods~\cite{EST,sironi2018hats,maqueda2018event}, (a) Correlation-based methods, (b) Multi-task learning methods, and unsupervised learning methods~\cite{zhu2019unsupervised, 2020Unsupervised30} (c) Multi-task learning methods, (d) self-supervised learning methods, and (e) SNN-based methods.} \label{fig:Optical flow}
\end{figure}

\noindent \textbf{Traditional methods}: Recent research has delved into understanding the principles and characteristics of event data that facilitate the estimation process. These studies have particularly focused on leveraging contrast maximization methods to estimate optical flow accurately~\cite{shiba2022secrets}. 
Furthermore, there is ongoing research aimed at designing innovative event camera platforms specifically tailored for the hardware implementation of adaptive block-matching optical flow. These platforms serve as practical demonstrations of the effectiveness of this approach~\cite{liu2022edflow}.

\noindent \textbf{Supervised methods}: In ~\cite{EST,sironi2018hats,maqueda2018event}, 
event streams are first converted into image-based or surface-based representations and then trained via standard convolutional neural networks (CNNs).
Kepple \etal \cite{kepple2020jointly} proposed to simultaneously generate the region's local flow and the reliability of the prediction. Gehrig \etal \cite{gehrig2021raft} proposed an RNN-based framework that utilizes the cost volumes and learns the feature correlation of the volumetric voxel grid of events, so as to estimate optical ﬂow. 

Some research employs SNNs for optical flow estimation \cite{cuadrado2023optical}. 
However, deep SNNs suffer from spike vanishing problems. To this end, \cite{lee2020spike} combined SNNs and CNNs in an end-to-end manner to estimate optical ﬂow, while \cite{paredes2019unsupervised} proposed a hierarchical SNN architecture
for feature extraction and local and global motion perception. Future research could explore combining SNNs and transformer to learn the global and local visual information from events.

\noindent \textbf{Unsupervised methods}: 
Recent research is focused on unsupervised learning for solving the data scarcity problem. 
Zhu \etal ~\cite{zhu2019unsupervised} introduced a novel event representation containing two channels for encoding the number of positive and negative events and two for the timestamp of the most recent positive and negative events. They utilized the grayscale, \ie, APS, images of the event camera as the self-supervision signals to train the network. 
However, these methods are still based on the photo consistency principle, while this assumption may not be valid in some adverse visual conditions (\eg, high-speed motion). To this end, Zhu \etal \cite{zhu2019unsupervised} proposed a discretized volumetric event representation to maintain the events' temporal distribution, and the input processed event data is used to predict the motions and remove the motion blur.

\noindent \textbf{Remarks}: 
Deep learning-based approaches for optical flow estimation have garnered extensive exploration. Yet, \textbf{\textit{a pertinent avenue of inquiry lies in examining real-time network architectures that can fully leverage the inherent benefits of event cameras, including their low latency and persistent operational capabilities.}}

\subsubsection{Depth estimation}
\begin{figure}[t!]
\centering
\includegraphics[width=0.48\textwidth]{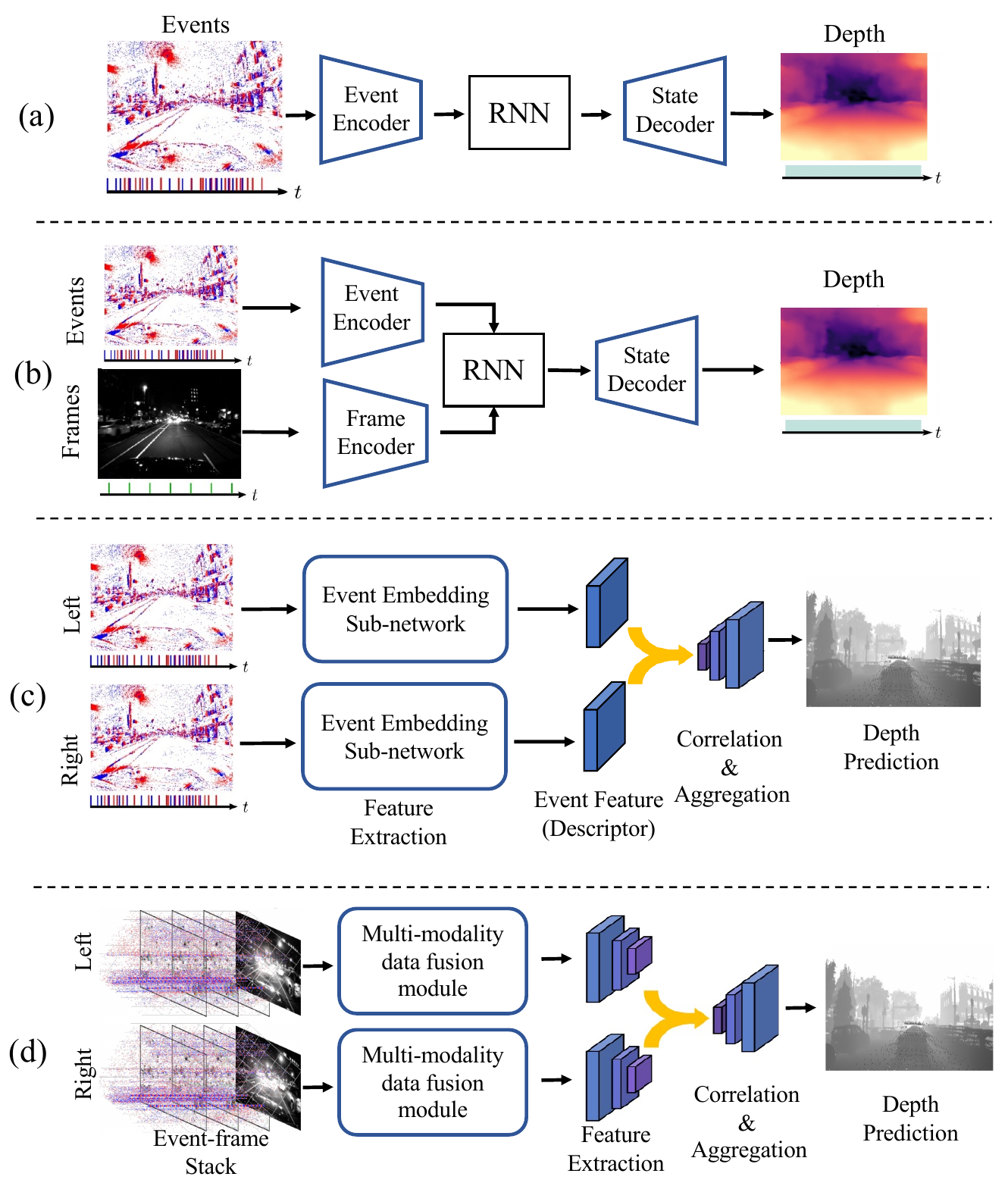} 
\caption{Event-based depth estimation methods, including (a) Monocular event only method~\cite{HidalgoCarrio2020LearningMD}, (b) Monocular event-frame based method ~\cite{Gehrig2021CombiningEA}, (c) Stereo event only method~\cite{Tulyakov2019LearningAE, Uddin2022UnsupervisedDE},  (d) Stereo event-frame based method ~\cite{MostafaviIsfahani2021EventIntensitySE,Nam2022StereoDF,Cho2022SelectionAC}.} \label{fig:depth}
\end{figure}

\textit{ \textbf{Insight} }: 
Events streams reflect abundant edge information, HDR, and high temporal resolution,
benefiting depth estimation tasks, especially in extreme conditions. Depth can be learned from either the monocular (single) input or stereo (multi-view of a scene) inputs. Under this outline, we categorize the depth estimation methods based on how events are used and learned.  

\noindent \textbf{Monocular depth estimation}:
Based on whether events are used alone or combined with the intensity frames, we divide the monocular depth estimation methods into two types. 1) Events-only approaches: \cite{HidalgoCarrio2020LearningMD} is a representative approach that adopts a recurrent network5 
to learn the temporal information from grid-like event inputs, depicted in Fig.~\ref{fig:depth} (a). However, as monocular depth estimation from events is an ill-posed problem, rendering such a learning framework difficult to achieve highly precise depth. Moreover, this learning paradigm may fail to predict depth in the static scene as events are only triggered by motion.
2) event-plus-frame approaches: RAM~\cite{Gehrig2021CombiningEA} employs the same RNN as \cite{HidalgoCarrio2020LearningMD} but combines events and frames (\ie, as complementary to each other) to learn to predict depth from the multi-modal inputs asynchronously, as shown in Fig.~\ref{fig:depth} (b). Nonetheless, the recurrent network is inevitably accompanied by long-term memory costs.

\noindent \textbf{Stereo depth estimation}: 
As the visual cues of the left and right event cameras are used in the stereo setting, the model complexity and memory cost of learning pipelines become more prohibitive. 
\cite{Tulyakov2019LearningAE,Uddin2022UnsupervisedDE} are two pioneering works in stereo depth estimation, as shown in Fig.~\ref{fig:depth} (c). 
In particular, DDES~\cite{Tulyakov2019LearningAE} is the first learning-based stereo-matching method, and in ~\cite{Uddin2022UnsupervisedDE}, the first unsupervised learning framework is proposed. 
Both methods store events at each position as a First-in First-out queue, enabling concurrent time and polarity reservation.
To adaptively extract features from sparse data, Zhang \etal ~\cite{Zhang2022DiscreteTC} proposed continuous time convolution and discrete time convolution to encode high dimensional spatial-temporal event data.

By contrast, some research explores multi-modality fusion under different settings which serves as a remedy to utilize the benefits of each modality. 
EIS~\cite{MostafaviIsfahani2021EventIntensitySE} is a representative work to combine events and frames with a recycling network, as depicted in Fig.~\ref{fig:depth} (d). However, since events are sparse, event stacking is an important factor that affects the quality of fusion and depth prediction because stacking inappropriate amounts of events can lead to information overriding or missing problems. To this end, ~\cite{Nam2022StereoDF,Cho2022SelectionAC} propose a selection module to filter more useful events. Specifically, 
Nam \etal ~\cite{Nam2022StereoDF} concatenated the event stacks with different densities and then adaptively learn these stacks to highlight the contribution of the well-stacked events. Moreover, considering the constant motion of the cameras, SCSNet~\cite{Cho2022SelectionAC} introduces a differentiable event selection network to extract more reliable events and correlate the feature from a neighbor region of events and images, diminishing the disruption of bad alignment intuitively.

\noindent\textit{\textbf{Remarks}}: From our review, inter-camera spatial correlation is the key to content matching between events and frames.
Although~\cite{MostafaviIsfahani2021EventIntensitySE,Nam2022StereoDF,Cho2022SelectionAC} combine events and frames, it still deserves exploring which part events contribute most to the multi-modal feature fusion and alignment. Also, it is possible to use an event camera and a frame-based camera for stereo depth estimation. Future research could consider exploring these directions. 

\subsection{3D Vision}
\label{sec:3D Vision}
\subsubsection{Visual SLAM}
\textit{\textbf{Insight}}: 
It is an essential module for various applications, \eg, robotic navigation and virtual reality. Visual SLAM receives the signals, \eg, 2D images, as the source for ego-motion estimation and builds 3D maps, which can generally be defined as the tracking thread and the mapping thread. Event-based visual SLAM shares a similar spirit and benefits from the robustness of event cameras to light-changing and fast-moving conditions.

\begin{figure}[t!]
\centering
\includegraphics[width=0.99\columnwidth]{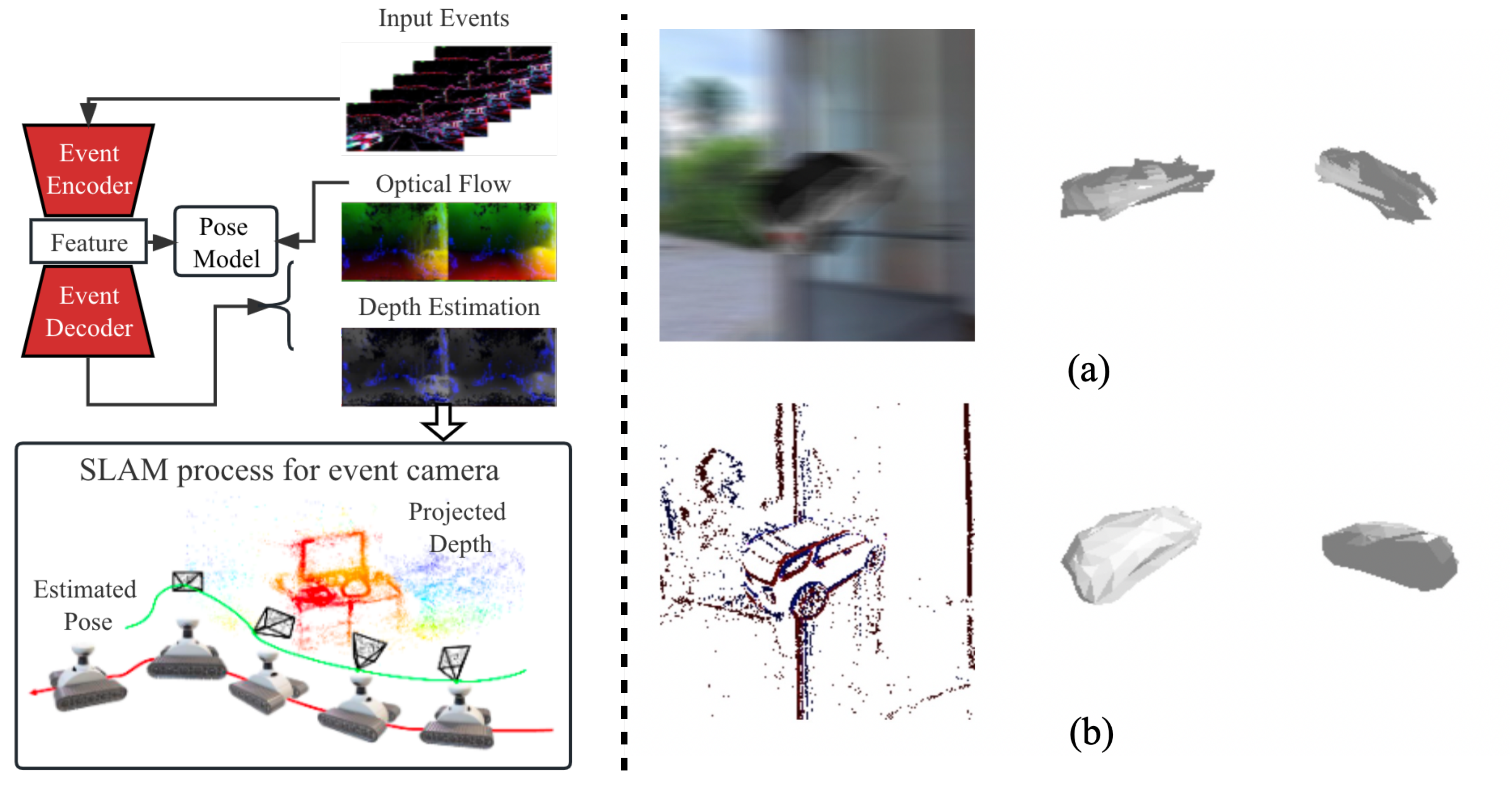}
\caption{(Left) Illustration of the general framework for event-based SLAM via deep learning, key elements are retrieved from \cite{event_slam_pic1,event_slam_pic2,2020Unsupervised30,zhu2019unsupervised}. The colored point cloud is a map reconstructed from event data of different sensor views (green trajectory) grounded by robot location (red directional trajectory). (Right) Visualization of 3D reconstruction from (a) blurred RGB images and (b) events from ~\cite{E3D}.
}
\label{fig:event_slam}
\end{figure}

In traditional SLAM, depth and ego-motion could be easily estimated through triangulation and the subsequent local pose estimation by, \eg, Perspective-N-Points. Similar approaches have been applied on event stream\cite{zhou2021event,chen2023esvio}.
Recently, the learning-based approaches are also been applied in SLAM, such as, \cite{2020Unsupervised30,zhu2019unsupervised}  propose united frameworks\textemdash with an encoder-decoder structure\textemdash for optical flow, depth, and ego-motion estimation. 

In particular, in \cite{2020Unsupervised30}, a united framework is proposed to estimate sparse optical flow, depths, and ego-motion, in which an encoder-decoder structure was adopted for sparse depth estimation. 
By contrast, Zhu \etal \cite{zhu2019unsupervised} proposed to directly learn 6-DOF poses from multi-view intensity frames, other than deriving from optical flow and depth, like \cite{2020Unsupervised30}.
The recent EAGAN \cite{lin_eagan_2022} adopts the vision transformer ~\cite{2017transformer} to boost the accuracy of optical flow estimation, yet no optimization is done to depth estimation. However, EAGAN shows an increase in the number of learnable parameters compared to the methods~\cite{zhu2019unsupervised,2020Unsupervised30}, and the ego-motion estimation is not considered. 

Moreover, a branch of research casts event-based SLAM as a re-localization problem.
This paradigm proposes to directly learn the camera poses from extracted deep features in an end-to-end trainable manner.
Later on, additional denoising modules are introduced in \cite{2020CNNLSTM32} to further increase the pose estimation accuracy.

\noindent\textit{\textbf{Remarks}}: 
Currently, the event-based SLAM systems employing deep learning are generally decoupled as separate modules rather than the cross-event frame and pose-map joint estimation as traditionally processed in visual SLAM. Some attempts \cite{ 2020Unsupervised30} generate frame association (as a form of optical flow), pose estimation, and depth map with a unified network architecture. Still, global consistency remains an unsolved problem. For the front end, applying DNNs for the intra-frame association purpose is a challenging problem. 
This leaves a vacancy for exploration since many frame-based SLAM systems have proved that the pose derived by cross-frame feature association is highly precise. Another interesting direction to investigate is the dense map or mesh reconstruction from event data that served as mapping for SLAM. we have seen mesh reconstruction with very limited precision and designed for small-size objects in Sec.\ref{3DRecon}, leaving learning-based scene reconstruction an open problem.
\subsubsection{3D Reconstruction}
\begin{table*}[t!]
\centering
\caption{Comparison of existing representative event-based 3D reconstruction methods.}
\label{3DRecon}
\begin{tabular}{llllccc}
\toprule
Publications & Methods          & Task                         & Representations & Frame Input & Real Time & Multi Cameras \\ \midrule
Arxiv 2020   & E3D~\cite{E3D}              & 3D Reconstruction            & Image-based     & \Checkmark         & \XSolidBrush         & \XSolidBrush             \\ 
ECCV 2020    & Stereo-event PTV~\cite{Classic3D1} & 3D Fluid Flow Reconstruction & Stream-based    & \XSolidBrush           & \XSolidBrush         & \Checkmark           \\ 
ICCV 2021    & EventHands~\cite{EventHands}       & 3D Hand Pose Estimation      & Surface-based   & \XSolidBrush           & \Checkmark       & \XSolidBrush             \\ 
ECCV 2022    & EvAC3D~\cite{EvAC3D}           & 3D Reconstruction            & Surface-based   & \XSolidBrush           & \XSolidBrush         & \XSolidBrush          
\\ \bottomrule
\end{tabular}
\end{table*}


\textit{\textbf{Insight:}} Event cameras capture dominant scene features,~\eg, edges and silhouettes, making them more suitable for some 3D reconstruction methods than the frame-based data (See Fig.~\ref{fig:event_slam}).

Unlike the traditional frame-based RGB and depth cameras widely explored in 3D reconstruction~\cite{3Dsurvey}, event cameras enjoy inherent benefits, such as low latency and HDR~\cite{6-DoF}. 
Intuitively, approaches designed for prior RGB and depth data can not be directly applied to event data due to the distinct data format. Thus, endeavors have been made in converting events to sparse, semi-dense point clouds and full-frame depth maps for existing RGB-based pipelines~\cite{Classic3D1, Gehrig2021CombiningEA}. 
These methods are committed to taking advantage of the off-the-shelf 3D reconstruction pipelines built for RGB-based inputs while ignoring the unique strengths of event cameras.
In the following research, event cameras are combined with RGB cameras for the advantages of both sensors. Besides, some approaches combine event cameras with other types of sensors, such as ELS~\cite{ELS} that uses a laser point-projector and an event camera.

To directly take advantage of the event data, EvAC3D~\cite{EvAC3D} explores the direct reconstruction of mesh from a continuous stream of events while defining the boundaries of the objects as apparent contour events and continuously carving out high-fidelity meshes. The comparison of existing representative event-based 3D reconstruction methods is shown in Tab.~\ref{3DRecon}.

\noindent\textit{\textbf{Remarks}}: 
Recent research provides important insights regarding how events can be utilized to understand the 3D world. However, a general and unified pipeline is expected, which is left for future research. The fusion of the RGB and event cameras is also valuable for achieving better reconstruction results. 
\subsubsection{3D Human Pose and Shape Estimation}
\label{sec:3DHPE}
\begin{figure}[t!]
\centering
\includegraphics[width=0.8\columnwidth]{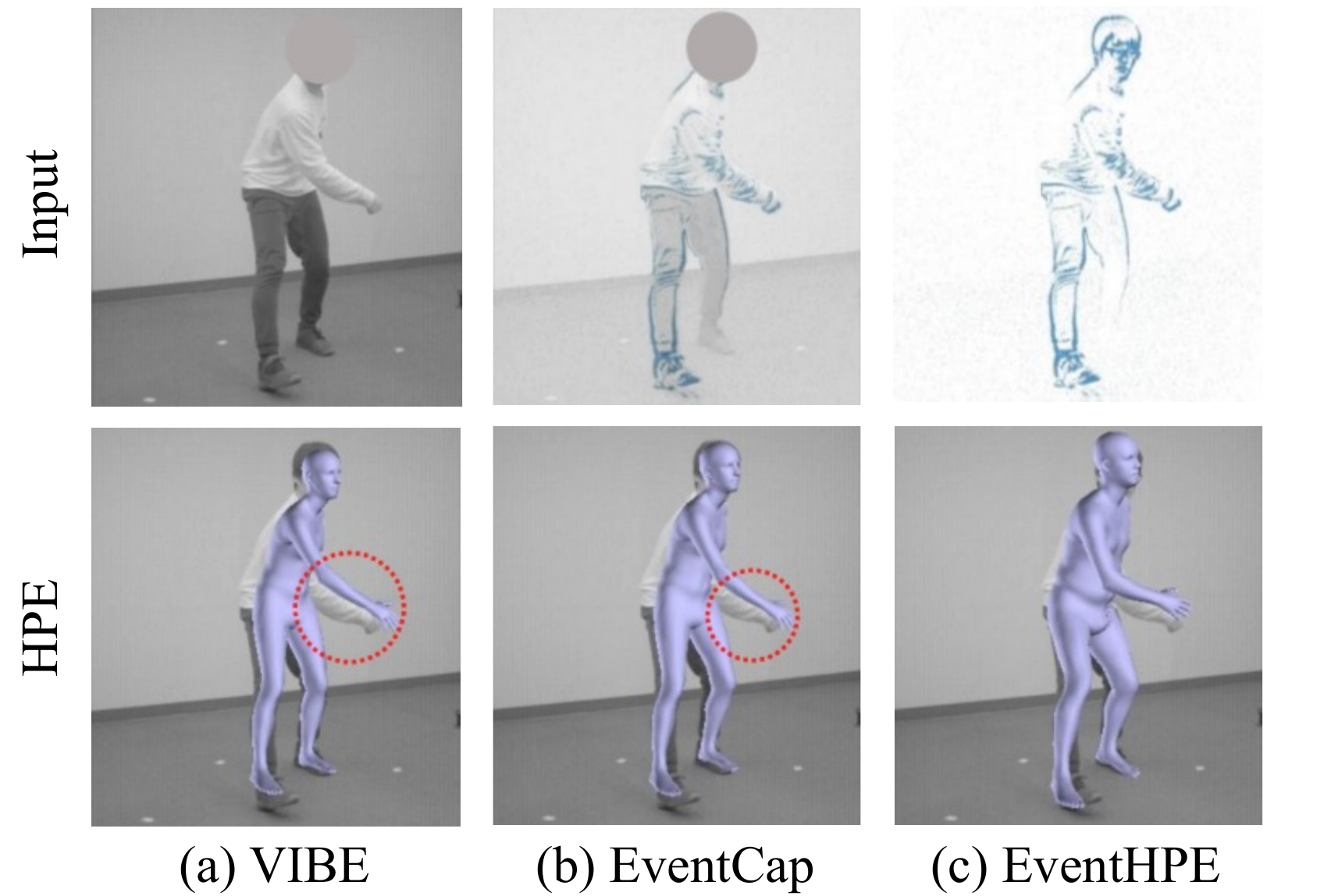}
\caption{Qualitative comparison between 3D HPE methods from ~\cite{EventHPE}. (a) VIBE~\cite{VIBE}; (b) EventCap~\cite{EventCap}, (c) EventHPE~\cite{EventHPE}}. \label{fig:HPE}
\centering
\end{figure}

\textit{\textbf{Insight}}: The ability to capture the dynamic motion makes event cameras superior for estimating 3D moving objects, especially for 3D human pose and shape estimation (3D HPE)  (see Fig.~\ref{fig:HPE}). 

In the past few years, 3D HPE has been extensively explored with RGB images and videos in the deep learning era~\cite{humansurvey}. The most challenging scenario in 3D HPE is always related to the high-speed motion~\cite{VIBE}, which is essential in many practical applications, such as sports performance evaluation. However, the RGB cameras suffer inevitable fundamental problems~\cite{EventCap}, including unsatisfactory frame rates and data redundancy. By contrast, event cameras are more advisable for fast-motion scenarios. 

DHP19~\cite{DHP19} is the first DL-based pipeline and provides the first dataset for event-based 3D HPE.
More recently, EventCap~\cite{EventCap} is the first work to capture high-speed human motions from a single event camera with the guidance of gray-scale images. 
To alleviate the reliance on frame inputs, the following research EventHPE~\cite{EventHPE}, proposes to infer 3D HPE from the sole source of event input, given the beginning shape from the first frame of the intensity image stream. 

\noindent \textit{\textbf{Remarks}}: 
A limitation of these approaches is that they need gray-scale images for initialization. \textit{For future research, it is worth investigating how to infer 3D HPE purely from event signals without additional priors. Meanwhile, it also promises to combine the advantages of both RGB and event sensors and design multi-modal learning frameworks for more robust 3D HPE.
}
\section{Research Trend and Discussions}

\subsection{Discussions}
\label{sec:importantdiscussions}

\noindent \textit{\textbf{Advanced applications of Event cameras with AI:}}
Compared with frame-based cameras, event cameras are energy-efficient sensors with low latency and an always-on ability, this preferable characteristic in real-time applications such as nano drones and VR/AR headsets. 
The cutting-edge event cameras, such as \eg, EVK4 from Prophesse \footnote{https://www.prophesee.ai/event-camera-evk4/} together with the advanced AI techniques can be applied in many industrial scenarios, \ie, spatter monitoring, edge-let tracking, and even in medical samples sterility testing.

\noindent \textit{\textbf{Low latency of event cameras vs. High computation of DNNs}:}
One of the notable superiorities of event cameras is the low latency which enables real-time applications of event cameras.
However, the computational complexity of the neural network is typically enormous, potentially negating the benefits of lower event latency.
An important research question is how to accelerate the neural network in the field of events.
We suggest that three angles be taken when conducting this issue.
1) Make use of a light-weight network architecture.
2) Make use of network quantization and compression techniques.
3) Construct a network based on SNN, which has low delay and sparsity.

\noindent \textit{\textbf{Do we really need deep networks for learning events?}}
The success of DL in frame-based computer vision has spurred interest in applying DL to event data. DL approaches have shown marked accuracy improvements over traditional event-based vision algorithms, such as a recent DL-based feature tracker surpassing non-DL methods. Despite these advancements, DL methods may not always retain the beneficial properties of event sensors. To optimize their combination, we suggest: 1) For ultra-high temporal resolution tasks like microfluidic analysis, DL is less suitable; 2) For high-level semantic tasks like action recognition, advanced DL models can improve accuracy; 3) Event cameras can effectively complement frame-based sensors using DL models. In summary, integrating DL with event cameras in proper scenarios offers significant potential in various applications. 

\noindent \textit{\textbf{Integrating DNNs with event cameras: Neuromorphic, Federated, and Edge Computing Approaches}
In integrating DNNs with event cameras, we highlight approaches that maintain the cameras' low latency and efficiency. Neuromorphic computing, inspired by the human brain, significantly reduces power and computational needs, ideal for real-time processing. Federated learning, through decentralized processing, enhances efficiency and privacy in robotics using event-based sensing. Edge computing, by processing data closer to its source, reduces latency for quicker robotic decisions. These methods demonstrate how unconventional approaches can leverage event cameras with DNNs, opening new applications in dynamic settings.
}

\noindent \textit{\textbf{How to better deal with noisy events with DNNs?}}
Random noise can be introduced throughout the trigger process of events due to many reasons. Thus the denoising procedure is necessary for accurate information capture. According to the existing denoising methods in Sec.~\ref{subsec:event+denoising}, we suggest dealing with the noisy events with DNNs in three steps:
1) Formulate the spatial and temporal distributions of raw events;
2) Separately denoising from both spatial and temporal perspectives, \eg, operations in spatial neighbourhoods and current event surfaces;
3) Further consider the correlation between spatial and temporal distributions, aiming at maintaining spatial-temporal correlation.

\noindent \textit{\textbf{Can the high temporal resolution of events be fully reflected by DNNs?}}
No matter what kind of event representation, summarized in Sec.~\ref{sec:Event Representation}, is used as DNN's input, the grid-like tensors always lose some partial information of raw events, such as temporal information in the image-based representations~\cite{maqueda2018event,zhu2018ev,EST,deng2020amae,bai2022accurate,deng2021mvf}. The high temporal resolution of events cannot be fully reflected by the existing DNNs which are proposed to solve frame-based vision.
The high temporal resolutions are predominantly fused to frame-like representations for the downstream tasks. SNNs can solve this problem according to their architectures, however, there are still technical difficulties in bringing SNNs into practical applications. Thus specific neural networks directly designed for event data are in the future outlook.


\noindent \textit{\textbf{ Lower accuracy of event-based vision models in the dark.}}
Low light conditions significantly increase shot noise and blur (due to prolonged photoreceptor response times) in DVS.
The noises can be briefly called "holes" or false negatives. This leads to unreliable scene understanding of event cameras. 
Employing DL for noise cancellation of event data is a promising approach \cite{wang2020joint}.
In addition, multi-sensor fusion is a valuable direction. 
For example, thermal sensors have been widely used in the dark \cite{zhang2021object}. Therefore,
developing DL-based methods that take thermal and event data as inputs is worth exploring.

\noindent \textit{\textbf{Are DL-based methods more advantageous than optimized-based ones?}}
Optimization-based methods are more applicable to edge computing ~\cite{battiti1992first} but are still unlikely to reach global optima\textemdash more likely to be trapped in saddle points ~\cite{Goodfellow-et-al-2016} ).
By contrast, DNNs are superior in that they can flexibly learn the multi-dimension data and extract better feature representations. Intuitively, DNNs have the potential to learn spatial-temporal information from events. Superior results has been demonstrated in exploring the temporal correlation of events with various DNNs in the literature.
Also, owning to the numerical stability and efficiency, DNNs allow hidden state encoding for effective prediction in various tasks. For example, in the context of event-based SLAM, implicit encoding of event streams exhibits higher spatial-temporal consistency than naively tracking fired pixels, enabling estimation of ego-motion and 3D scene reconstruction jointly, whereas the optimization-based method shows significantly lower accuracy.

\noindent \textit{\textbf{Is focal alignment necessary between RGB and event pixels in event cameras?}}
Event cameras have emerged as an efficient alternative for capturing motion information. Recent studies highlight the potential of DNNs in leveraging both RGB and event data to enhance images, enabling crucial tasks like deblurring and video frame interpolation. This capability holds significant practical applications in areas such as augmented reality and virtual reality. However, achieving the desired image quality mandates RGB characteristics comparable to advanced mobile RGB sensors, along with precise focal alignment between RGB and event pixels on the sensor. To address these challenges, one promising approach involves the development of hybrid-type sensors that seamlessly integrate high-frame-rate event pixels with advanced mobile RGB pixels.

\subsection{New Directions}
\label{sec:new-direction}
\noindent\textbf{\textit{NeRF for Event-based Neural Rendering:}}
NeRF ~\cite{NeRF} is a representative neural implicit 3D representation method that synthesizes 3D objects with volume rendering techniques. 
Most existing research in NeRF is investigated based on RGB cameras, which suffer from inevitable shortcomings, \eg, low dynamic range and motion blur in unfavorable lighting conditions. 
Thus, recent attention has been paid to the usage of event cameras for NeRF~\cite{EventNeRF, E-NeRF, Ev-NeRF}.
The first work is EventNeRF~\cite{EventNeRF}, which is trained with pure event-based supervision. It demonstrates that the NeRF estimation from a single fast-moving event camera in unfavourable scenarios (\eg, fast-moving objects, motion blur, or insufficient lighting) is feasible while frame-based approaches fail. 
Moreover, E-NeRF~\cite{E-NeRF} takes the strengths of RGB and event cameras by combining color frames and events to achieve sharp and colorize reconstruction results.
From the review of recent progress, event-based NeRF methods show superior abilities than the frame-based NeRF methods.
Since the event-based NeRF is an emerging direction, future research could focus on improving the technique pipelines and more lightweight DNNs for mobile applications.

\noindent\textbf{\textit{Multi- and Cross-modal Learning for Event-based Vision:}}
In practical scenarios, event cameras always play an auxiliary role to provide multi-modal guidance in many aforementioned computer vision tasks, \eg,  image and video SR~\cite{han2021evintsr,wang2020event,jing2021turning}. With the development of event cameras, event-based vision will occupy an increasingly dominant position in both research and industry. Especially in some specific domains, there are already attempts to leverage pure event data to facilitate task accuracy, such as reconstructing RGB images and videos from pure event data~\cite{MostafaviIsfahani2018EventBasedHD,rebecq2019events,zou2021learning,rebecq2019high}, \etc. Consequently, how to utilize the domain knowledge from the frame-based vision to the emerging event-based vision deserves more intensive research. Recently, CTN~\cite{CTN} and EventDance~\cite{zheng2024eventdance} are proposed as early attempts of transferring knowledge from frame-based vision to event-based vision. 

\noindent\textbf{\textit{Event-based model Pretaining:}}
Pre-trained neural networks are the foundations of almost all downstream task models in the deep learning era. With the growing interest in event-based vision, a wide range of datasets are collected in many downstream tasks. Thus, unified pre-trained weights with large-scale data are required for better accuracy.
Since the event data is a totally distinct format containing spatial-temporal information, a reliable and efficient pre-trained method is required for various applications.
In ~\cite{eventdatapretrain}, Yang \etal~proposed the first pre-training pipeline for dealing with event camera data, including various event data augmentations, a masking sample strategy, and a contrastive learning approach. Recently developed, Masked Event Modeling (MEM)~\cite{klenk2024masked}, a self-supervised framework, pretrains neural networks using unlabeled data from event camera recordings. The pretrained model is then finetuned for specific downstream tasks, consistently enhancing task accuracy.
Moreover, several frameworks, such as E-CLIP~\cite{zhou2023clip} and Ev-LaFOR~\cite{cho2023label} are proposed to unleash the potential of CLIP for event-based recognition to compensate for the lack of large-scale event-based datasets.
\textbf{\textit{More details of new directions can be found in the supplementary material.}}
\bibliographystyle{IEEEtran}
\bibliography{references_1}
\vspace{-40pt}
\begin{IEEEbiography}[{\includegraphics[width=0.6in,height=0.8in,clip,keepaspectratio]{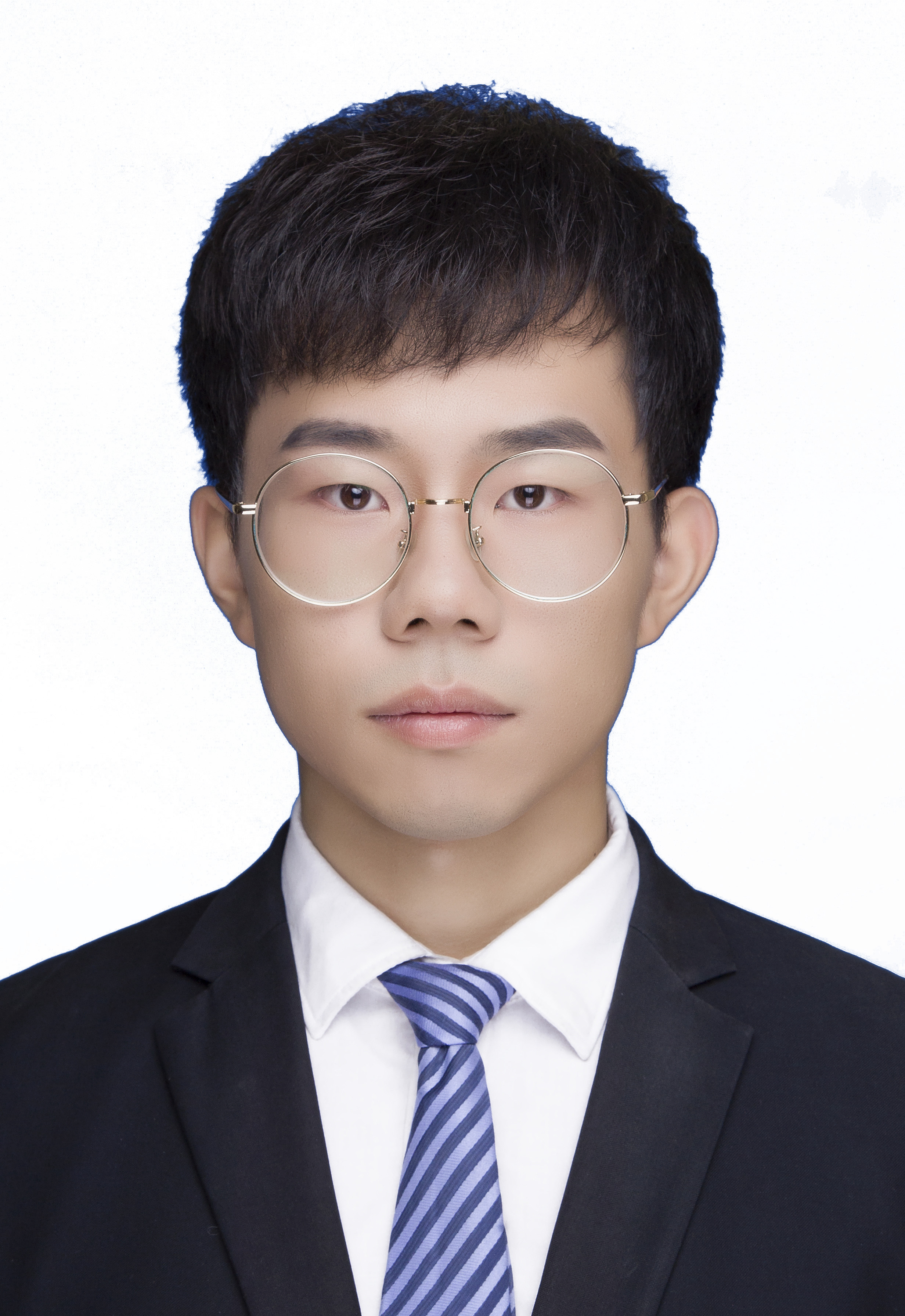}}] {Xu Zheng} (IEEE Student Member) is a Ph.D. student in the Visual Learning and Intelligent Systems Lab,  Artificial Intelligence
Thrust, The Hong Kong University of Science and Technology,  Guangzhou (HKUST-GZ). His research interests include multi-modal learning, event-based vision, transfer learning, \etc.
\vspace{-70pt}
\end{IEEEbiography}

\begin{IEEEbiography}[{\includegraphics[width=0.6in,height=0.8in,clip,keepaspectratio]{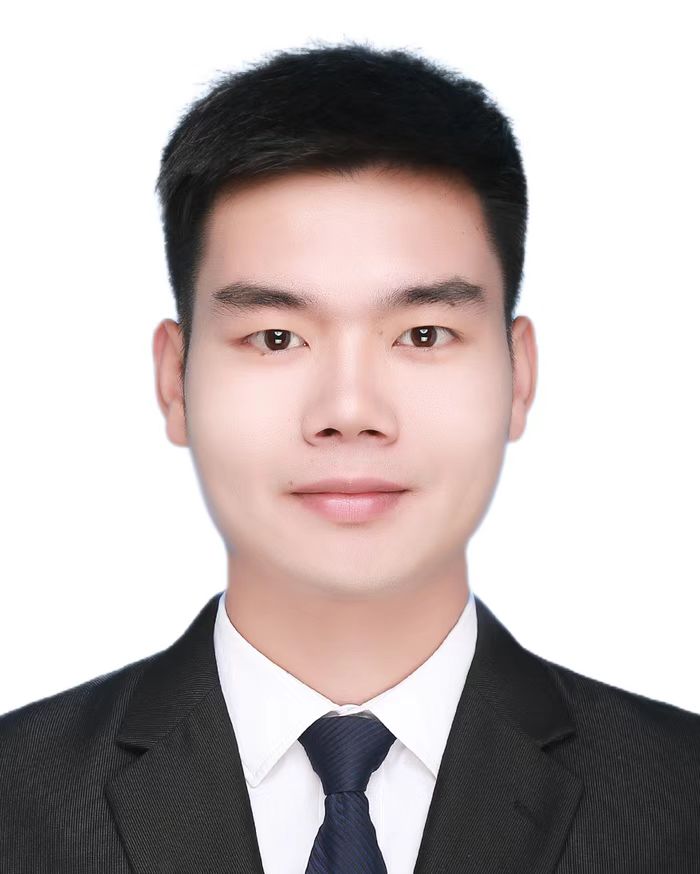}}] {Yexin Liu} is a Mphil. student in the Visual Learning and Intelligent Systems Lab,  Artificial Intelligence
Thrust, The Hong Kong University of Science and Technology,  Guangzhou (HKUST-GZ).
His research interests include infrared- and event-based vision, and unsupervised domain adaptation.
\vspace{-60pt}
\end{IEEEbiography}

\begin{IEEEbiography}[{\includegraphics[width=0.6in,height=0.8in,clip,keepaspectratio]{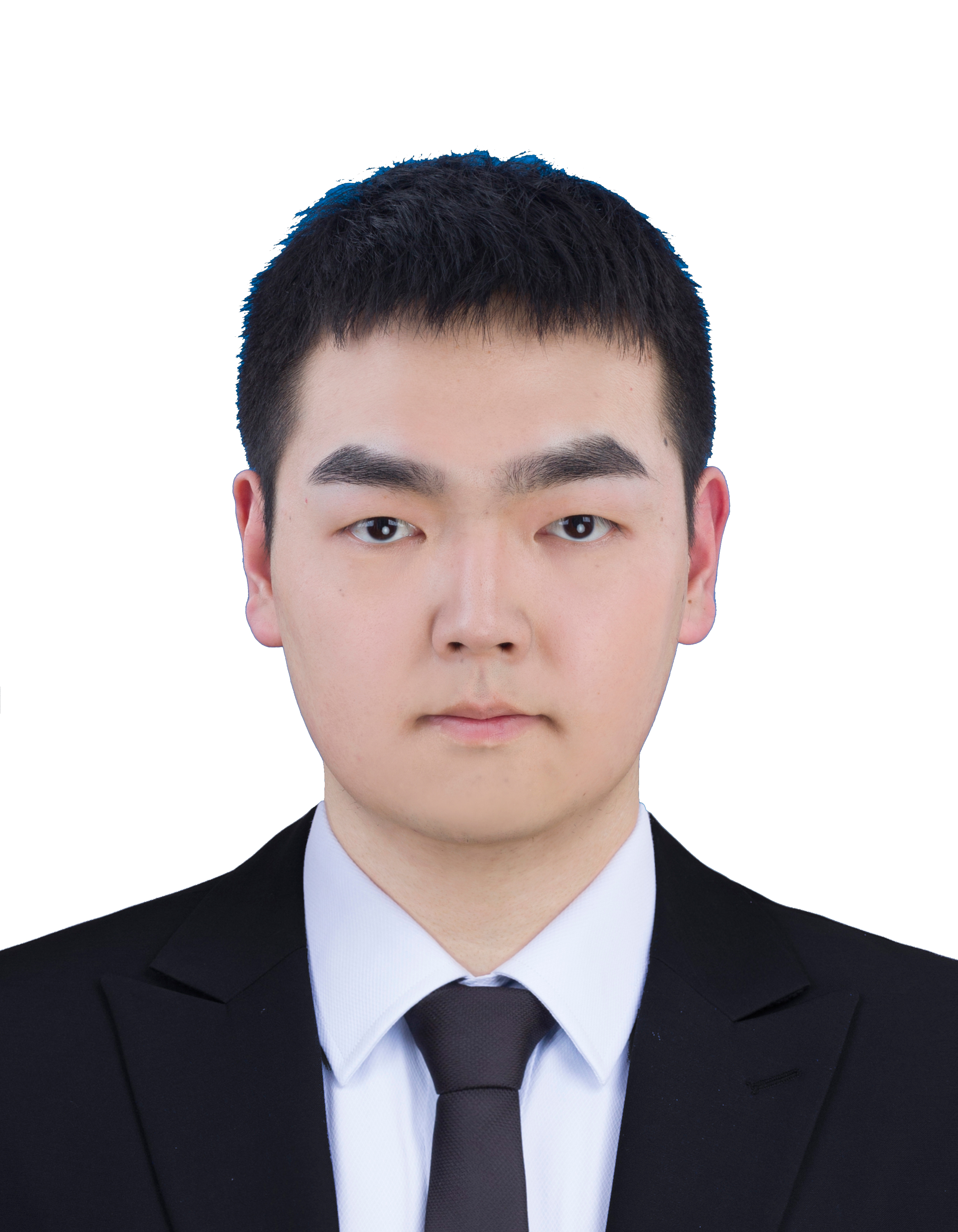}}] {Yunfan Lu}
is a Ph.D. student in the Visual Learning and Intelligent Systems Lab,  Artificial Intelligence
Thrust, The Hong Kong University of Science and Technology,  Guangzhou (HKUST-GZ).
His research interests include low-level vision (event camera, deblurring, SR), pattern recognition (image classification, object detection), and DL (transfer learning, unsupervised learning).
\vspace{-60pt}
\end{IEEEbiography}

\begin{IEEEbiography}[{\includegraphics[width=0.6in,height=0.8in,clip,keepaspectratio]{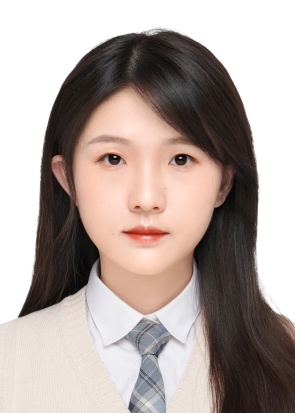}}] {Tongyan Hua}
is a research assistant in the Visual Learning and Intelligent Systems Lab,  Artificial Intelligence
Thrust, The Hong Kong University of Science and Technology,  Guangzhou (HKUST-GZ).
Her research interests include robotics vision, Simultaneous localization and mapping (SLAM), Deep Learning, \etc.
\vspace{-60pt}
\end{IEEEbiography}

\begin{IEEEbiography}[{\includegraphics[width=0.6in,height=0.8in,clip,keepaspectratio]{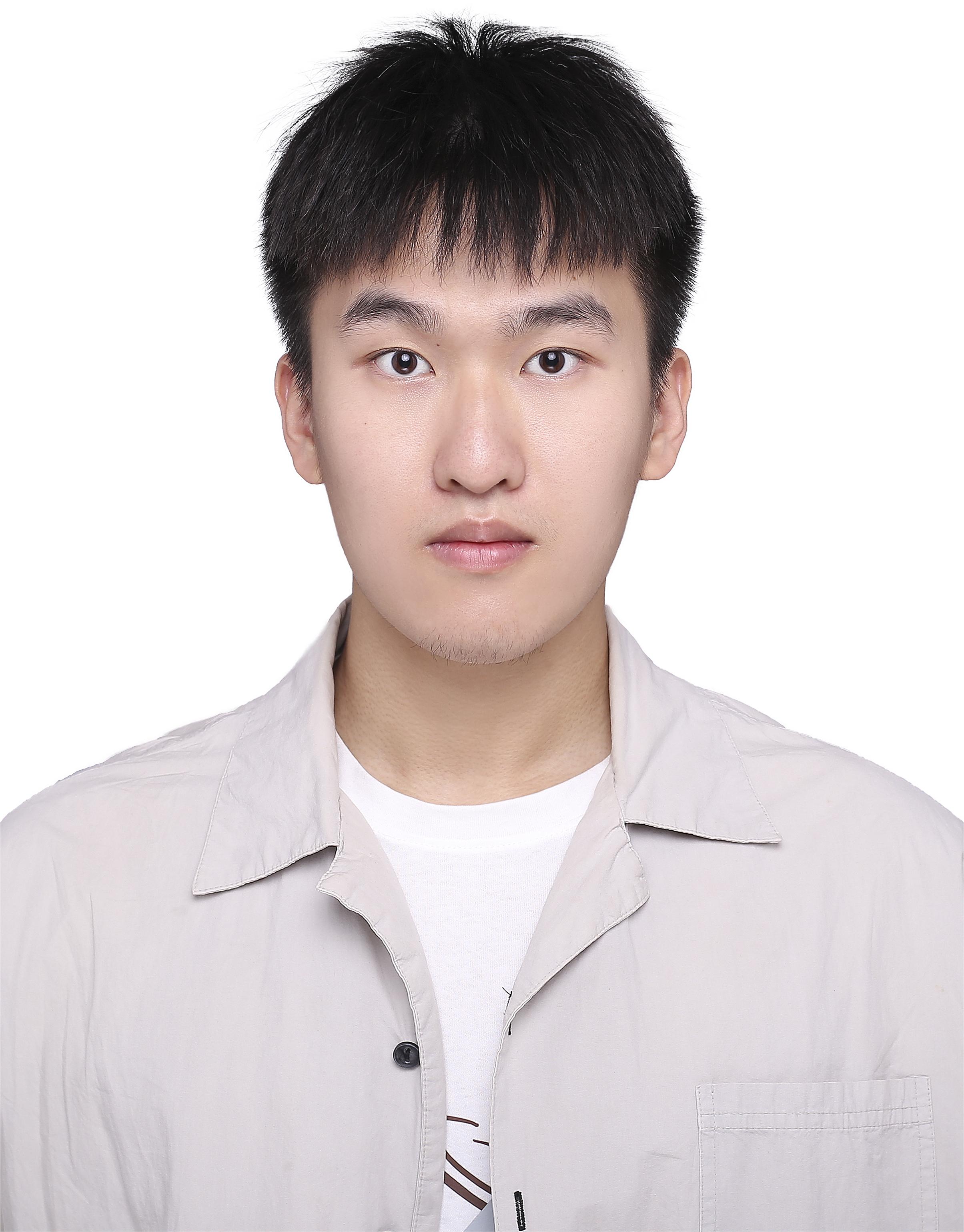}}] {Tianbo Pan}
is a Mphil student in the Visual Learning and Intelligent Systems Lab,  Artificial Intelligence
Thrust, The Hong Kong University of Science and Technology,  Guangzhou (HKUST-GZ).
His research interests include event-based vision, Simultaneous localization and mapping (SLAM), 3D vision, \etc.
\vspace{-60pt}
\end{IEEEbiography}

\begin{IEEEbiography}[{\includegraphics[width=0.6in,height=0.8in,clip,keepaspectratio]{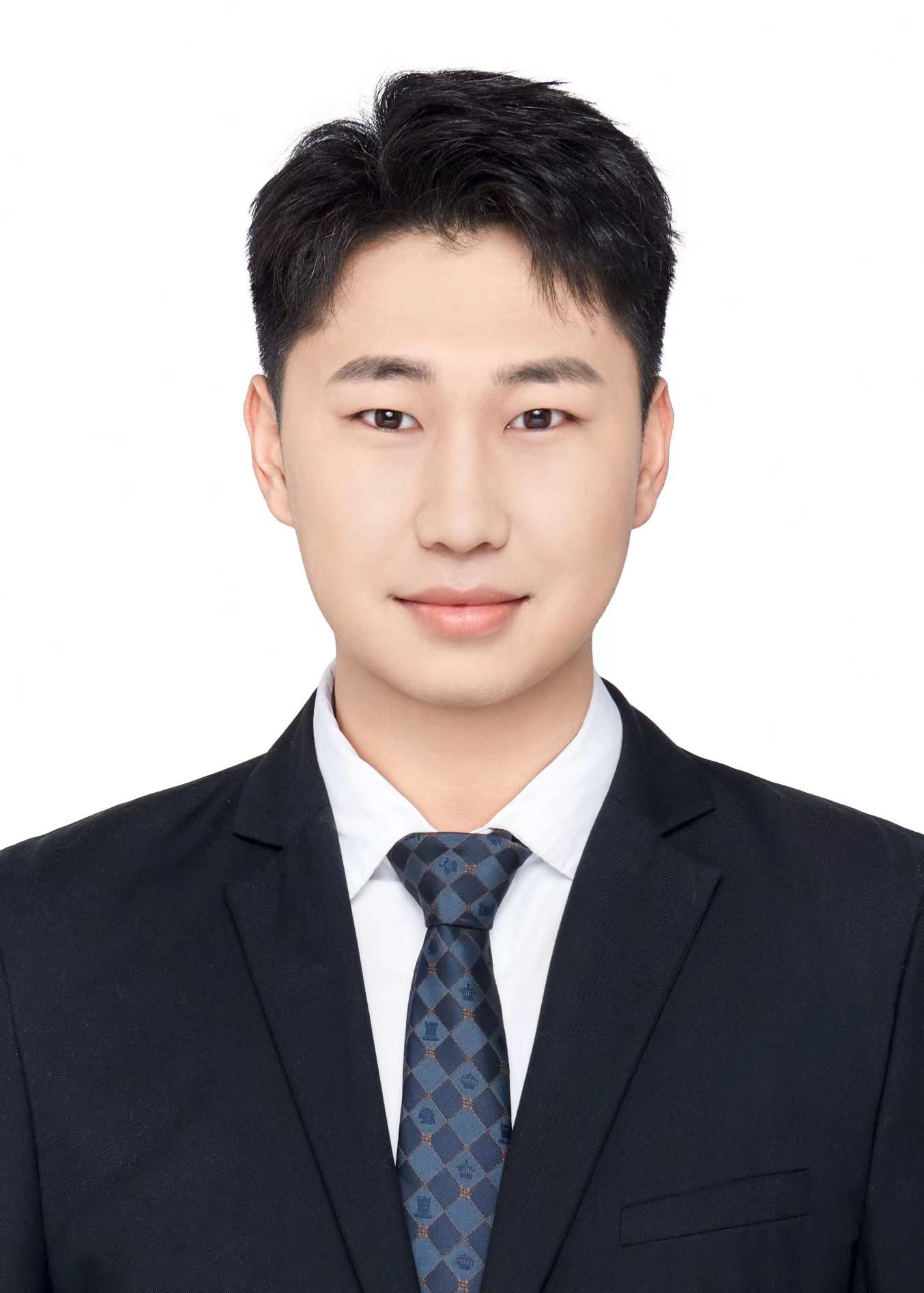}}] {Weiming Zhang}
is a research assistant in the Visual Learning and Intelligent Systems Lab,  Artificial Intelligence
Thrust, The Hong Kong University of Science and Technology,  Guangzhou (HKUST-GZ).
His research interests include event-based vision, Deep Learning, \etc.
\vspace{-60pt}
\end{IEEEbiography}

\begin{IEEEbiography}[{\includegraphics[width=0.6in,height=0.8in,clip,keepaspectratio]{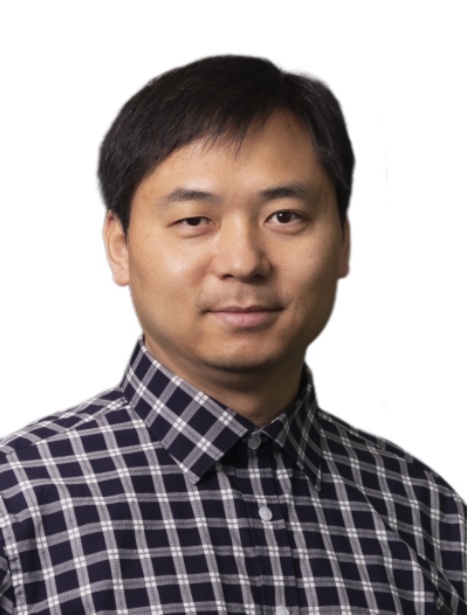}}] {Dacheng Tao} (Fellow IEEE) the Inaugural Director of the JD Explore Academy and a Senior Vice President of JD.com. He is also an advisor and chief scientist of the digital sciences initiative at the University of Sydney. He mainly applies statistics and mathematics to AI and data science, and his research is detailed in one monograph and over 200 publications in prestigious journals and proceedings at
leading conferences. He received the 2015 Australian Scopus-Eureka Prize, the 2018 IEEE ICDM Research Contributions Award, and the 2021 IEEE Computer
Society McCluskey Technical Achievement Award. He is a fellow of the
Australian Academy of Science, AAAS, ACM and IEEE.
\vspace{-40pt}
\end{IEEEbiography}

\begin{IEEEbiography}[{\includegraphics[width=0.6in,height=0.8in,clip,keepaspectratio]{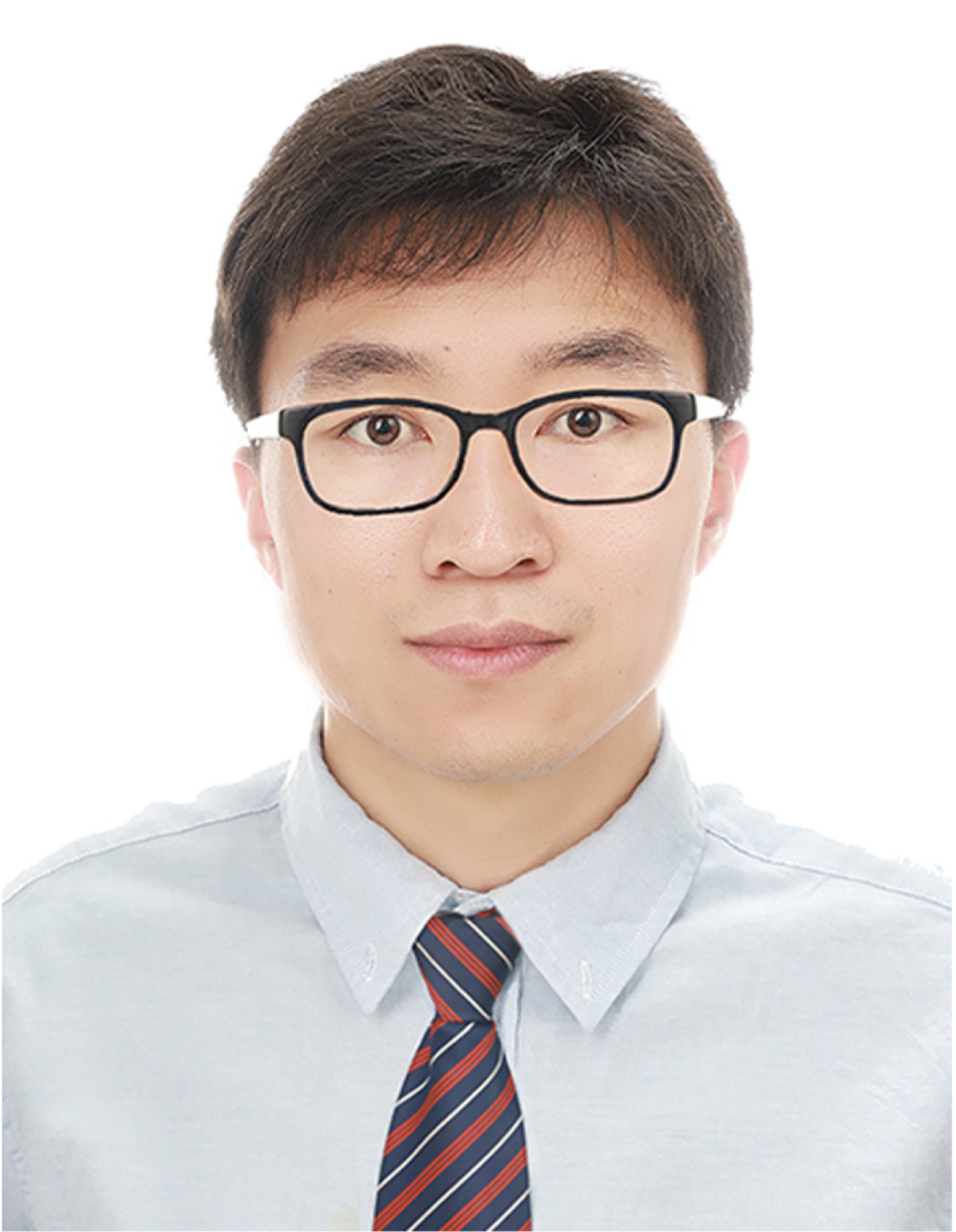}}] 
{Lin Wang} (IEEE Member) is an assistant professor in the AI Thrust, HKUST-GZ, HKUST FYTRI, and an affiliate assistant professor in the Dept. of CSE, HKUST. He did his Postdoc at the Korea Advanced Institute of Science and Technology (KAIST). He got his Ph.D. (with honors) and M.S. from KAIST, Korea. He had rich cross-disciplinary research experience, covering mechanical, industrial, and computer engineering. His research interests lie in computer and robotic vision, machine learning, intelligent systems (XR, vision for HCI), etc. 
\vspace{-50pt}
\end{IEEEbiography}

\end{document}


\title{
Deep Learning for Event-based Vision: \\A Comprehensive Survey and Benchmarks \\
–Supplementary Material–
}
\maketitle

\IEEEdisplaynontitleabstractindextext

\IEEEpeerreviewmaketitle
\begin{abstract}
Due to the spatial constraints within the main paper, this supplementary material provides an expansive elucidation of the survey. 
Sec.~\ref{sec1} offers details and exact obtaining steps of the mentioned event representations of the main paper, including image-based and surface-based event representations.
Sec.~\ref{sec2} provides more potential new directions for event cameras, such as federated learning, adversarial robustness, and sparse computing.
Sec.~\ref{sec3} dedicates the datasets and evaluation metrics for deep learning methods in event-based vision.
Sec.~\ref{sec4} gives more details of qualitative and quantitative comparisons for a comprehensive overview the representative works, including image reconstruction, deblur, denoising, object classification, optical flow estimation, and depth estimation.
Sec.~\ref{sec5} provides more qualitative comparison including event super-resolution and high dynamic range.
\end{abstract}

\section{Details of some event representations}
\label{sec1}

\begin{figure*}[]
\centering
\includegraphics[width=.98\textwidth]{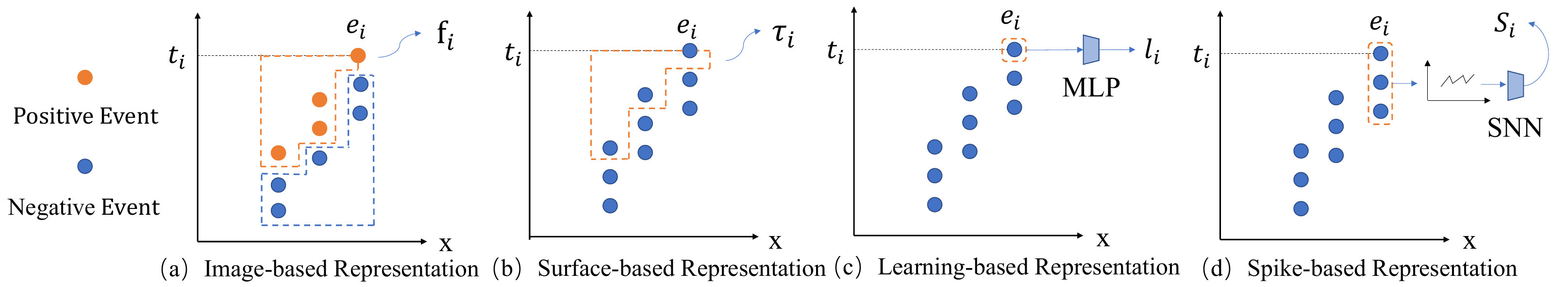}
Typical event representation methods for DNNs: (a) image-based representation $f_i$, (b) surface-based representation $\tau_i$, (c) learning-based representation $l_i$, (d) spike-based representation $S_i$. The events surrounded by yellow dashed lines refer to the output representations. We use negative events in (b), (c), and (d) to show the procedure while it is the same for the positive ones. \label{fig:evenr-representation}
\vspace{-10pt}
\end{figure*}

An event camera responds asynchronously to each independent pixel and generates a stream of events. 
An event is interpreted as a tuple  $(\textbf{u}, t, p)$, which is triggered whenever a change in the logarithmic intensity $L$ surpasses a constant value (threshold) $C$, formulated as follows: 

\begin{equation}
    p = \left\{
    \begin{aligned}
         +1 ,& L(\textbf{u},t) -L(\textbf{u}, t-\Delta t) \geq C \\
         -1 ,& L(\textbf{u},t) -L(\textbf{u}, t-\Delta t) \leq -C \\
         0  ,& other\\
    \end{aligned}
    \right.
\end{equation}

where $\textbf{u}=(x,y)$ denotes the pixel location, $t$ is the timestamp and $p\in \{-1, 1\}$ is the polarity, indicating the sign of brightness changes (-1 and 1 represent positive and negative events, respectively). $\Delta t$ is a time interval since the last event at pixel $\textbf{u}=(x,y)$. A stream of events are triggered 
which can be denoted as: 

\begin{equation}
    \mathcal{E}=\{e_i\}_{i=1}^N = \{\textbf{u}_i,t_i,p_i\}, i \in N,
\end{equation}
\subsection{Image-based Event Representation}
As the event camera's output is a sparse event stream, if we perform asynchronous event-by-event processing on the event stream, each event triggers pixel-wise computation, this makes the processing computationally intensive and generates features that focus more on local information but ignore global information. Therefore, in order to benefit from conventional processors based on well-established deep learning algorithms, an image-based representation that converts asynchronous events into synchronous frames has emerged in event-based vision research, as shown in Fig.~\ref{fig:evenr-representation} (a).

Most existing image-based approaches construct multi-channel frames as event representations to aggregate information including spatial coordinates, timestamps and polarities.
These methods(\cite{maqueda2018event,zhu2018ev,EST,deng2020amae,bai2022accurate,deng2021mvf} )often set channels to transform a set of events into frames based on the event's polarities (positive and negative), timestamps and event-counts. Maqueda \etal \cite{maqueda2018event} set up two separate channels to evaluate the histograms for positive and negative events to obtain two-channel event images that were finally merged together into synchronous event frames. To consider the importance of event counts and timestamps for holistic information, Zhu \etal \cite{zhu2018ev} proposed EV-FlowNet to provide a self-supervised neural network by generating four-channel images. The four channels contain two event-count channels and two timestamps of the most recent event channels based on the different event polarities. Gehrig \etal \cite{EST} took the coordinates and timestamps of the event as input to a multilayer perceptron (MLP), and repeatedly evaluated and summed the activation maps for each point to obtain a grid-like representation. But these approaches ignore the fact that different trajectories of the same object will have different semantic information. In such cases, Deng \etal \cite{deng2020amae} used timestamps of different polarities of events to form a two-channel image. Meanwhile, they assigned a higher value to recent events to avoid recent semantic information being blurred caused by overlapping motion messages. However, weighting timestamps with different values may result in corrupting the temporal information of the event stream. And in view of solving the outlier value due to the environment or hardware, Bai \etal \cite{bai2022accurate} proposed a three-channel event representation (one event-count channel and two temporal channels for positive and negative events). Deng \etal \cite{deng2021mvf} generated the three-channel images (x-y, x-t and y-t) by projecting event streams at the x, y, and t axes. Specifically, according to \cite{zihao2018unsupervised}, we can obtain the x-y frame by the following equation:
\begin{equation}
    \mathbb{F_\mathrm{xy(x,y,t)} }=\sum_{i}^{N}p_{i}\delta (x-x_{i},y-y_{i} ) max(0,1-\left |t-t_{i}^{\ast }   \right | ),
\end{equation}
where $\delta$ stands for $Dirac$ $delta$ function and $t_{i}^{\ast }$ represents the normalized event timestamp. The generated $\mathbb{F_\mathrm{(x,y)} }$ is in $\mathbb{R^{\mathrm {H\times W\times B_{t} } } } $. Deng \etal\cite{deng2021mvf} proposed to obtain $\mathbb{F_\mathrm{(x,t)} }$ and  $\mathbb{F_\mathrm{(y,t)} }$ by the similar integration procedure. However, due to the huge value discrepancy between $x-t$ and $y-t$, we should first scale the timestamps $t$ to corresponding ranges([0,$T_{xt}$-1] or [0,$T_{yt}$-1]).
So,take $\mathbb{F_\mathrm{xt(x,t,y)} }$ as an example , the scaled timestamps is 
\begin{equation}
    t_{i}^{\diamond }=\Gamma ((T_{xt}-1)(t_{i}-t_{0}  )/(t_{N-1}-t_{0}  ) ,
\end{equation}
where $\Gamma$ is the floor operation and the corresponding normalized coordinate of the y-axis is:
\begin{equation}
    y_{i}^{\diamond }=(B_{xt}-1)y_{i}/W .
\end{equation}
The final $s-t$ frame generation formulation is:
\begin{equation}
    \mathbb{F_\mathrm{xt(x,t,y)} }=\sum_{i}^{N}p_{i}\delta (x-x_{i},t-t_{i}^{\diamond }  ) max(0,1-\left |y-y_{i}^{\diamond }   \right | ) .
\end{equation}
This is a more comprehensive and complementary spatiotemporal representation compared to the single view (only generate x-y frame) methods discussed aforementioned. 

However, the current treatment of outlier values is too simple and brutal, as the threshold value is set and the higher value is changed. This would be a point that could be optimised.

\subsection{Surface-based Event Representation}
\begin{figure}[t!]
    \centering
    \includegraphics[width=0.46\textwidth]{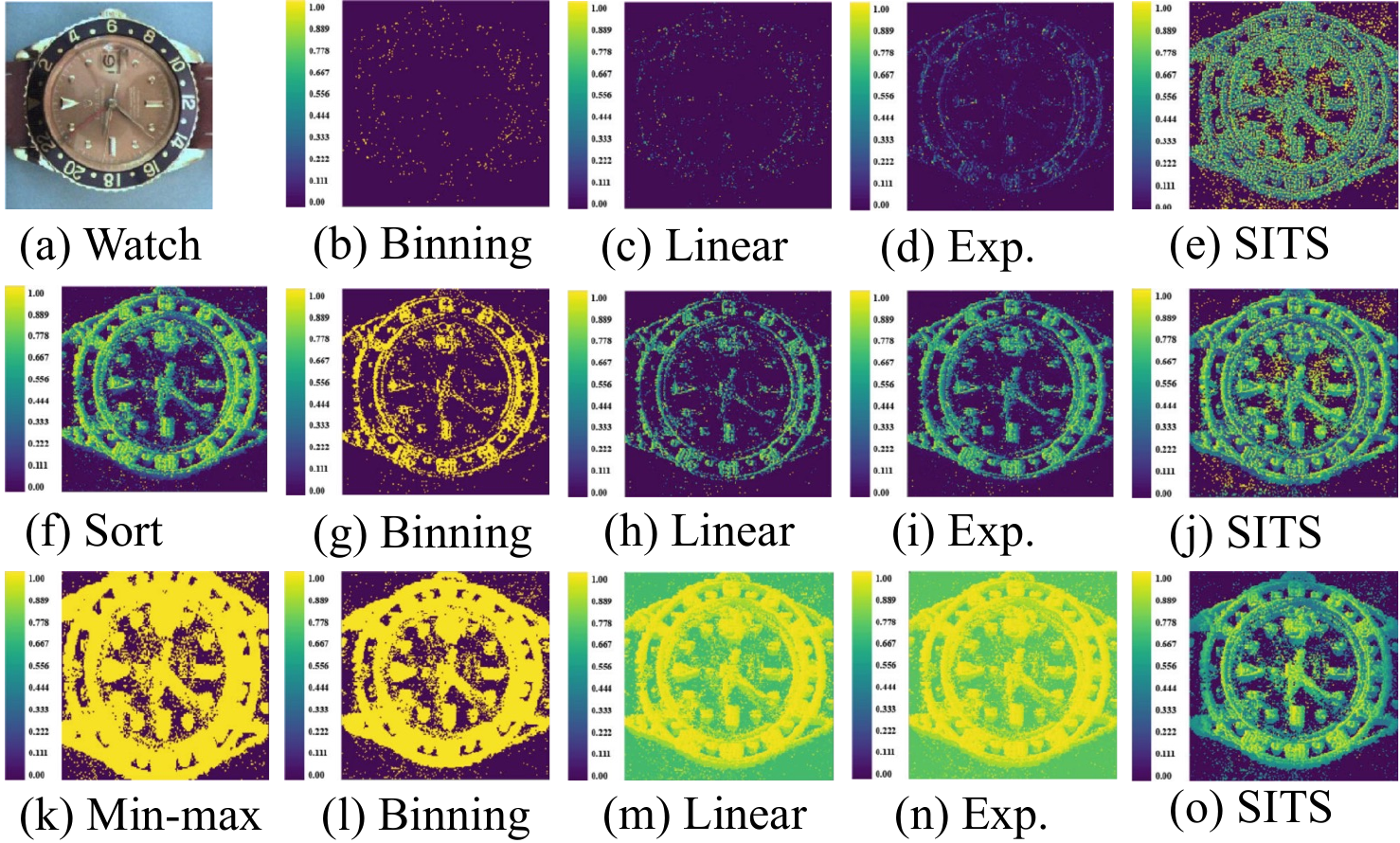}
    \caption{Visualization comparison of normalization approaches~\cite{ChainSAE}.}
    \label{fig:surface}
\end{figure}
The Surface of Active Event (SAE) is first proposed in ~\cite{SAE}, it maps the event streams to a time-dependent surface and tracks the activity around the spatial location of $ev_i$, as illustrated in Fig.~\ref{fig:evenr-representation} (b). The SAE which maps the timestamps of the latest event triggered in the neighbourhood region can be briefly formulated as:
\begin{equation}
    SAE: \mathbb{R}^2 \rightarrow [0,\infty \}.
\end{equation}
Different from the basic image-based representation which utilizes intensity images to provide context content, the SAE achieves this through a totally different perspective, \ie, the temporal-spatial perspective. Specifically, the time surface of the $i$ th event $ev_i$ can be formulated as a spatial operator acting on the neighbouring region of $e_i$:
\begin{equation}
\label{SAE}
    \tau_i([x_n,y_n]^T,p) = \underset{j\le i}{max}\{t_j|[x_i+x_n,y_i+y_n],p_j=p\} 
\end{equation}
where $x_n \in \{-r,r\}$, $y_n \in \{-r,r\}$, $p \in \{-1,1\}$ and $r$ is the radius of the neighbouring region used to obtain the time surface. As shown in Eq.~\ref{SAE}, the time surface $\tau_i([x_n,y_n]^T,p)$ encodes the time context in the $(2r+1)\times(2r+1)$ neighbourhood region of the incoming event $ev_i$, hence maintaining both temporal and spatial information for downstream tasks. 

As discussed before, the SAE captures the spatial-temporal characteristics of event streams. However, the monotonically increasing timestamps of events result in time surface values ranging from 0 to infinity~\cite{ChainSAE}.
Therefore, appropriate normalization approaches are required to help preserve the temporal invariant data representation from raw SAE by mapping the timestamps to $[0,1]$. According to the widely adopted normalization methods in image-based images, min-max normalization is also can be utilized in the event-based vision to map the timestamps of SAE to $[0,1]$~\cite{alzugaray2018ace}. Since the ranges of SAE are from 0 to infinity, the min-max normalization makes the current pattern information indistinguishable, as shown in (Fig.~\ref{fig:surface} (k)). To solve the indistinguishable problems, Afshar \etal ~\cite{time-window} proposed to use time-window normalization with event binning and linear decay. Specifically, the event binning changes all the values to 1 within a fixed time window $\delta t$, and other values are set to 0 as follows:
\begin{equation}
    B_i([x_i,y_i]^T, p_i)=\begin{cases}
  &1,t_i - \tau_i([x_n,y_n]^T,p_i) \le \delta t \\
  &0, other,
\end{cases}
\end{equation}
where the $B_i([x_i,y_i]^T, p_i)$ is the normalized surface of $ev_i$.
Linear decay improves the mapping by using a linear function to obtain the normalized surface $L_i([x_i,y_i]^T, p_i)$ as: 
\begin{equation}
    L_i([x_i,y_i]^T, p_i)=\begin{cases}
  &1 + \delta t(\tau_i - t_i), t_i - \tau_i \ge \frac{1}{\delta t}  \\
  &0, other,
\end{cases}
\end{equation}
where $\tau_i$ is the abbreviation of $\tau_i([x_n,y_n]^T,p_i)$. As shown in Fig.~\ref{fig:surface}, an inappropriate parameter leads to information loss. A too-small time interval makes the pattern unrecognizable as illustrated in Fig.~\ref{fig:surface} (b), (c) and (d). By contrast, a too-large parameter also introduces undistinguishable problems in Fig.\ref{fig:surface} (l), (m) and (n). Only the properly tuned parameter approximately brings the pattern (watch) back in Fig.~\ref{fig:surface} (b), (h) and (I).

Besides the direct and the linear function mapping, there is also an exponential decay normalization described in ~\cite{Hots, N-HAR}. 
It uses a non-linear exponential function to generate the normalized surface $E_i([x_i,y_i]^T, p_i)$ within $\delta t$ by:
\begin{equation}
    E_i([x_i,y_i]^T, p_i)=e^{\delta t(\tau_i([x_n,y_n]^T,p_i) - t_i)},
\end{equation}
where $\delta t$ is a time constant that requires parameter tuning. 
All these above normalization methods rely on empirical parameter tuning, leading to additional computational costs. To avoid this problem, sort normalization is proposed by Alzugaray \etal ~\cite{alzugaray2018ace} to sort all the timestamps within an SAE at each pixel. After the sorting, the minimum and maximum values of the SAE can be utilized to map the SAE to $[0,1]$. However, though the sort normalization method alleviates the dependence on parameter tuning, the by-product of time complexity impedes the whole procedure's efficiency.

In order to build an efficient SAE and achieve robust speed invariant characteristics, Manderscheid \etal ~\cite{SITS} introduced a normalization scheme to obtain the Speed Invariant Time Surface (SITS). The normalization scheme of SITS initializes all the pixel locations $\tau_i([x_i+x_n,y_i+y_n]^T,p_i)$ in $ev_i$'s neighbourhood region to 0. For each incoming event $ev_i$, all event locations $[x_i+x_n,y_i+y_n]^T$ are considered:
\begin{equation}
    \begin{aligned}
&\tau_i([x_i+x_n,y_i+y_n]^T,p_i)=\tau_i([x_i+x_n,y_i+y_n]^T,p_i)-1, \\
  & \tau_i([x_i+x_n,y_i+y_n]^T,p_i) \ge \tau_i([x_i,y_i]^T,p_i).
\end{aligned} 
\end{equation}
After the operation of all the events in the neighbourhood region, finally, the $SITS_i([x_i+x_n,y_i+y_n]^T,p_i)$ are set to $(2r+1)^2$. Thus the range of $SITS_i$ is from 1 to $(2r+1)^2$. Obviously, as shown in Fig~\ref{fig:surface} (e), (j) and (o), the SITS is influenced by the constant $r$ and an inappropriate value will lead to insufficient representation. So it is critical to find the trade-off between time consumption and robustness by fine-tuning the constant $r$. Overall, the SITS updates the time surface when a new event is triggered, thus leading to inefficiency in the on-demand tasks. Lin \etal ~\cite{ChainSAE} suggested solving and alleviating this imbalance between normalization and the number of events by using a chain update strategy. 
The proposed chain SAE method demonstrates superior efficiency compared to other normalization methods, as depicted in Fig.~\ref{fig:surface} (f). Notably, it eliminates the need for parameter tuning, further enhancing its practicality and ease of implementation.
\begin{table*}[t!]
\centering
\caption{Some mainstream datasets for event-based vision. - means no results available.}
\label{tab:Datasets}
\resizebox{0.99\textwidth}{!}{
\begin{tabular}{ccccc}
\toprule
Dataset  & Description &  Tasks & Cameras & Metric \\
\midrule
\makecell[c]{HQF ~\cite{stoffregen2020reducing}} & \makecell[c]{Events and ground truth frames} & Image Reconstruction  & DAVIS240C & - \\
\hdashline
\makecell[c]{DR ~\cite{delmerico2019we}} & \makecell[c]{Indoor and outdoor environment} & Image Reconstruction  & miniDAVIS346 & - \\
\hdashline
\makecell[c]{DAVIS240C~\cite{mueggler2017event}} & \makecell[c]{Events and ground-truth, intensity frames and IMU} & Image Reconstruction and SLAM & DAVIS+Simulator & SSIM \\
\midrule
\makecell[c]{DHP19~\cite{DHP19}} & \makecell[c]{33 movements recorded \\from 17 subjects} & 3D Human Pose Estimation  & DVS cameras & MPJPE \\
\hdashline
\makecell[c]{EventCap~\cite{EventCap}} & \makecell[c]{12 sequences of 6 actors \\performing different activities} & 3D Human Pose Estimation & DAVIS240C & MPJPE \\
\midrule
\makecell[c]{DDD17~\cite{ESS}} & \makecell[c]{12 hours of driving data} & Semantic Segmentation & DAVIS & mIOU  \\
\hdashline
\makecell[c]{DDD20~\cite{hu2020ddd20}} & \makecell[c]{ 51h of DAVIS event+frame camera and vehicle human control data \\ collected from 4000 km of highway and urban driving} & Semantic Segmentation & DAVIS & mIOU  \\
\hdashline
\makecell[c]{DSEC-Semantic~\cite{ESS}} & \makecell[c]{53 driving sequences collected in a variety of \\urban and rural environments in Switzerland~\cite{Dsec}} & Semantic Segmentation & DAVIS & mIOU  \\
\hdashline
\makecell[c]{EVIMO2~\cite{burner2022evimo2}} & \makecell[c]{3.75 minutes of independently moving household objects, \\22.55 minutes of static scenes, \\and 14.85 minutes of basic motions in shallow
scenes} & Semantic Segmentation & Prophesee Gen3 VGA & mIOU  \\
\midrule
\makecell[c]{KITTI Simulated~\cite{KITTI}} & \makecell[c]{20 videos for object tracking which are converted\\ from videos to events using the V2E simulator} & Object Detection & Simulator & mAP  \\
\hdashline
\makecell[c]{Gen1 Automative~\cite{Gen1_Auto}} & \makecell[c]{226719 cars and 27658 pedestrians \\with manual bounding box annotations} & Object Detection & ATIS & mAP  \\
\hdashline
\makecell[c]{1Mpx Detection~\cite{1MpxDetection}} & \makecell[c]{event streams and more than 25 million bounding\\ boxes of cars, pedestrians and two-wheelers} & Object Detection & 1 megapixel event camera\\+GoPro
Hero6 & mAP  \\
\hdashline
\makecell[c]{PKU-DDD17-CAR~\cite{PKUDDD17}} & \makecell[c]{discontinuous images and event streams \\of driving scenes} & Object Detection & DAVIS346 & mAP  \\
\midrule
\makecell[c]{N-Cars~\cite{sironi2018hats}} & \makecell[c]{4, 029 event samples for \\the binary task of car recognition}  & Object Recognition & ATIS & 
Accuracy \\
\hdashline
\makecell[c]{N-Caltech101~\cite{orchard2015converting}} & \makecell[c]{8, 246 samples and 100 classes\\ based on Caltech101 dataset \cite{fei2006one}} & Object Recognition & ATIS & 
Accuracy \\
\midrule

\makecell[c]{ENFS-real ~\cite{duan2021eventzoom}} & \makecell[c]{First real-world dataset involving \\multi-scale LR-HR pairs for event SR} & Super-resolution &  AUO80e+DAVIS34& RMSE \\
\hdashline
\makecell[c]{ENFS-syn ~\cite{weng2022boosting}} & \makecell[c]{Involves 2(4, 8, 16)× \\LR-HR pairs for 65 scenes} & Super-resolution & Event Simulator & RMSE \\
\hdashline
\makecell[c]{RGB-DAVIS-syn ~\cite{weng2022boosting}} & \makecell[c]{Similar to ENFS-syn \\that utilizes the HR frames from RGB-DAVIS} & Super-resolution & Event Simulator & RMSE \\
\hdashline
\makecell[c]{DVSNOISE20 ~\cite{baldwin2020event}} & \makecell[c]{contains 16 different scenes \\mostly under static conditions} & Denoising \& Super-resolution & DAVIS346 & IoU \\
\hdashline
\makecell[c]{DVSCLEAN ~\cite{fang2022aednet}} & \makecell[c]{First real dataset shares the same \\recording method with simulated dataset} & Denoising & CeleX-V & SNR \\
\hdashline
\makecell[c]{DND21 ~\cite{guo2022low}} & \makecell[c]{The only one based on clean DVS plus\\ realistic DVS leak and shot noise} & Denoising & DAVIS346+Event Simulator & AUC \\
\hdashline
\makecell[c]{IEBCS ~\cite{joubert2021event}} & \makecell[c]{An extended DVS pixel simulator which simplifies \\ the latency and the noise models} & Denoising & Event Simulator & Accuracy \\
\hdashline
\makecell[c]{v2e ~\cite{hu2021v2e}} & \makecell[c]{Including pixel level Gaussian\\ event threshold mismatch,\\ finite intensity dependent bandwidth, \\and intensity-dependent noise} & Denoising & Event Simulator & Accuracy \\
\hdashline
\makecell[c]{ESIM~\cite{wang2020eventsr}} & \makecell[c]{Multiple synthetic events and APS images} & Image Super-resolution  & Event Simulator & PSNR / FSIM / SSIM \\
\midrule
\makecell[c]{MVSEC ~\cite{Zhu2018TheMS}} & \makecell[c]{consists of day/night-time outdoor driving sequences\\and flying indoor sequences, with depth GT} & Depth Estimation & DAVIS &  AAE / MAE\\
\hdashline
\makecell[c]{DSEC ~\cite{Gehrig2021DSECAS}} & \makecell[c]{contains 53 driving scenarios \\taken in various lighting conditions} & Depth Estimation &  Stereo RGB and event camera & MAE / RMSE\\
\hdashline
\makecell[c]{VIVID ~\cite{lee2022vivid++}} & \makecell[c]{contains sequences for
visual navigation \\ in poor illumination condition} & Depth Estimation &  DAVIS 240 C & MAE / RMSE\\
\midrule
\makecell[c]{VECtor ~\cite{gao2022vector}} & \makecell[c]{benchmark datasets for research on \\multi-sensor SLAM including data from\\an event stereo camera, a stereo camera,\\ an RGB-D sensor, a LiDAR, and an IMU} & Visual Odometry and SLAM & Prophesee Gen3 VGA & -\\
\hdashline
\makecell[c]{TUM-VIE ~\cite{klenk2021tum}} & \makecell[c]{benchmark datasets for developing \\3D perception and navigation algorithms,\\including stereo events and grayscale frames,\\ IMU and 6dof motion data} & Visual Odometry and SLAM & Prophesee Gen4 HD & -\\
\hdashline
\makecell[c]{EDS ~\cite{hidalgo2022event}} & \makecell[c]{A dataset released for direct \\monocular visual odometry, including \\ data from an event camera\\ and a RGB camera} & Visual Odometry and SLAM & Prophesee EVALUATION \\
\hdashline
\makecell[c]{Stereo DAVIS~\cite{event_slam_pic2}} & \makecell[c]{Stereo events data and their \\corresponding grayscale frames, an IMU data} & Visual Odometry and SLAM & DAVIS+Simulator & - \\
\bottomrule
\end{tabular}
}
\end{table*}

\section{More New Directions}
\label{sec2}
\noindent\textbf{\textit{Federated Learning for Event Data Privacy:}}
The traditional DL-based visual training is to collect data and then conduct unified training. This pipeline in event-based vision faces three challenges: 1) Collecting large-scale event datasets is extremely expensive; 2) The data collection may violate the privacy of individuals; 3) The cycle of centralized collection, centralized training, and re-release of the model is time-consuming, which is not conducive to rapid iteration. The introduction of federated learning can solve these challenges. For example, federated learning makes it possible to iterate the model at each terminal node. It is not the data itself that is passed into the central node, but the gradient of the terminal node update. This can effectively alleviate the above three problems in event-based vision.

\vspace{2pt}
\noindent \textbf{Adversarial Robustness of Event-based DNNs Models:}
The research on the adversarial robustness of DNNs on the 2D and 3D data, \eg, canonical images and point cloud, have been extensively explored in the literature. However, little research has considered the adversarial robustness of DNNs for event-based vision. Recently, Marchisio \etal ~\cite{Dvs-attacks} first focused on this problem and proposed various categories of adversarial examples for SNNs. Also, Lee \etal ~\cite{lee2022adversarial} proposed to train robust DNNs with generated adversarial examples. As event cameras bring a distinct imaging shift, and deep learning for event-based vision attracts more and more attention, the robustness of event-based DNN models with respect to adversarial attack and defence is worth exploring in future research.

\vspace{2pt}
\noindent \textbf{Sparse Computing:}
The utilization of event camera sparsity in deep neural networks (DNNs) and physical DNN inference engines presents opportunities to improve system throughput, latency, and energy efficiency. Recent advancements~\cite{mirabbasi2022through, liu2022energy, gao2022spartus} in weight, activation, and temporal sparsity have demonstrated significant advantages. Event cameras, with their activity-driven nature, allow for the exploitation of sparsity in computational processes. By updating the network only when necessary, event cameras effectively reduce computations, particularly for tasks involving temporal changes in input. This approach, known as input activity-driven computing, in conjunction with other forms of sparsity, holds the potential to enable highly energy-efficient edge devices that emulate the efficiency observed in the brain operating in natural environments. Further investigation and exploration of sparsity computing with event cameras can enhance the performance of DNN systems across various domains, including wearables, brain-machine interfaces, and IoT applications.

\section{Dataset and evaluation metric}
\label{sec3}
Datasets are pivotal for building deep learning methods for event-based vision.  In Tab.~\ref{tab:Datasets}, we provide an overview of representative datasets and standard evaluation metrics for different visual tasks in the context of deep-learning-based event-based vision. It is important to note that certain datasets are excluded as they consist of synthetic events generated using the widely-used video-to-event simulator ESIM~\cite{rebecq2018esim}, which utilizes RGB or image-based datasets as input.

\section{Details of some qualitative and quantitative comparisons}
\label{sec4}
In order to provide a comprehensive overview of the representative work in this section, we have expanded certain summary tables pertaining to various visual tasks, such as image reconstruction, deblurring, denoising, object classification, optical flow estimation, and depth estimation. However, due to space constraints, we encourage readers to refer to the original paper for detailed information and specific experimental details regarding these studies.

\subsection{Image Reconstruction}
\subsubsection{Experiment Setting and Dataset}
During the evaluation of their approaches, the authors~\cite{zhang2022formulating} conducted comparisons between their methods and state-of-the-art (SoTA) image reconstruction techniques using standard datasets. They specifically mention E2VID~\cite{Rebecq2019HighSA}, ECNN~\cite{stoffregen2020reducing}, and BTEB~\cite{paredes2021back} as the approaches they compare against.

The comparison is performed by reconstructing images at the timestamps corresponding to ground truth images from datasets such as DAVIS frames. The evaluation metrics employed encompass mean square error (MSE), structural similarity (SSIM), and perceptual similarity (LPIPS). The evaluation sequences consist of 60-second durations with varying motion speeds. To prevent motion blur corruption, the authors adhere to the sequence cuts suggested by~\cite{stoffregen2020reducing} and selectively utilize timestamps within a specific range.

In the case of the slider sequences, image reconstruction is performed at all available timestamps of the frames, as minimal motion blur is present. The authors use a certain number of events per individual waveform event (IWE) and estimate motion using contrast maximization techniques. Optical flow is computed from the estimated motion and fed into their method.

The authors provide information about the range of values for the regularizer weight in various regularization techniques, including Tikhonov regularization (Tikh.), total variation (TV), and image prior to denoiser.

\begin{table}[t!]
\centering
\caption{\textbf{Tab. 2 in the main paper.} Qualitative comparison results of some image reconstruction methods~\cite{zhang2022formulating} on event dataset~\cite{mueggler2017event}.}
\resizebox{0.95\linewidth}{!}{
\begin{tabular}{cccccc}
\toprule
Method & Type & MSE~$\downarrow$ & SSIM~$\uparrow$ & LPIPS~$\downarrow$ & Time\\ \midrule
E2VID ~\cite{Rebecq2019HighSA} & DL-based & 0.069 & 0.395 & 0.438 & 0.2448 s\\
ECNN~\cite{stoffregen2020reducing} & D-based & 0.056 & 0.416 & 0.442 & 0.2839 s \\
BTEB~\cite{paredes2021back} & DL-based & 0.090 & 0.357 & 0.520 & 0.4059 s\\
Tikhonov ~\cite{zhang2022formulating} & Model-based &  0.121 & 0.356 & 0.485 & 0.4401 s\\
TV~\cite{zhang2022formulating} & Mode-based & 0.113 & 0.386 & 0.502 & 4.0443 s\\
CNN~\cite{zhang2022formulating} & DL-based & 0.080 & 0.437 & 0.485 & 28.3904 s\\
\bottomrule
\end{tabular}}
\label{Reconstruction}
\end{table}

\subsubsection{Experimental Analysis}
\begin{figure*}[t!]
    \centering
    \includegraphics[width=0.9\textwidth]{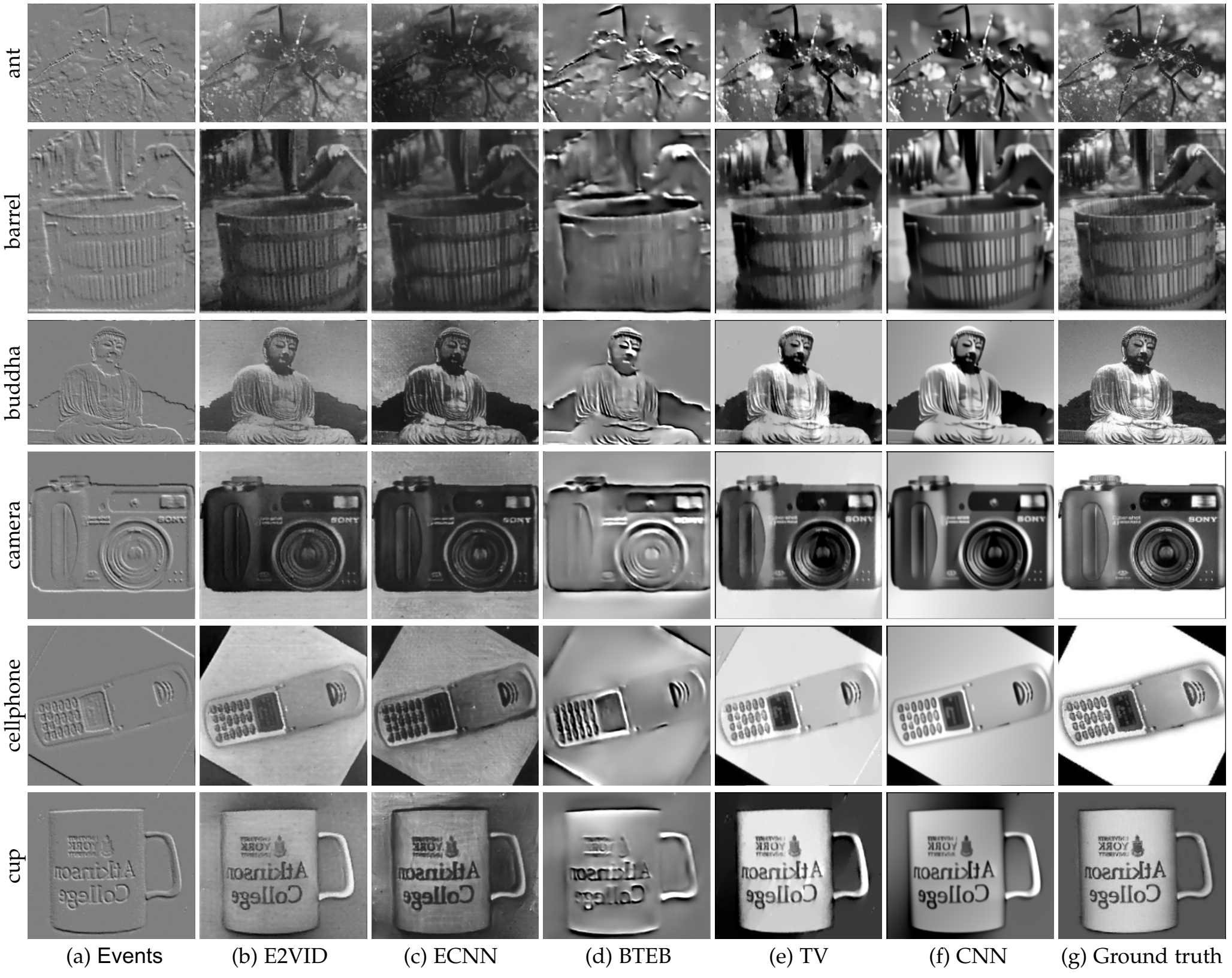}
    \caption{Qualitative comparison of image reconstruction methods from~\cite{zhang2022formulating}.}
    \label{fig: Reconstruction}
\end{figure*}

The authors present qualitative and quantitative results for several methods in terms of evaluation metrics, which are provided in Fig.~\ref{fig: Reconstruction} and Tab.~\ref{Reconstruction}. They highlight that the choice of evaluation metric and protocol significantly impact the reported results.

Among the methods evaluated, the authors' method achieves the best results in terms of SSIM, while E2VID and ECNN perform better in other metrics. It is noteworthy that the authors' three methods (Tikh., TV, and CNN) achieve reconstruction results that are comparable to the SoTA methods, despite not relying on ground truth images and lacking recurrent connections to past events beyond the individual waveform event (IWE). This highlights the effectiveness of their approaches and suggests that even without certain advantages typically associated with more advanced techniques, such as ground truth information or recurrent connections, their methods are able to achieve competitive reconstruction quality.

Remarkably, traditional and straightforward techniques like classical Tikhonov or total variation regularizers, which operate on linear systems of equations, demonstrate comparable results to more intricate learning-based approaches. Particularly in scenes with dynamics and complex shapes, the authors' proposed CNN regularizer exhibits superior performance compared to ECNN in terms of SSIM and perceptual similarity (LPIPS) distributions. The distributions of the proposed regularizers (Tikh.-TV-CNN) consistently exhibit trends in mean square error (MSE) and SSIM, although the trend is less distinct when considering LPIPS. These findings highlight the impressive efficacy of even simple regularization techniques in achieving reconstruction quality comparable to more complex learning-based methods.

Overall, the authors provide an analysis of the quantitative results obtained using different metrics, highlighting the performance of their methods compared to the SoTA methods and emphasizing the effectiveness of even simple regularization techniques in achieving comparable reconstruction quality.

\subsection{Deblur}
\subsubsection{Experiment Setting and Dataset}
To generate a synthetic dataset for the GoPro dataset, the authors employed the REDS dataset as the foundation. Initially, the images were downsampled and cropped to a size of 160 × 320. To increase the frame rate, RIFE was utilized to interpolate 7 additional frames between each pair of consecutive frames. For the purpose of creating blurry frames and events, sequences with a high frame rate were utilized. The blurry frames were generated by averaging a specific number of sharp images, while the events were simulated using ESIM. This process resulted in a purely synthetic dataset that emulates the characteristics of the GoPro dataset.

In contrast to the synthetic nature of the GoPro dataset, the HQF dataset comprises real-world events and high-quality frames captured simultaneously using a DAVIS240C camera. The images within this dataset exhibit minimal blur. Similarly to the GoPro dataset, the frame rate of the HQF dataset is increased by up-converting the frames, and blurry frames are synthesized using the same methodology. Through this process, a semi-synthetic dataset of blurry videos is generated, incorporating both real-world events and synthesized blur effects.

The EVDI method adopts a lightweight network architecture, comprising a 5-layer LDI network and a fusion network composed of 6 convolution layers, 2 residual blocks, and 1 CBAM (Convolutional Block Attention Module) block. The implementation is carried out using PyTorch and trained on NVIDIA GeForce RTX 2080 Ti GPUs, with a batch size of 4. The training process consists of two stages: deblurring and unified deblurring with interpolation, each lasting for 100 epochs. Separate models are trained for each dataset. Importantly, the evaluation of the EVDI method does not rely on ground-truth images, showcasing its ability to perform assessment without reference data.

\subsubsection{Experimental Analysis}

\begin{figure*}[t!]
    \centering
    \includegraphics[width=\textwidth]{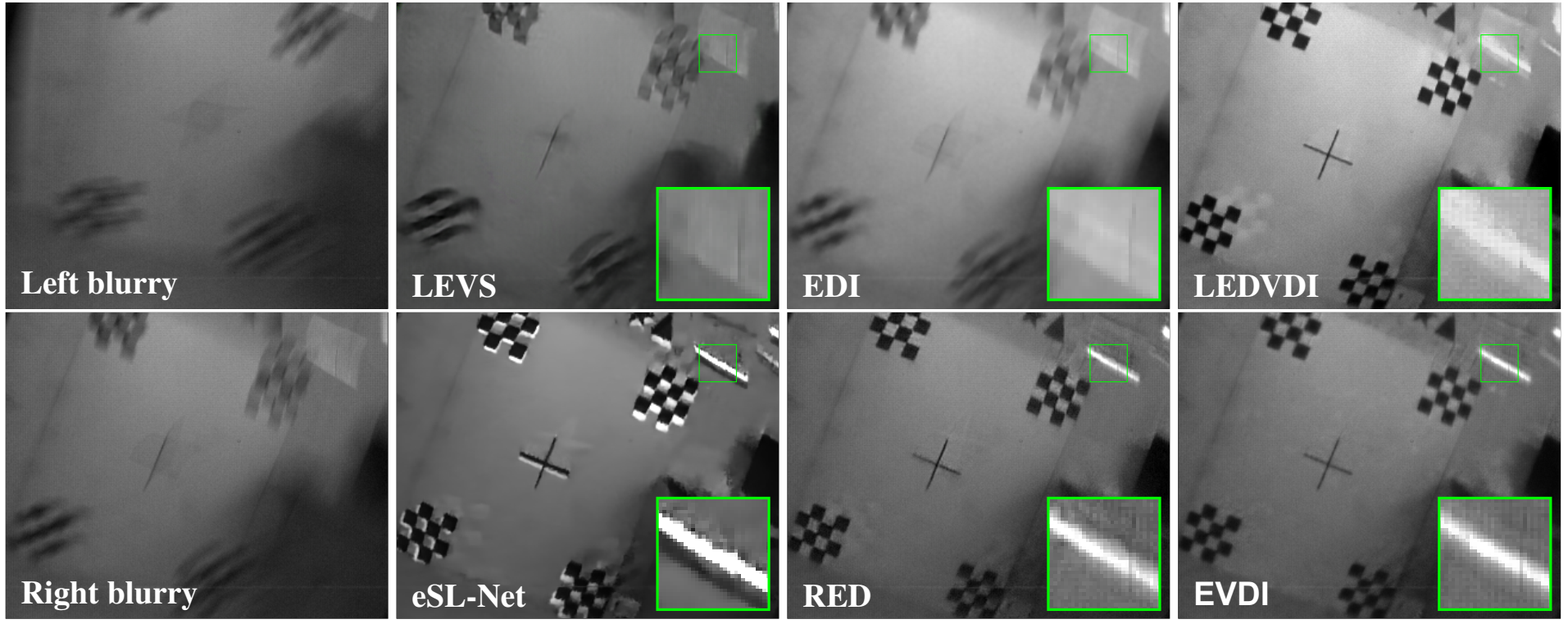}
    \caption{Visualization results of deblurring from EVDI~\cite{zhang2022unifying}.}
    \label{fig: deblur}
\end{figure*}

The qualitative and quantitative results of the deblurring task are presented in Fig~\ref{fig: deblur} and Tab.~\ref{Deblurring}, respectively. The EVDI method showcases impressive deblurring performance when compared to SoTA approaches. While LEVS exhibits limitations in highly dynamic scenes, EDI and LEDVDI achieve competitive results by effectively utilizing precise motion information extracted from events. However, it is worth noting that the performance of LEDVDI varies across different datasets. RED achieves competitive results through semi-supervised learning, although it compromises some performance in order to balance different data distributions. In contrast, the EVDI method tackles this challenge by fitting the specific data distribution using a self-supervised framework, resulting in superior performance on each dataset. Furthermore, the EVDI model stands out for its significantly fewer network parameters and overall computational efficiency compared to other methods, with the exception of eSL-Net.

\begin{table}[t!]
\centering
\caption{\textbf{Tab. 5 in the main paper.} Qualitative comparison of deblurring methods on GoPro and HQF dataset from ~\cite{zhang2022unifying}. `N/A' means no results are available.}
\vspace{-10pt}
\resizebox{0.94\linewidth}{!}{
\begin{tabular}{ccccccccc}
\toprule
\multirow{2}{*}{Method} & \multicolumn{3}{c}{GoPro} & \multicolumn{3}{c}{HQF} & \multirow{2}{*}{Param.}\\ \cmidrule{2-7}
& PSNR $\uparrow$ & SSIM $\uparrow$ & LPIPS $\downarrow$ & PSNR $\uparrow$ & SSIM $\downarrow$ & LPIPS $\uparrow$ \\ \midrule
LEVS ~\cite{jin2018learning} & 20.84 & 0.5473 & 0.1111 & 20.08 & 0.5629 & 0.0998 & 18.21M \\
EDI ~\cite{pan2019bringing} & 21.29 & 0.6402 & 0.1104 & 19.65 & 0.5909 & 0.1173 &  N/A\\
eSL-Net ~\cite{wang2020event} & 17.80 & 0.5655 & 0.1141 & 21.36 & 0.6659 & 0.0644 & 0.188M\\
LEDVDI ~\cite{lin2020learning} &  25.38 & 0.8567 & 0.0280 & 22.58 & 0.7472 & 0.0578 & 4.996M \\
RED~\cite{xu2021motion} & 25.14 & 0.8587 & 0.0425 & 24.48 & 0.7572 & 0.0475 & 9.762M\\
EVDI~\cite{zhang2022unifying} & 30.40 & 0.9058 & 0.0144 & 24.77 & 0.7664 & 0.0423 & 0.393M\\
\bottomrule
\end{tabular}}
\label{Deblurring}
\end{table}

\subsection{Denoising}
\subsubsection{Experiment Setting and Dataset}

For DL-based methods, researchers generated a simulated dataset with explicit labels to train the AEDNet~\cite{fang2022aednet}. Typically, simulating real-world data using simulated data can lead to inaccuracies due to different generation mechanisms. Nevertheless, AEDNet is specifically designed to capture the spatial affinity between neighboring events, a shared characteristic present in both simulated and real-world data. Additionally, real-world data was collected to demonstrate the usefulness of AEDNet in real-world scenarios. The dataset used for evaluation is called DVSCLEAN. In the experimental section, the researchers conducted tests on both a denoising dataset and an object classification dataset to evaluate the performance of AEDNet. Initially, the researchers compared the performance of AEDNet with other SoTA denoising algorithms (STP~\cite{huang20221000}, Liu et al.~\cite{liu2015design}, ICM~\cite{wu2020probabilistic}
and EDnCNN~\cite{baldwin2020event}) using the DVSCLEAN dataset. Since the simulated data in the dataset contained explicit labels, the denoising performance could be directly assessed. For real-world data, the researchers visualized the event stream to demonstrate the visual impact of the denoising process. Subsequently, they tested AEDNet on another publicly available dataset, DVSNOISE20~\cite{baldwin2020event}, to further validate its robustness.

\subsubsection{Experimental Analysis}
\begin{figure*}[h!]
    \centering
    \includegraphics[width=1.0\textwidth]{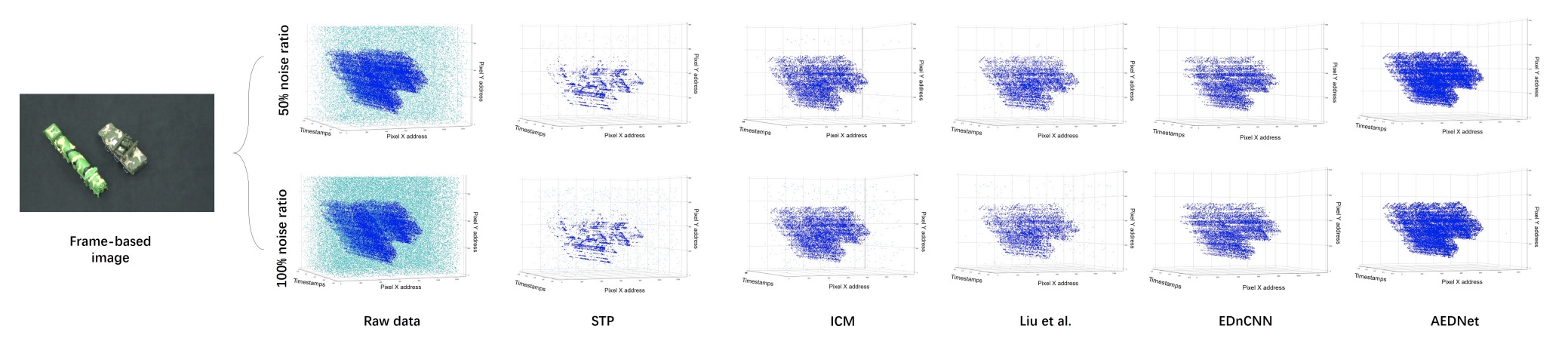}
    \caption{Comparative analysis of event stream denoising results for five algorithms using the simulated data from the DVSCLEAN dataset.
    The results are taken from ~\cite{fang2022aednet}.}
    \label{fig:Denoise}
\end{figure*}

AEDNet is benchmarked against other SoTA denoising algorithms. Firstly, the testing is conducted on simulated data, using Signal-to-Noise Ratio (SNR) as the metric to evaluate the denoising performance. The denoising results of the five algorithms on the simulated data are compared in Fig. \ref{fig:Denoise} and Tab. \ref{Denoising}. It can be observed that after denoising with AEDNet, there are significantly fewer isolated points remaining. However, STP, ICM, and Liu \etal still leave a significant amount of isolated noise. Furthermore, the remaining four algorithms encounter a common challenge of incorrectly identifying real events as noise, resulting in the removal of some real events, which can negatively impact subsequent tasks. In contrast, AEDNet exhibits consistent performance even under higher noise ratios, whereas the SNR of the other algorithms noticeably decreases with increasing noise ratio. This highlights the superior robustness of AEDNet in accurately identifying real events within complex scenes.

\begin{figure*}[h]
    \centering
    \includegraphics[width=1.0\textwidth]{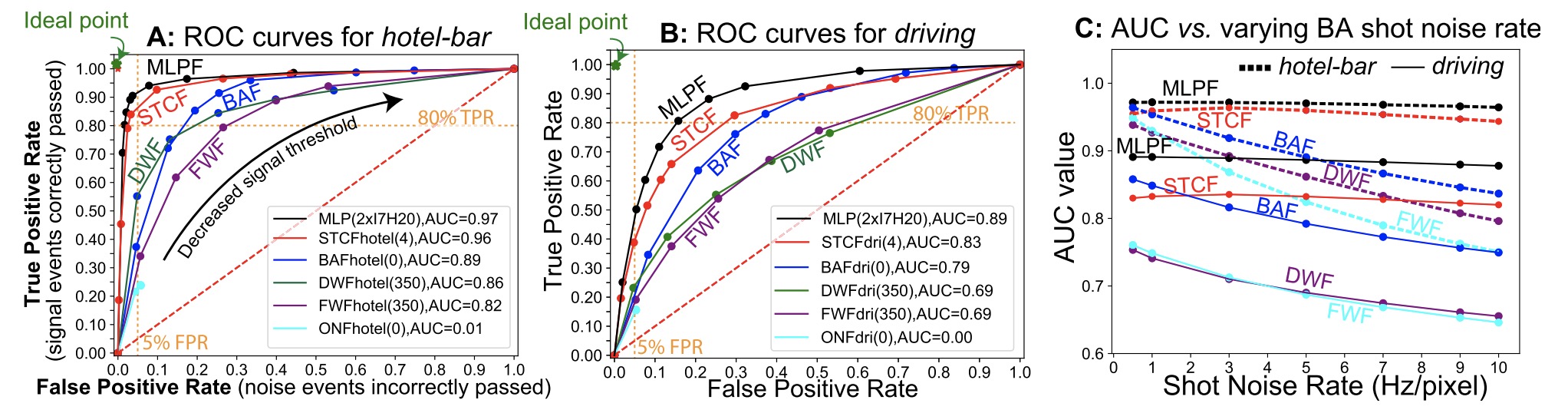}
    \caption{Performance evaluation of denoising algorithms using ROC curves and summarized AUC values on two datasets. The results are taken from ~\cite{guo2022low}.}
    \label{fig:ROC}
\end{figure*}

SNR measurement depends on the discrimination threshold. Overly aggressive denoising might remove useful signals, falsely inflating SNR. Instead, we use the Area Under the Curve (AUC) metric for a more comprehensive model assessment. AUC evaluates overall performance across various thresholds, offering an unbiased view unlike metrics focused on a single threshold.
We showcase ROC curves and AUC values for different denoising methods in Fig. \ref{fig:ROC}. Results for the hotel-bar and driving sequences are presented in Fig. \ref{fig:ROC} (A) and (B), respectively, with Fig. \ref{fig:ROC} (C) summarizing AUC accuracies against shot noise rates. Notably, the Multilayer Perceptron Denoising Filter (MLPF) outperforms handcrafted methods by efficiently filtering signal-like events, maintaining high AUC even as noise increases. Conversely, other filters' performance declines with higher noise.
These outcomes underscore MLPF and STCF's denoising efficacy and resilience against noise.

\begin{table}[h!]
\centering
\caption{Comparision of SNR scores of Denoising algorithms on DVSCLEAN dataset from ~\cite{fang2022aednet} .}
\begin{tabular}{cccc}
\toprule
\multirow{1}{*}{} &50\% noise ratio &  100\% noise ratio & Average \\
\midrule
\multirow{1}{*}{Raw data} & 3 & 0 & 1.5\\
\midrule
STP ~\cite{huang20221000} & 20.34 & 14.53 & 17.44 \\
Liu et al. ~\cite{liu2015design} & 23.80 & 18.70 & 21.25 \\
ICM ~\cite{wu2020probabilistic} & 21.64 & 15.68 & 18.66 \\
EDnCNN ~\cite{baldwin2020event}& 24.75 & 18.80 & 21.78\\
AEDNet ~\cite{fang2022aednet}  & 26.11 & 25.08 & 25.60 \\
\bottomrule
\end{tabular}
\label{Denoising}
\end{table}

\begin{figure*}[t!]
    \centering
    \includegraphics[width=\textwidth]{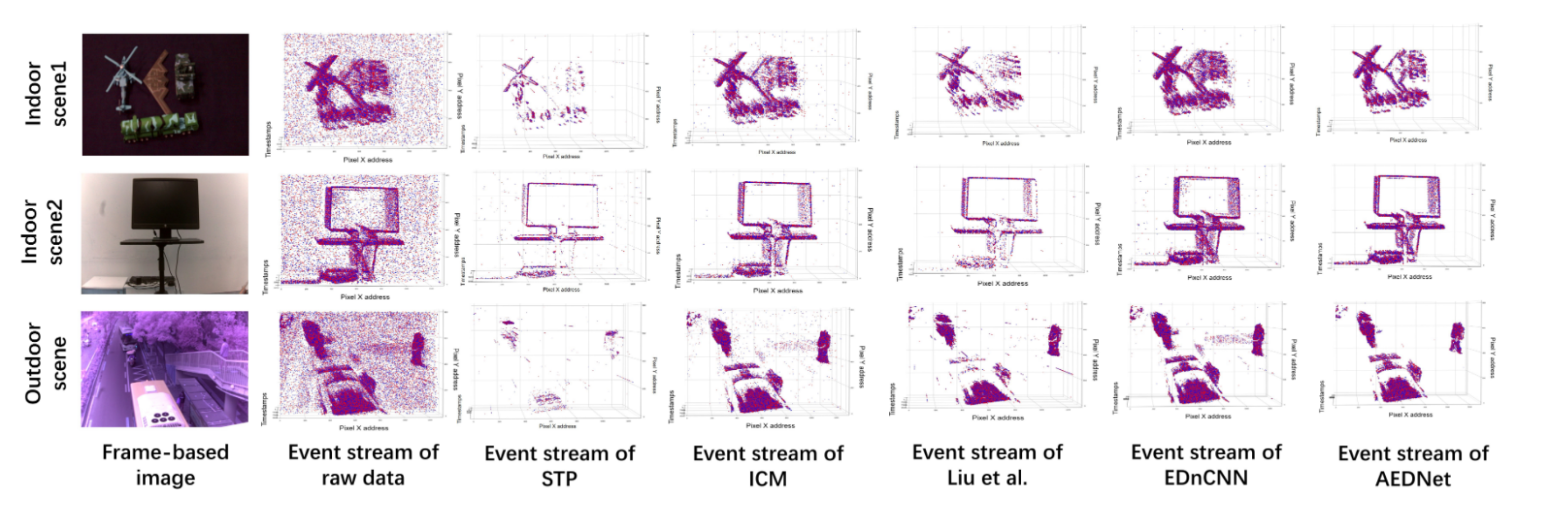}
    \caption{The visual comparisons among STP\cite{huang20221000}, ICM \cite{wu2020probabilistic}, Liu \etal \cite{liu2015design}, EDnCNN \cite{baldwin2020event}, AEDNet\cite{fang2022aednet} five denoising algorithms based on real-world data of the DVSCLEAN dataset.}
    \label{fig: DenoisingVisual_comparisons}
\end{figure*}

\subsection{Object Classification}
\subsubsection{Experiment Setting and Dataset}
\noindent \textbf{N-MINIST~\cite{orchard2015converting}} is a spiking version of the MINIST dataset which contains the same amount of training and testing samples (60000 and 10000, respectively) and has the same size as the original MNIST dataset (28 $\times$ 28 pixels). The N-MNIST dataset is captured by moving an event sensor while viewing the MNIST examples displayed on an LCD monitor.

\noindent \textbf{N-Caltech101~\cite{orchard2015converting}} dataset is a spiking version of the original Caltech-101 dataset. To avoid confusion, the "Faces" class in the original Caltech-101 is removed. Consequently, N-Caltech-101 consists of 100 object classes along with a background class. N-Caltech101 is also captured by monitor display and moving sensors.

\noindent \textbf{N-Cars~\cite{sironi2018hats}} dataset is a substantial event-based dataset especially designed for car classification in real-world scenarios which encompasses 12336 car samples and 11693 non-car samples (background). An ATIS camera was positioned behind the windshield of a car to record different driving sessions. For training purposes, the N-Cars dataset is divided into 7940 car samples and 7482 background samples while the testing set consists of 4396 car samples and 4211 background samples. Each sample in the dataset has a duration of 100 milliseconds.

All experiments are conducted using the official split of the aforementioned datasets. To ensure fair comparisons across different methods, all influencing factors, such as the random seed, are kept consistent.

\subsubsection{Experimental Analysis}

All quantitative results can be found in Tab. 5 of the main paper. For the N-MNIST~\cite{orchard2015converting} dataset, EV-VGCNN~\cite{Ev-VGCNN} (0.994) and TORE~\cite{baldwin2022time} (0.994) achieved SoTA performance, while MVF-Net~\cite{deng2021mvf} (0.991) outperformed other methods on the N-Caltech101~\cite{orchard2015converting} dataset. Additionally, TORE~\cite{baldwin2022time} (0.977) demonstrated the best performance on the N-Cars~\cite{sironi2018hats} dataset. It is worth noting that the learning-based representations, specifically the Event-Spike-Tensor (EST)~\cite{EST}, exhibited promising performance compared to other event representations, such as HOTS. This highlights the superior performance of learning-based representations discussed in Sec. 2.1.3 of the main paper.

\subsection{Optical Flow Estimation}

\subsubsection{Experiment Setting and Dataset}
The authors discuss their approach to training and evaluating their event camera optical flow model. They utilize the ESIM dataset for pre-training, which provides full ground truth flow annotations. To evaluate their model's performance in different scenarios, they use the EV-IMO dataset for dynamic scenes and the Sintel dataset for general evaluation. By incorporating these datasets, the authors aim to enhance the robustness and accuracy of their model. The inclusion of diverse and challenging datasets helps in training the model on a wide range of real-world scenarios, enabling it to generalize better and perform well in various conditions. This approach ensures that the model is equipped with a comprehensive understanding of different visual characteristics, leading to improved performance and better overall results.

The authors describe the training process for their model using the FlyingChairs2 dataset. They train the model for 100 epochs. They adopted a batch size of 8. The training process takes approximately 42 hours on two NVIDIA 2080Ti GPUs. They use the AdamW optimizer with specific parameters and apply geometric and photometric augmentations to the data. The OneCycle policy is used for learning rate scheduling. After training, the pre-trained model is evaluated on the MVSEC dataset using different time intervals.

\subsubsection{Experimental Analysis}
The authors present their model's performance in Tab.~\ref{tab:optical_flow}, showcasing its SoTA performance on indoor sequences for both Endpoint Error and outlier metrics. While their model may not achieve the best EPE on the outdoor day1 sequence, it excels in the outlier metric, and its performance on indoor1-3 sequences is sufficiently impressive to demonstrate its advantage. Particularly in the evaluation setting with longer time intervals (dt=4) and larger motion, their results show significant improvement compared to other methods. Moreover, their model achieves comparable performance to existing SoTA methods that rely on two-frame inputs, despite being a dense optical flow estimation method using a single image with events. This demonstrates the effectiveness of their framework in accurately predicting dense flow by combining the information from the first image and events.

\begin{figure*}[t!]
    \centering
    \includegraphics[width=0.75\textwidth]{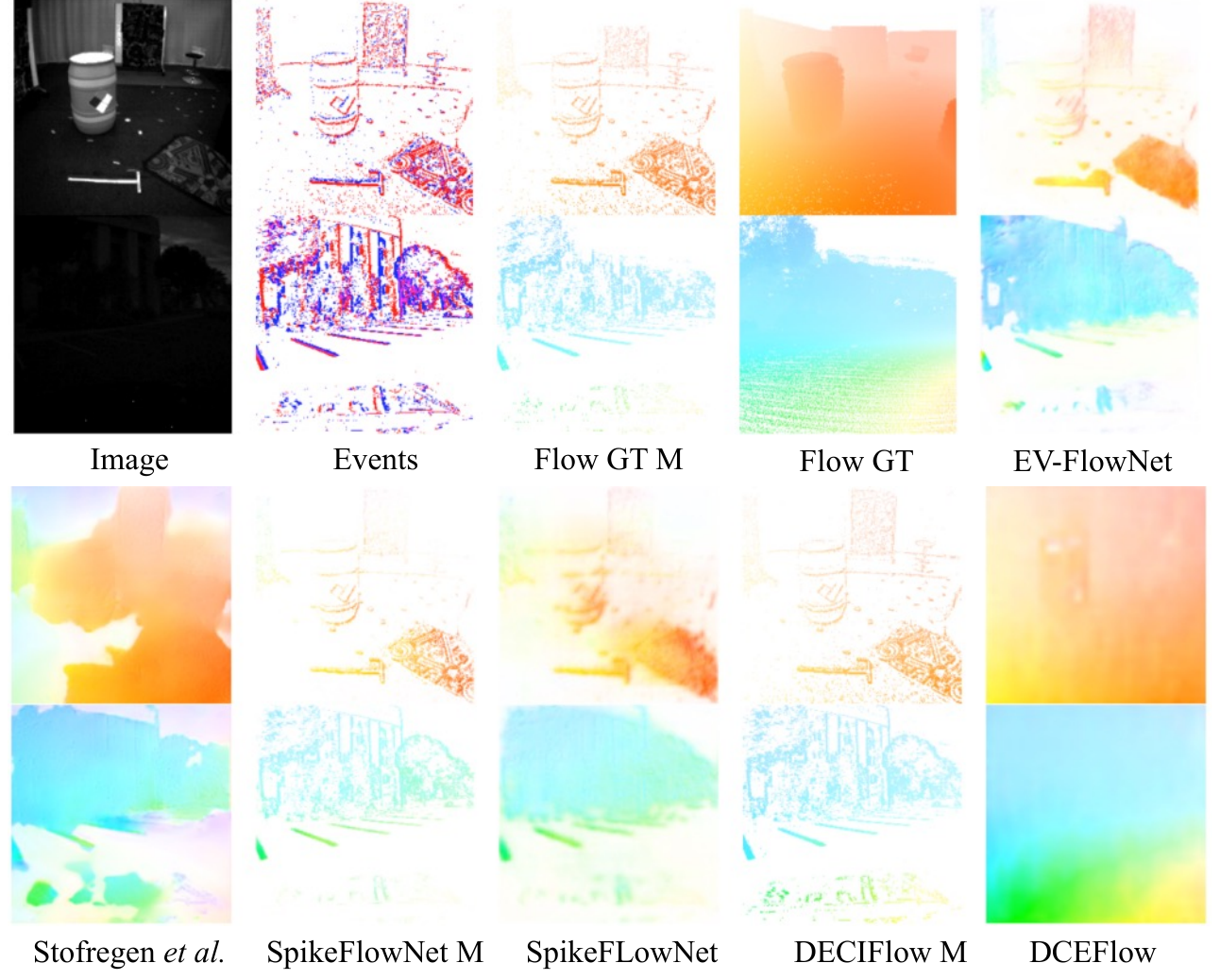}
    \caption{Qualitative comparison on the MVSEC dataset. M means that the masked flow at the pixels with events. (EV-FlowNet~\cite{LIF-EV-FlowNet}, Stoffregen \etal ~\cite{stoffregen2020reducing}, SpikeFlowNet~\cite{lee2020spike}, DCEIFlow~\cite{DECIFlow}.)}
    \label{fig: opticalflow}
\end{figure*}

The authors perform visual comparisons with various event-based methods, as shown in Fig.~\ref{fig: opticalflow}. They compare their proposed DCEIFlow model with EV-FlowNet, SpikeFlowNet, and Stoffregen \etal. They observe that EV-FlowNet and SpikeFlowNet, which rely only on events, produce sparse flow predictions with many incorrect predictions. 
The method proposed by Stoffregen faces challenges in predicting complete dense flow in regions without events, resulting in incomplete predictions. The MVSEC dataset, which is affected by camera motion, introduces spatial inconsistencies between images and optical flow, making it difficult to perform visual comparisons. Despite these challenges, the authors conclude that their DCEIFlow model performs better with fewer errors and more dense estimations, aligning with the quantitative comparisons conducted earlier.

\begin{table*}[t!]
\centering
\caption{\textbf{Tab. 10 in the main paper.} Experiments of representative methods on event-based optical flow estimation from ~\cite{weng2022boosting}. \\ 
UL: unsupervised learning. SL: supervised learning. MB: model-based methods. ($\cdot$): evaluation on both \textit{$outdoor\_day1$} and \textit{$outdoor\_day2$} sequences. 
[$\cdot$]: evaluation on \textit{$outdoor\_day2$} sequences. N/A means no results are available.
The results without any brackets mean that they are not trained on any sequence of MVSEC.)
}
\label{tab:optical_flow}
\vspace{-10pt}
\setlength{\tabcolsep}{4mm}
\resizebox{0.99\textwidth}{!}{
\begin{tabular}{cccccccccccc}
\toprule
\multirow{2}{*}{\begin{tabular}[c]{@{}c@{}}Type\end{tabular}} & \multirow{2}{*}{Method} & \multirow{2}{*}{Metric} & \multicolumn{2}{c}{$indoor\_flying1$} & \multicolumn{2}{c}{$indoor\_flying2$} & \multicolumn{2}{c}{$indoor\_flying3$} & \multicolumn{2}{c}{$outdoor$\_$day1$} & \multirow{2}{*}{Param.}\\ \cmidrule{4-11} 
 &  &  & EPE & $\%Out$ & EPE & $\%Out$ & EPE & $\%Out$ & EPE & $\%Out$ \\ \midrule
\multirow{9}{*}{UL} & Ev-FlowNet~\cite{zhu2018ev} & sparse & (1.03) & (2.2) & (1.72) & (15.1) & (1.53) & (11.9) & {[}0.49{]} & {[}0.2{]} & N/A \\  \cmidrule{2-12}
 & Zhu \etal ~\cite{2019Unsupervised31} & sparse & (0.58) & (0.0) & (1.02) & (4.0) & (0.87) & (3.0) & {[}0.32{]} & {[}0.0{]} & N/A \\ \cmidrule{2-12}
 & Matrix-LSTM~\cite{Matrixlstm} & sparse & (0.82) & (0.53) & (1.19) & (5.59) & (1.08) & (4.81) & N/A & N/A & N/A\\ \cmidrule{2-12} 
 & Spike-FLowNet~\cite{lee2020spike} & sparse & {[}0.84{]} & N/A & {[}1.28{]} & N/A & {[}1.11{]} & N/A & {[}0.49{]} & N/A & 13.039\\ \cmidrule{2-12}
 & Paredes \etal ~\cite{paredes2021back} & sparse & (0.79) & (1.2) & (1.40) & (10.9) & (1.18) & (7.4) & {[}0.92{]} & {[}5.4{]} & N/A\\ \cmidrule{2-12}
 & LIF-EV-FlowNet~\cite{LIF-EV-FlowNet} & sparse & 0.71 & 1.41 & 1.44 & 12.75 & 1.16 & 9.11 & 0.53 & 0.33 & N/A\\ \cmidrule{2-12}
 & Deng \etal~\cite{deng2021} & sparse & (0.89) & (0.66) & (1.31) & (6.44) & (1.13) & (3.53) & N/A & N/A & N/A\\ \cmidrule{2-12}
 & Li \etal~\cite{Li2020}  & sparse & (0.59) & (0.83) & (0.64) & (2.26) & N/A & N/A & {[}0.31{]} & {[}0.03{]} & N/A\\\cmidrule{2-12}
 & STE-FlowNet~\cite{STE-FlowNet} & sparse & {[}0.57{]} & {[}0.1{]} & {[}0.79{]} & {[}1.6{]} & {[}0.72{]} & {[}1.3{]} & {[}0.42{]} & {[}0.0{]} & N/A\\ \midrule
\multirow{4}{*}{SL} & Stoffregen \etal ~\cite{stoffregen2020reducing}&  dense & \textbf{0.56} & 1.00 & 0.66 & 1.00 & 0.59 & 1.00 & 0.68 & 0.99 & N/A\\ \cmidrule{2-12}
 & EST~\cite{EST} & sparse & (0.97) & (0.91) & (1.38) & (8.20) & (1.43) & (6.47) & N/A & N/A & N/A\\ \cmidrule{2-12}
 & {DCEIFlow~\cite{DECIFlow}} & \multicolumn{1}{c}{dense} & \multicolumn{1}{c}{\textbf{0.56}} & \multicolumn{1}{c}{0.28} & \multicolumn{1}{c}{0.64} & \multicolumn{1}{c}{0.16} & \multicolumn{1}{c}{0.57} & \multicolumn{1}{c}{0.12} & \multicolumn{1}{c}{0.91} & \multicolumn{1}{c}{0.71} & \multicolumn{1}{c}{N/A}\\ \cmidrule{2-12}
 & {DCEIFlow~\cite{DECIFlow}} & \multicolumn{1}{c}{sparse} & \multicolumn{1}{c}{0.57} & \multicolumn{1}{l}{0.30} & \multicolumn{1}{c}{0.70} & \multicolumn{1}{c}{0.30} & \multicolumn{1}{c}{0.58} & \multicolumn{1}{c}{0.15} & \multicolumn{1}{c}{0.74} & \multicolumn{1}{c}{0.29} & \multicolumn{1}{c}{N/A}\\ \midrule 
\multirow{4}{*}{MB} & \multicolumn{1}{c}{Pan \etal ~\cite{pan2020single}} & \multicolumn{1}{c}{sparse} & \multicolumn{1}{c}{0.93} & \multicolumn{1}{c}{0.48} & \multicolumn{1}{c}{0.93} & \multicolumn{1}{c}{0.48} & \multicolumn{1}{c}{0.93} & \multicolumn{1}{l}{0.48} & \multicolumn{1}{l}{0.93} & \multicolumn{1}{c}{0.48} & \multicolumn{1}{c}{N/A}\\ \cmidrule{2-12}
 & Shiba~\cite{shiba2022secrets} & sparse & 0.42 & 0.10 & 0.60 & 0.59 & 0.50 & 0.28 & 0.30 & 0.10 & N/A\\\cmidrule{2-12}
 & $\text{Fusion-FlowNet}_{Early}$~\cite{Fusion-FlowNet} & dense & (\textbf{0.56}) & N/A & (0.95) & N/A & (0.76) & N/A & {[}0.59{]} & N/A & 12.269\\ \cmidrule{2-12}
 & $\text{Fusion-FlowNet}_{Late}$~\cite{Fusion-FlowNet} & sparse & (0.57) & N/A & (0.99) & N/A & (0.79) & N/A & {[}0.55{]} & N/A & 7.549\\ \bottomrule
\end{tabular}
}
\end{table*}

\subsection{Depth estimation}
\subsubsection{Experiment Setting and Dataset}
The dataset called MVSEC (Multi-Vehicle Stereo Event Camera)~\cite{Zhu2018TheMS} comprises recorded data from a stereo rig consisting of two DAVIS event cameras. These cameras have a resolution of 346 × 260 pixels and a focal length of $83^{\circ}$, providing a wide horizontal field of view. LiDAR is utilized  to measure ground truth depth. MVSEC offers a diverse collection of driving sequences, both during the day and at night, as well as indoor sequences captured using a quadcopter. The depth maps were recorded at 20Hz, while the grayscale images were captured at 10 Hz for nighttime sequences and 45 Hz for daytime sequences.

To evaluate various methods for depth estimation, five common metrics are employed: absolute relative error (Abs.Rel.), logarithmic mean squared error (RMSELog), scale-invariant logarithmic error (SILog), accuracy ($\delta < 1.25^{n}$, n = 1, 2, 3), average absolute depth errors at different cut-off distances (i.e., 10m, 20m, and 30m), and running time (ms).

In Tab.~\ref{tab:depth_estimation}, the listed methods predict depth in the logarithmic scale, which is subsequently normalized and restored to absolute values by multiplying with the maximum depth. To prevent excessively large depth values, the absolute depth is clipped at 80 m. For training purposes, the outdoor day2 sequence is utilized, while testing is conducted on four sequences: outdoor day1 and outdoor night1 to outdoor night3. As an example, the implementation details of EReFormer~\cite{Liu2022EventbasedMD} indicate their adoption of the AdamW optimizer with a weight decay of 0.1. The learning rate follows a 1-cycle policy, with a maximum value of 3.2 × 10-5. Training the network spans 200 epochs, with a batch size of 2. All experiments were performed on NVIDIA Tesla V100-PCLE GPUs.

\subsubsection{Experimental Analysis}
DPT~\cite{Ranftl2021VisionTF}, which utilizes vision transformers, achieves superior performance compared to the best CNN-based method MDDE~\cite{HidalgoCarrio2020LearningMD}. This outcome provides evidence supporting the notion that leveraging global spatial information from sparse events aids in predicting more accurate depth maps in various scenarios. However, the performance of the spatial-temporal transformer in EReFormer~\cite{Liu2022EventbasedMD} highlights the sub-optimal result stemming from insufficient utilization of rich temporal cues derived from continuous event streams. This could potentially explain why EReFormer~\cite{Liu2022EventbasedMD} suffers minimal performance degradation when applied to night scenarios with higher noise levels.

\section{More qualitative comparison}
\label{sec5}
\begin{figure*}[t!]
    \centering
    \includegraphics[width=\textwidth]{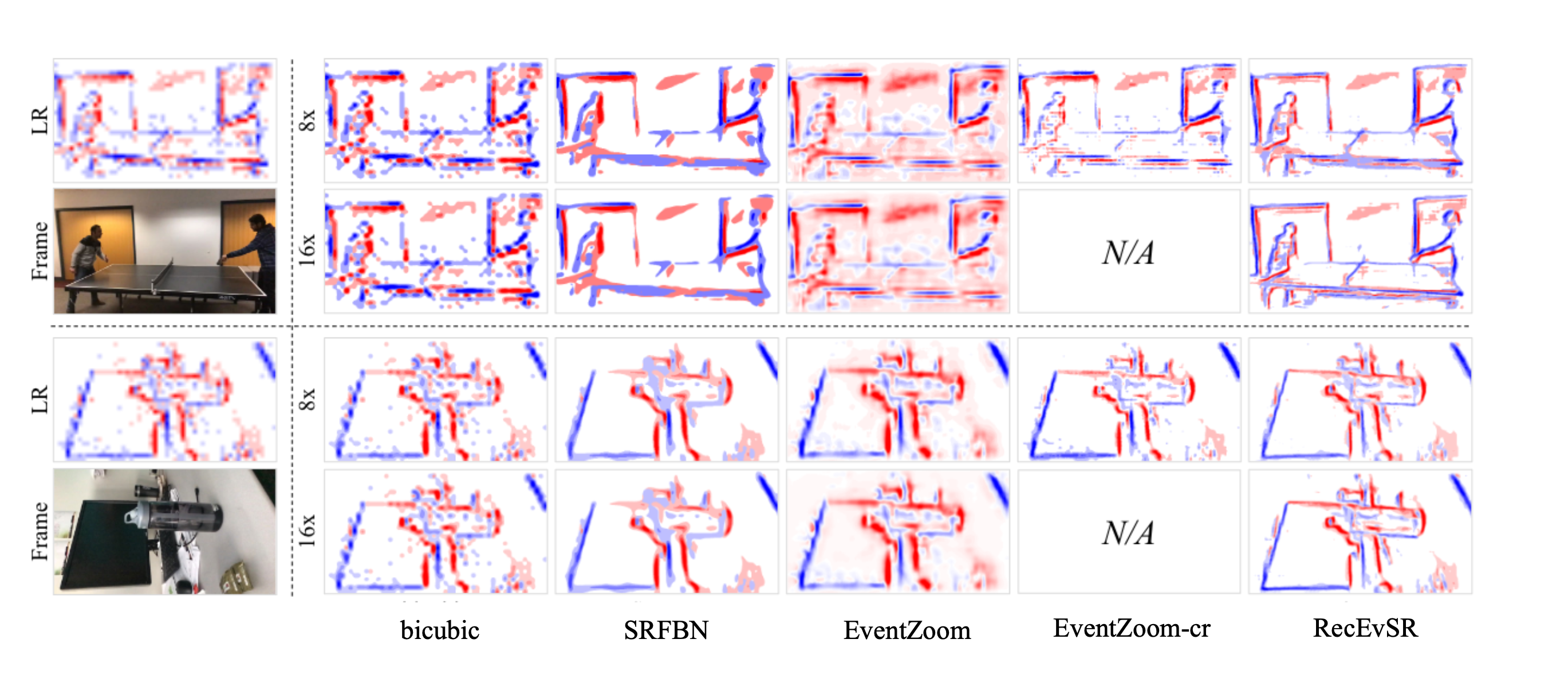}
    \caption{The visual comparisons among bicubic, SRFBN\cite{li2019feedback}, EventZoom\cite{duan2021eventzoom}, EventZoom-cr\cite{weng2022boosting} and RecEvSR \cite{weng2022boosting} for large factor SR (8(16) X) based on ENFS-real dataset.}
    \label{fig: SRVisual_comparisons}
\end{figure*}

Due to the lack of space in the main paper, we provide several experimental results of different visual tasks, including event super-resolution (Fig.~\ref{fig: SRVisual_comparisons} and Tab.~\ref{SR}) and high dynamic range (Tab.~\ref{tab:HDR}). 
The comparison results on synthetic datasets (ENFS-syn and RGB-DAVIS-syn) and real-world dataset (ENFS-real), as illustrated in Tab.~\ref{SR}. Notably, the method RecEvSR ~\cite{weng2022boosting} exhibits an impressive average 30\% boost in performance in terms of RMSE for 2(4)× SR, demonstrating a clear superiority with over 80\% average RMSE gain for 8(16)× SR on ENFS-syn. Against frame-based techniques such as bicubic and SRFBN~\cite{li2019feedback}, particularly for 8(16)× SR, the method showcases a notable relative gain of over 19\% regarding RMSE on ENFS-syn. Moreover, the integration of coordinate relocation upsampling in EventZoom-cr~\cite{duan2021eventzoom} leads to substantial enhancements in performance for large factor SR, achieving remarkable RMSE gains of 65.49\% for 8× SR on ENFS-syn and 44.18\% for 4× SR on RGB-DAVIS-syn. In real-world evaluations employing the ENFS-real dataset, the method surpasses EventZoom (EventZoom-cr)~\cite{duan2021eventzoom}by 20\% on average for 2(4)× SR and frame-based methods by 17\% in terms of RMSE.
Fig. 8 shows the visual outcomes of different super-resolution (SR) techniques, such as bicubic interpolation, SRFBN~\cite{li2019feedback}, EventZoom~\cite{duan2021eventzoom}, EventZoom-cr~\cite{duan2021eventzoom} and RecEvSR~\cite{weng2022boosting}, on the ENFSreal dataset. Notably, due to resource constraints, EventZoom-cr's 16× SR results are not presented. The RecEvSR exhibits superior performance compared to the baseline methods in handling large-factor SR tasks.

The Tab.~\ref{tab:HDR} displays a comparison of various state-of-the-art HDR imaging methods on a synthetic dataset~\cite{yang2023learning}, encompassing a frame-based approach: Liu\etal~\cite{liu2020single}; an event based method: the colored variant of E2VID~\cite{rebecq2019high}; three event-guided HDR image reconstruction methods: Han\etal~\cite{Han2020NeuromorphicCG}, eSL-Net~\cite{wang2020event} and HDRev-Net~\cite{yang2023learning}; along two exposure-based methods ~\cite{Debevec1997RecoveringHD, li2020fast}. The HDRev-Net combines event and frame and shows best performance under most of the metrics while the frame-based method Liu\etal~\cite{liu2020single} demonstrates the lowest LPIPS.



\begin{table*}[h!]
\centering
\caption{Qualitative comparison of Super-Resolution methods on synthetic and real-world datasets from ~\cite{weng2022boosting} based on RMSE. - means no results are available.}
\resizebox{0.7\linewidth}{!}{
\begin{tabular}{ccccccccccc}
\toprule
Methods  & 2x &  4x & 8x & 16x & & 2x & 4x & & 2x & 4x \\
\midrule
\multirow{1}{*}{} & \multicolumn{4}{c}{ENFS-syn} & \multicolumn{3}{c}{RGB-DAVIS-syn} & \multicolumn{3}{c}{ENFS-real}\\ \cmidrule{2-5} \cmidrule{7-8} \cmidrule{10-11}

bicubic  & 0.821 & 0.784 & 0.791 & 0.764 & & 0.387 & 0.378 &  &0.899 & 0.969 \\
SRFBN ~\cite{li2019feedback} & 0.694 & 0.690 & 0.708 & 0.678 & & 0.366 & 0.362 & & 0.669 & 0.753 \\
EventZoom ~\cite{duan2021eventzoom} & 0.843 & 1.036 & 2.385 & 5.970 & & 0.583 & 1.100 & & 0.773 & 0.910 \\
EventZoom-cr ~\cite{duan2021eventzoom} & 0.844 & 0.833 & 0.823 & -  & & 0.604 & 0.614 & & 0.775 & 0.828 \\
RecEvSR ~\cite{weng2022boosting}  & 0.686 & 0.653 & 0.617 & 0.582 & & 0.352 & 0.329 & & 0.663 & 0.663 \\
\bottomrule
\end{tabular}}
\label{SR}
\end{table*}

\begin{table*}[h!]
\centering
\caption{Qualitative comparison of HDR reconstruction methods on synthetic and real-world datase}
\label{tab:HDR}
\begin{tabular}{cccccc}
\toprule
Methods   & PSNR $\uparrow$   & SSIM $\uparrow$  & LPIPS $\downarrow$ & VDP $\uparrow$   & VQM $\downarrow$   \\
\midrule
eSL-Net~\cite{wang2020event}   & 16.575 & 0.713 & 0.413 & 5.903 & 0.467 \\
E2VID~\cite{Rebecq2019HighSA}     & 13.734 & 0.589 & 0.451 & 4.143 & 0.343 \\
Liu \etal ~\cite{liu2020single}       & 23.159 & 0.901 & 0.104 & 7.543 & 0.107 \\
Han \etal ~\cite{Han2020NeuromorphicCG}       & 20.697 & 0.861 & 0.208 & 6.709 & 0.243 \\
Debevec \etal ~\cite{Debevec1997RecoveringHD}   & 23.596 & 0.877 & 0.264 & 6.192 & 0.264 \\
Li \etal ~\cite{li2020fast}     & 20.673 & 0.890 & 0.151 & N/A     & N/A     \\
HDRev-Net~\cite{yang2023learning} & 24.071 & 0.928 & 0.110 & 8.108 & 0.103\\
\bottomrule
\end{tabular}
\end{table*}

\begin{table*}[t!]
\centering
\caption{Qualitative comparison of pure event-based depth estimation method on real dataset MVSEC~\cite{Zhu2018TheMS} outdoor split day1 and night1}
\label{tab:depth_estimation}
\begin{tabular}{cccccccccccc}
\toprule
\multicolumn{1}{l}{Datasets} & Methods    & Abs.Rel. $\downarrow$ & RMSELog $\downarrow$ & SILog $\downarrow$ & $\delta < 1.25$ $\uparrow$ & $\delta < 1.25^{2}$ $\uparrow$ & $\delta < 1.25^{3}$ $\uparrow$ & 10m $\downarrow$ & 20m $\downarrow$ & 30m $\downarrow$ & Runtime(ms) \\ \midrule
\multirow{4}{*}{day1} & MDDE~\cite{HidalgoCarrio2020LearningMD}      & 0.450      & 0.514     & 0.251   & 0.472              & 0.711               & 0.823               & 2.70  & 3.46  & 3.84  & 7.67        \\
& DTL~\cite{wang2020dual}      & 0.390      & 0.436     & 0.176   & 0.510              & 0.757               & 0.865               & 2.00  & 2.00  & 3.35  & 6.01        \\
& DPT~\cite{Ranftl2021VisionTF}       & 0.291      & 0.341     & 0.105   & 0.668              & 0.829               & 0.914               & 1.44  & 2.40  & 2.82  & 24.51       \\
& EReFormer~\cite{Liu2022EventbasedMD} & 0.271      & 0.333     & 0.102   & 0.664              & 0.831               & 0.923               & 1.29  & 2.14  & 2.59  & 35.17      \\ \midrule
\multirow{4}{*}{night1} & MDDE~\cite{HidalgoCarrio2020LearningMD}     & 0.770 & 0.638 & 0.346 & 0.327 & 0.582 & 0.732 & 5.36 & 5.32 & 5.40 & 7.67        \\
& DTL~\cite{wang2020dual}      &  0.474 & 0.555 & 0.299 & 0.429 & 0.657 & 0.791 & 2.61 & 3.11 & 3.82 & 6.01        \\
& DPT~\cite{Ranftl2021VisionTF}       & 0.344 & 0.405 & 0.153 & 0.564 & 0.768 & 0.891 & 1.80 & 2.67 & 3.22 & 24.51       \\
& EReFormer~\cite{Liu2022EventbasedMD} & 0.317 & 0.415 & 0.158 & 0.547 & 0.753 & 0.881 & 1.52 & 2.28 & 2.98 & 35.17      \\

\bottomrule
\end{tabular}
\end{table*}

\clearpage
\clearpage
\bibliographystyle{IEEEtran}
\bibliography{supplref}